\documentclass[10pt,twocolumn,letterpaper]{article}

\usepackage[pagenumbers]{cvpr} 

\usepackage{graphicx}
\usepackage{amsmath}
\usepackage{amssymb}
\usepackage{booktabs}
\usepackage{multirow}
\usepackage{color}
\usepackage{microtype}
\usepackage{makecell}
\makeatletter
\@namedef{ver@everyshi.sty}{}
\makeatother
\usepackage{pgfplots}
\usepackage{bbm}
\usepackage{enumitem}
\usepackage{diagbox}

\setlist[itemize]{leftmargin=*}

\pgfplotsset{compat=1.3}
\usepackage[pagebackref,breaklinks,colorlinks]{hyperref}
\renewcommand{\vec}[1]{\boldsymbol{#1}}

\DeclareMathOperator*{\argmax}{\arg\!\max}

\usepackage[capitalize]{cleveref}
\crefname{section}{Sec.}{Secs.}
\Crefname{section}{Section}{Sections}
\Crefname{table}{Table}{Tables}
\crefname{table}{Tab.}{Tabs.}

\definecolor{colorflatblue}{RGB}{31,119,180}
\definecolor{colorflatorange}{RGB}{255,127,14}

\def\cvprPaperID{5733} 
\def\confName{CVPR}
\def\confYear{2023}

\begin{document}

\title{3D Line Mapping Revisited}

\author{Shaohui Liu$^{1}$ \quad
Yifan Yu$^{1}$ \quad
R\'emi Pautrat$^{1}$ \quad
Marc Pollefeys$^{1,2}$ \quad
Viktor Larsson$^{3}$ \\
$^{1}$Department of Computer Science, ETH Zurich \quad $^{2}$Microsoft \quad $^{3}$Lund University \\
}
\maketitle

\begin{abstract}
In contrast to sparse keypoints, a handful of line segments can concisely encode the high-level scene layout, as they often delineate the main structural elements. 
In addition to offering strong geometric cues, they are also omnipresent in urban landscapes and indoor scenes.
Despite their apparent advantages, current line-based reconstruction methods are far behind their point-based counterparts.
In this paper we aim to close the gap by introducing LIMAP,
a library for 3D line mapping that robustly and efficiently creates 3D line maps from multi-view imagery. This is achieved through revisiting the degeneracy problem of line triangulation, carefully crafted scoring and track building, and exploiting structural priors such as line coincidence, parallelism, and orthogonality.
Our code integrates seamlessly with existing point-based Structure-from-Motion methods and can leverage their 3D points to further improve the line reconstruction. 
Furthermore, as a byproduct, the method is able to recover 3D association graphs between lines and points / vanishing points (VPs).
In thorough experiments, we show that LIMAP significantly outperforms existing approaches for 3D line mapping.
Our robust 3D line maps also open up new research directions. We show two example applications: visual localization and bundle adjustment, where integrating lines alongside points yields the best results. Code is available at \href{https://github.com/cvg/limap}{\color{cyan}{https://github.com/cvg/limap}}.

\end{abstract}

\section{Introduction}

\label{sec::intro}
\begin{figure}[tb]
\setlength\tabcolsep{6pt} 
\begin{tabular}{cc}
{\includegraphics[trim={1000 200 1000 200}, clip, width=0.34\linewidth, height=70pt]{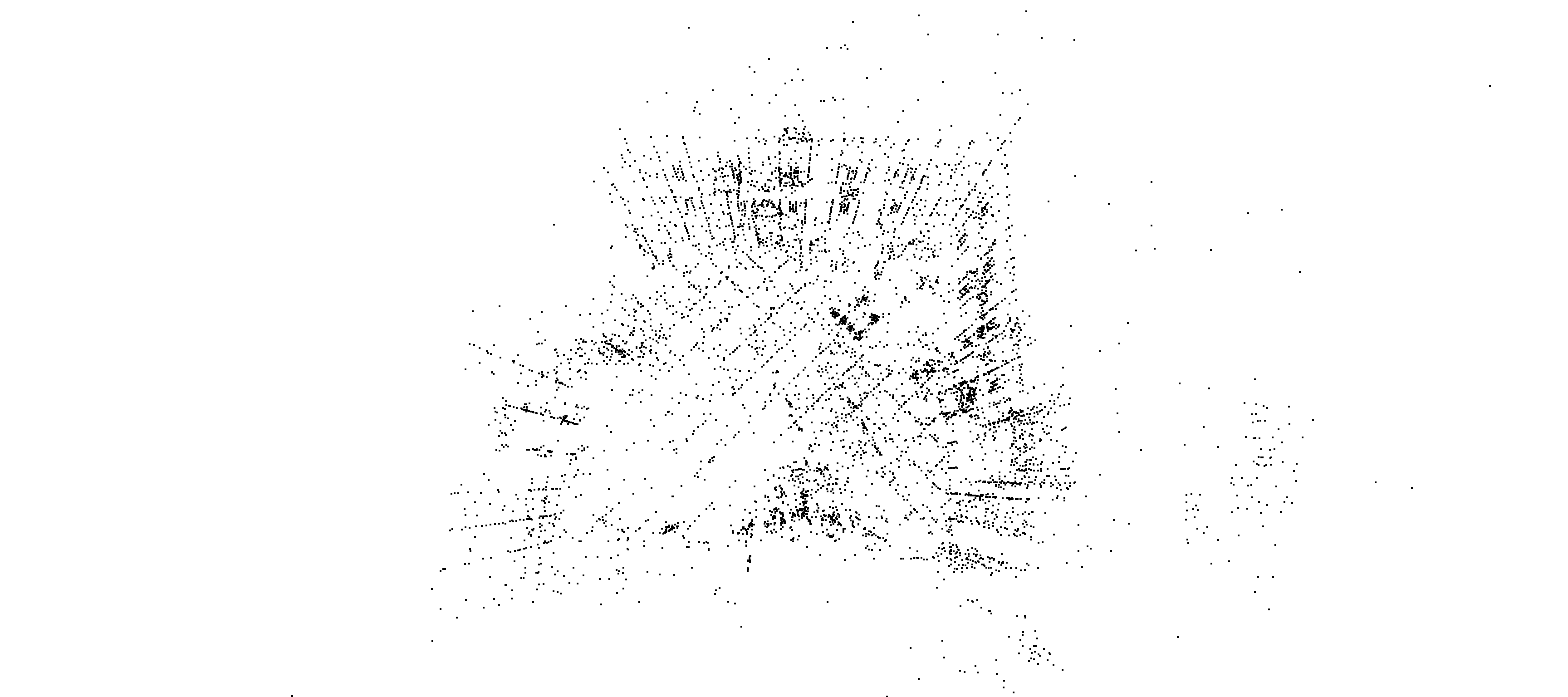}} & 
{\includegraphics[trim={620 50 600 50}, clip, width=0.34\linewidth, height=70pt]{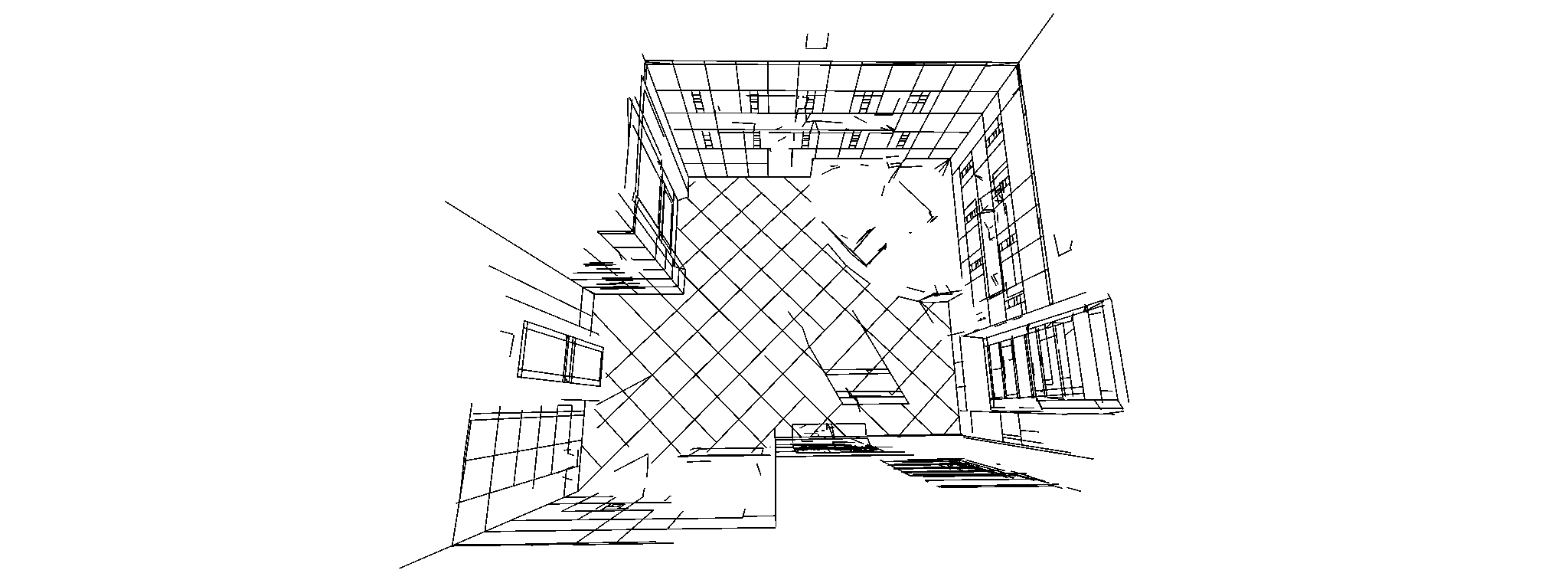}} \\
(a) Point mapping \cite{schonberger2016structure,detone2018superpoint} & (b) Line mapping \\
{\includegraphics[trim={410 30 350 50}, clip, width=0.34\linewidth, height=70pt]{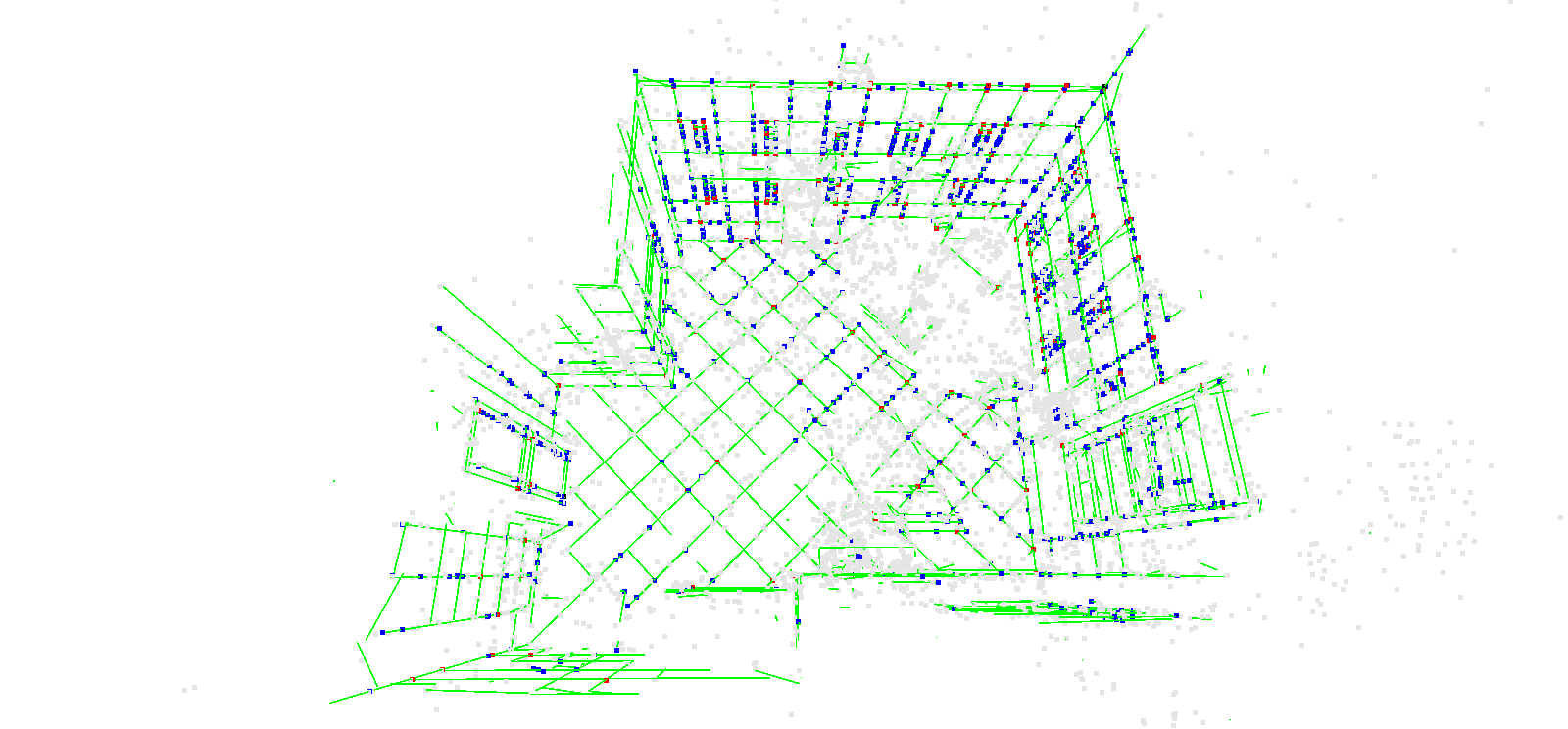}} &
{\includegraphics[trim={500 80 350 40}, clip, width=0.34\linewidth, height=70pt]{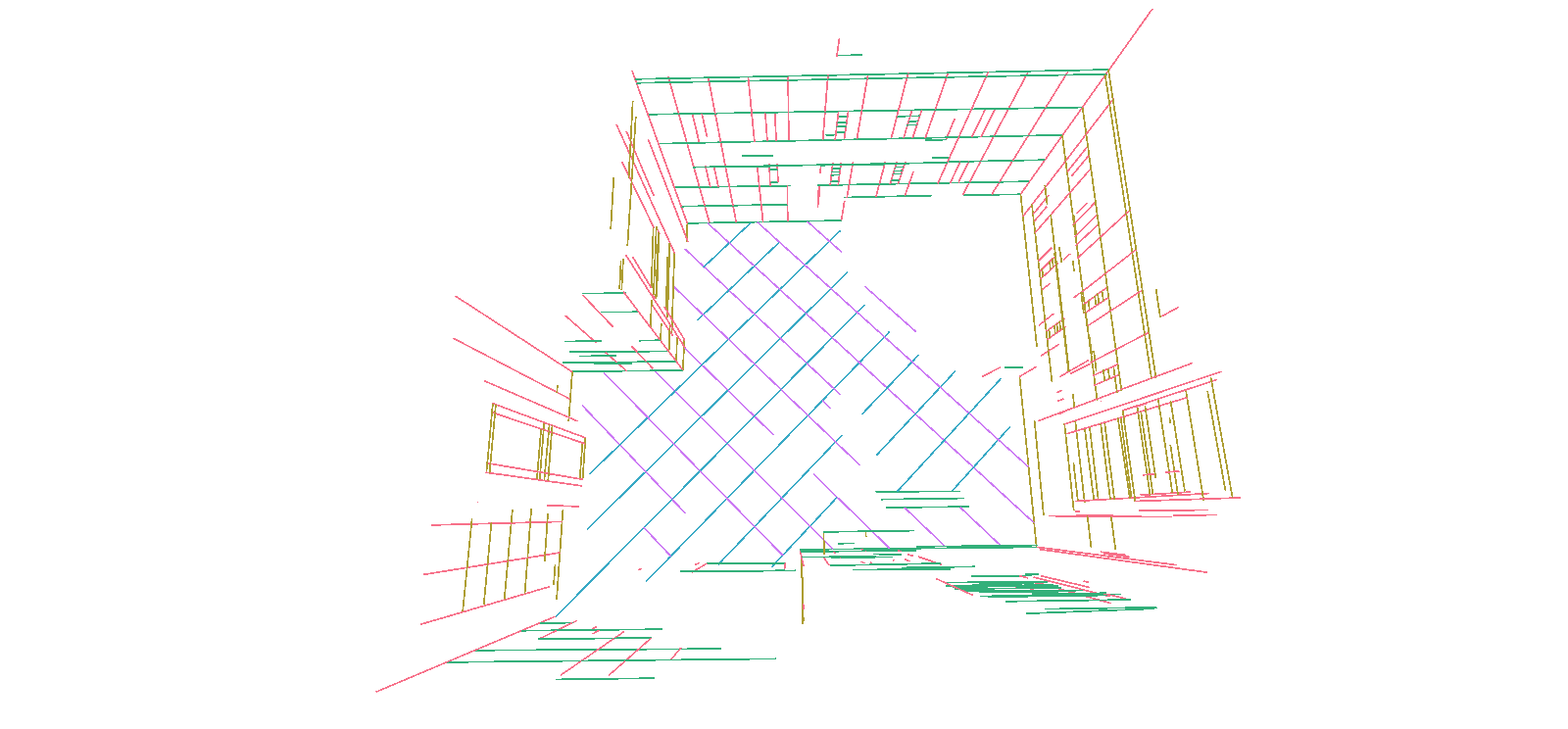}} \\
(c) Line-point association & (d) Line-VP association \\
\end{tabular}
\centering
\caption{In this paper, we propose a robust pipeline for mapping 3D lines (b), which offers stronger geometric clues about the scene layout compared to the widely used point mapping (a). Part of the success of our pipeline attributes to the modeling of structural priors such as coincidence (c), and parallelism / orthogonality (d). The corresponding 3D association graphs between lines and points / vanishing points (VPs) are also recovered from our system as a byproduct. The degree-1 point and degree-2 junctions are colored in blue and red respectively in (c), while parallel lines associated with the same VP are colored the same in (d).}
\label{fig::teaser}
\end{figure}

The ability to estimate 3D geometry and build sparse maps via Structure-from-Motion (SfM) has become ubiquitous in 3D computer vision. 
These frameworks enable important tasks such as building maps for localization~\cite{sattler2011}, providing initial estimates for dense reconstruction and refinement~\cite{schoenberger2016mvs}, and novel view synthesis~\cite{nerf,pittaluga2019revealing}. 
Currently, the field is dominated by point-based methods in which 2D keypoints are detected, matched, and triangulated into 3D maps~\cite{heinly2015,schonberger2016structure}.
These sparse maps offer a compact scene representation, only reconstructing the most distinctive points. 

While there have been tremendous progress in point-based reconstruction methods, they still struggle in scenes where it is difficult to detect and match sufficiently many stable keypoints, such as in indoor areas.
On the contrary, these man-made scenes contain abundant lines, e.g.~in walls, windows, doors, or ceilings. Furthermore, lines exhibit higher localization accuracy with less uncertainty in pixels \cite{forstner2016photogrammetric}. 
Last but not least, lines appear in highly structured patterns, often satisfying scene-wide geometric constraints such as co-planarity, coincidence (line intersections), parallelism, and orthogonality. 
In practice, lines suffer from different issues, such as poor endpoint localization and partial occlusion.
However, recent line detectors and matchers are bridging the gap of performance between points and lines~\cite{huang2018learning,pautrat2021sold2,yoon2021line}, making it timely to revisit the line reconstruction problem.

Despite their rich geometric properties and abundance in the real world, there exist very few line-based reconstruction methods in the literature~\cite{hofer2015line3d,micusik2017structure,hofer2017efficient,wei2022elsr}.
In practical applications, they have also not achieved the same level of success as their point-based counterparts. 
We believe this is due to several intrinsic challenges specific to line mapping:
\begin{itemize}[noitemsep,nolistsep]
    \item \textbf{Inconsistent endpoints.} Due to partial occlusion, lines often have inconsistent endpoints across images.
    \item \textbf{Line fragmentation.} In each image there might be multiple line segments that belong to the same line in 3D. This makes the process of creating track associations more complex compared to building 3D point tracks.
    \item \textbf{No two-view geometric verification.} While point matches can be verified in two views via epipolar geometry, lines require at least three views to filter.
    \item \textbf{Degenerate configurations.} In practice line triangulation is more prone to unstable configurations (see \cref{fig::synthetic_tests}), e.g.~becoming degenerate whenever the line is parallel with the camera motion (i.e.~to epipolar lines).
    \item \textbf{Weaker descriptor-based matching.} State-of-the-art descriptors for line segments are far behind their point-based counterparts, putting more emphasis on geometric verification and filtering during reconstruction. 
\end{itemize}

In this paper we aim to reduce the gap between point-based and line-based mapping solutions.
We propose a new robust mapping method, LIMAP, that integrates seamlessly into existing open-source point-based SfM frameworks \cite{snavely2006photo,wu2011visualsfm,schonberger2016structure}.
By sharing the code with the research community we hope to enable more research related to lines; both for low-level tasks (such as improving line segment detection and description) and for integrating lines into higher-level tasks (such as visual localization or dense reconstruction).
In particular, we make the following contributions in the paper:
\begin{itemize}\itemsep0pt
\item We build a new line mapping system that \textbf{reliably reconstructs 3D line segments from multi-view RGB images}. Compared to previous approaches, our line maps are significantly more complete and accurate, while having more robust 2D-3D track associations.

\item We achieve this by \textbf{automatically identifying and exploiting structural priors} such as coincidence (junctions) and parallelism. \textbf{Our technical contribution spans all stages of line mapping} including triangulating proposals, scoring, track building, and joint optimization, with 3D line-point / VP association graphs output as a byproduct.

\item The framework is \textbf{flexible} such that researchers can easily change components (e.g.~detectors, matchers, vanishing point estimators, etc.) or integrate additional sensor data (e.g.~depth maps or other 3D information). 

\item We are the first to go beyond small test sets by quantitatively evaluating on both synthetic and real datasets to benchmark the performance, with hundreds of images for each scene, in which \textbf{LIMAP consistently and significantly outperforms existing approaches}.

\item Finally, we demonstrate the usefulness of having robust line maps by showing \textbf{improvement over purely point-based methods} in tasks such as visual localization and bundle adjustment in Structure-from-Motion.
\end{itemize}

\section{Related Work}

\label{sec::related_work}
\noindent
\textbf{Line Detection and Matching. }
Detecting 2D line segments conventionally relies on grouping image gradients 
\cite{von2008lsd,akinlar2011edlines}. To improve the robustness and repeatability, learning-based line detectors were later proposed to tackle the problem of wireframe parsing \cite{huang2018learning,zhou2019learning,xue2019learning,zhang2019ppgnet,xue2020holistically,meng2020lgnn}. Recent deep detectors \cite{huang2020tp,pautrat2021sold2,xu2021line} manage to achieve impressive results for detecting general line segments. Matching of the detected line segments is often based on comparing either handcrafted \cite{bay2005wide,wang2009msld,zhang2013efficient,verhagen2014scale} or learning-based \cite{lange2019dld,vakhitov2019learnable,pautrat2021sold2,yoon2021line,abdellali2021l2d2} descriptors. Some recent methods also exploit point-line \cite{fan2010line,fan2012robust} and line-junction-line structures \cite{li2014robust,li2016hierarchical} to improve matching results, yet still not reaching the reliability level of advanced point matchers \cite{sarlin2020superglue,sun2021loftr}. Our method can leverage any line detector and matcher, and is robust to outliers.

~\\
\noindent
\textbf{Line Reconstruction.}
As a seminal work, Bartoli and Sturm \cite{bartoli2004framework,bartoli2005structure} proposed a full SfM pipeline for line segments, later improved by Schindler \cite{schindler2006line} with Manhattan-world assumption \cite{coughlan2000manhattan}. 
Jain et al. \cite{jain2010exploiting} proposed to impose global topological constraints between neighboring lines, which were further explored in \cite{ramalingam2015line,ranade2018novel,ren2021intuitive} to build wireframe models. Some learning-based methods \cite{zhou2019learning,luo2022LC2WF} were introduced as well to predict 3D wireframes. Hofer et al. \cite{hofer2014improving,hofer2015line3d,hofer2017efficient} proposed checking weak epipolar constraints over exhaustive matches and graph clustering, and introduced the Line3D++ software (referred as L3D++ in this paper), which remains the top choice \cite{gao2022pose,luo2022LC2WF} for acquiring 3D line maps so far. Recently, ELSR \cite{wei2022elsr} employed planes and points to guide the matching. However, all prior work mainly shows qualitative results and provides quantitative evaluation only on relatively small image sets \cite{strecha2008benchmarking,jain2010exploiting}. In this paper, we set up a quantitative evaluation on benchmarks with hundreds of images, where our proposed system significantly surpasses prior work by improving all stages in the mapping pipeline.

~\\
\noindent
\textbf{Line-based Applications.}
The resulting 3D line maps can be used for many downstream applications. \cite{hofer2017efficient} advocates the complementary nature of line reconstruction for structure visualization. Some incremental line-based SfM systems are introduced in \cite{zhang2014structure,micusik2017structure,holynski2020reducing}. To improve quality and robustness, recent methods \cite{zuo2017robust,pumarola2017pl,he2018pl,gomez2019pl,wei2019real,lim2021avoiding,lim2022uv} jointly employ point and line features in SLAM. While their line maps are often noisy and incomplete, noticeable improvement has been achieved in the accuracy of the recovered camera motion. There has also been development on VP estimation \cite{bazin_cvpr_2012,zhang2015,Li_2019_ICCV,Qian2022ARO} and solvers for joint point-line pose estimation \cite{ramalingam2011pose,vakhitov2016accurate,zhou2018stable,agostinho2019cvxpnpl}. Recently, promising performance in visual localization has been achieved by combining point and line features in a refinement step~\cite{gao2022pose}. In this paper, we show that our line maps can benefit multiple applications such as localization, SfM, and MVS. In particular, we present very competitive results on point-line visual localization.

\begin{figure*}[tb]
\includegraphics[width=0.98\linewidth]{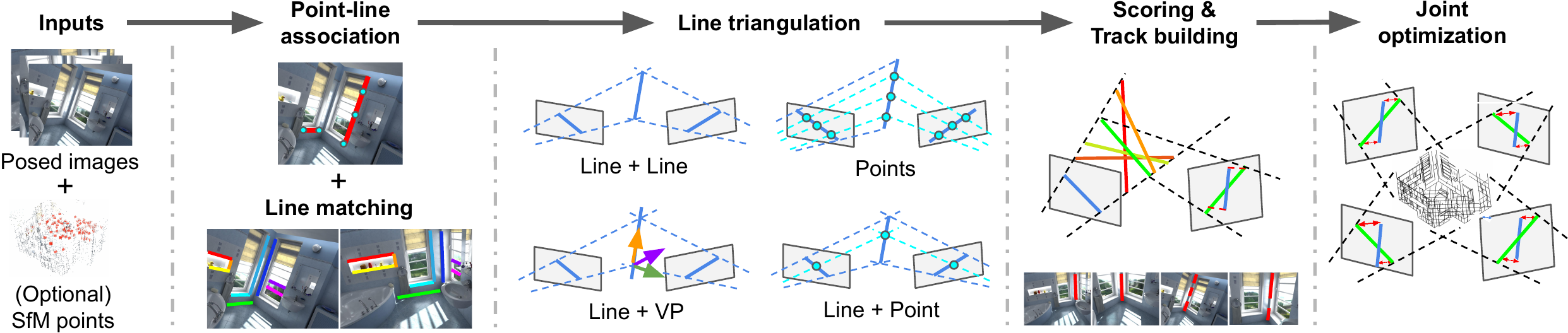}
\centering
\caption{\textbf{Overview.} Given a set of posed images and optional 3D points, we associate nearby points to lines, match the lines, triangulate them with 4 different strategies, score 3D line proposals, build line tracks, jointly optimize all features, before obtaining our final reconstruction.}
\label{fig::overview}
\end{figure*}

\section{The Proposed 3D Line Mapping Pipeline}

\label{sec::mapping}

We now present our proposed pipeline for 3D line mapping. Our method takes as input a set of images with 2D line segments from any existing line detectors. We assume the camera pose for each image is available (e.g.~from SfM/SLAM), and optionally we can also leverage a 3D point cloud (e.g.~obtained from point-based SfM). The pipeline consists of three main steps:
\begin{itemize}\itemsep0pt
    \item \textbf{Proposal Generation (Sec.~\ref{sec:proposals})}: For each 2D line segment, we generate a set of 3D line segment proposals.
    \item \textbf{Scoring and Track Association (Sec.~\ref{sec:scoring})}: Considering multi-view consistency, we score each proposal, select the best candidate for each 2D line, and associate them into a set of 3D line tracks.
    \item \textbf{Joint Refinement (Sec.~\ref{sec:refinement})}: Finally, we jointly perform non-linear refinement over the 3D line tracks along with 3D points and VP directions, integrating additional structural priors as soft constraints.
\end{itemize}
Figure~\ref{fig::overview} shows an overview of the overall pipeline. In the following sections, we detail each of the three main steps.

By design our pipeline is robust to scale changes and \textbf{we use the same hyper-parameters for all experiments across datasets}, which are provided in Sec. F.2 in the supp.

\subsection{Generating 3D Line Segment Proposals}\label{sec:proposals}

The first step is to generate a set of 3D line proposals for each 2D line segment.
Given a segment in an image, we use any existing line matcher to retrieve the top $K$ line matches in each of the $n_v$ closest images. Using the top $K$ line matches instead of a single match increases the chance of getting a correct match, while wrong matches will be filtered out in subsequent steps.

Let $(\vec{x}^r_1, \vec{x}^r_2) \in \mathbb{R}^3\times \mathbb{R}^3$ be the two endpoints (in homogeneous coordinates normalized by the intrinsics) for the reference line segment that we wish to generate proposals for. For ease of notation, we let the world-coordinate system align with the reference view. The endpoints of the 3D line proposals that we generate can all be written as
\begin{equation}
    \vec{X}_1 = \lambda_1 \vec{x}_1^r, \quad \vec{X}_2 = \lambda_2 \vec{x}_2^r ,
\end{equation}
for some values of $\lambda_1, \lambda_2 \in \mathbb{R}$. Having the 3D endpoints of all proposals lie on the camera rays of the 2D endpoints simplifies the scoring procedure in the second step (Sec.~\ref{sec:scoring}).

\subsubsection{Line Triangulation}
For each matched 2D line segment $(\vec{x}^m_1, \vec{x}^m_2)$ we generate one proposal via algebraic line triangulation. Let $(R^m, \vec{t}^m)$ be the camera pose of the matched view. We can then solve linearly for the endpoint ray depths $\lambda_i$ as
\begin{equation} \label{eq:depth_constraint}
   (\vec{x}^m_1 \times \vec{x}^m_2) ^T \left( R^m (\lambda_i\vec{x}_i^r) + \vec{t}^m\right) = 0,~~~i=1,2.
\end{equation}
The proposals are then filtered with cheirality checks (positive $\lambda$) and degeneracy check via the angle between ray $\vec{x}_i^r$ and $\vec{\ell_m} = \vec{x}_1^m \times \vec{x}_2^m$. Note that line triangulation becomes inherently unstable close to degenerate configurations when {\small $\vec{\ell}_m^TR^m\vec{x}^r_i = 0$}, where we get zero or infinite solutions from \eqref{eq:depth_constraint}. 
Geometrically, this happens when the line is parallel with the epipolar plane: If $\vec{\ell}_m^T\vec{t}^m \neq 0$ they have no intersection, otherwise they intersect fully and we get infinite solutions $\vec{\ell}_m \sim \vec{t}^m\times R^m\vec{x}_i^r = E\vec{x}_i^r$, i.e.~the line segment coincides with the epipolar line from $\vec{x}_i^r$. This issue is further illustrated in Figure~\ref{fig::synthetic_tests}. Since we solve for each $\lambda_i$ independently, the triangulation problem can have zero, one, or two degenerate endpoints. We term the case with one degenerate endpoint as a \textit{weakly degenerate} one, and the case with two degenerate endpoints as \textit{fully degenerate}. In contrast to the point case, two-view line triangulation is minimal such that any solution fits the measurements exactly with zero error, preventing filtering with 2D reprojection error at this stage. 

\subsubsection{Point-Line Association}
\label{sec::association_2d}
To obtain meaningful proposals in degenerate cases, we leverage additional geometric information coming from either points or associated vanishing points (VPs). 2D-3D point correspondences can either come from a point-based SfM model or be triangulated from matched endpoints/junctions. For each 2D line segment, we associate all 2D points within a fixed pixel threshold and thereby associate with their corresponding 3D points. For each image, we also estimate a set of VPs and their association to 2D lines using JLinkage~\cite{toldo2008robust}.

\subsubsection{Point-guided Line Triangulation}
\label{sec::point-guided-triangulation}
We now generate a second set of proposals for each 2D line segment with the assistance of the associated 2D-3D point correspondences and vanishing points. In the following parts we present three different methods. M1 employs multiple associated 3D points so it is stable for all cases including the \textit{fully degenerate} ones, while M2 and M3 with one known point / VP can help generate stable proposals in \textit{weakly degenerate} cases, which are more common in practice. Cheirality tests are applied to all proposals with respect to both views.

\noindent\textbf{M1. Multiple Points}. For each matched line segment we generate one proposal by collecting all of the associated 3D points that are common between the reference and the match. On top of those common points, we fit a 3D line that is then projected onto two camera rays corresponding to $\vec{x}_1^r$ and $\vec{x}_2^r$.

\noindent\textbf{M2. Line + Point}. For each matched line segment we also generate one proposal for each shared 3D point. We first project the 3D point onto the plane spanned by $\vec{x}_1^r$ and $\vec{x}_2^r$. We then aim to find a line that passes through the projection and minimizes the residuals in \eqref{eq:depth_constraint} to the matched line. This can be formulated as a quadratic optimization problem in the two endpoint depths $\vec{\lambda}=(\lambda_1,\lambda_2)$ with a single constraint:
\begin{equation}
    \min_{\vec{\lambda}\in\mathbb{R}^2} \vec{\lambda}^T A \vec{\lambda} + \vec{b}^T\vec{\lambda}, \quad \text{s.t.} \quad \vec{\lambda}^TQ\vec{\lambda} + \vec{q}^T\vec{\lambda} = 0 .
\end{equation}
Due to the low-dimensionality of the problem, a closed-form solution can be derived by reducing it to a univariate quartic polynomial. We show the full derivation in Sec. B in supp.

\noindent\textbf{M3. Line + VP}. Each VP corresponds to a 3D direction. For each associated VP, we generate one proposal based on its direction (again projected onto the plane spanned by $\vec{x}_1^r$ and $\vec{x}_2^r$). This gives a single linear constraint on the ray depths,
\begin{equation}
    \left(\vec{v} \times  (\vec{x}_1^r \times \vec{x}_2^r) \right)^T \left(\lambda_2 \vec{x}_2^r - \lambda_1\vec{x}_1^r \right) = 0 .
\end{equation}
where $\vec{v} \in \mathbb{R}^3$ is the VP. Using the constraint, we then solve for $\vec{\lambda}=(\lambda_1,\lambda_2)$ by minimizing the two residuals of \eqref{eq:depth_constraint} in a least squares sense. Note that $\vec{v}$ can either come from the reference image, or from a matched line in another image.

\noindent
\textbf{Extension: Line Mapping Given Depth Maps.}
\label{sec::fitnmerge}
The proposal generation step can be improved when each image has a corresponding depth map (e.g.~from an RGB-D sensor), which can be leveraged with robust line fitting to generate the 3D line proposals. Refer to Sec. E in our supplementary material for more details and results.

\subsection{Proposal Scoring and Track Association} \label{sec:scoring}

\begin{figure}[tb]
\scriptsize
\setlength\tabcolsep{3pt} 
\begin{tabular}{ccc}
\includegraphics[width=0.27\linewidth, height=50pt]{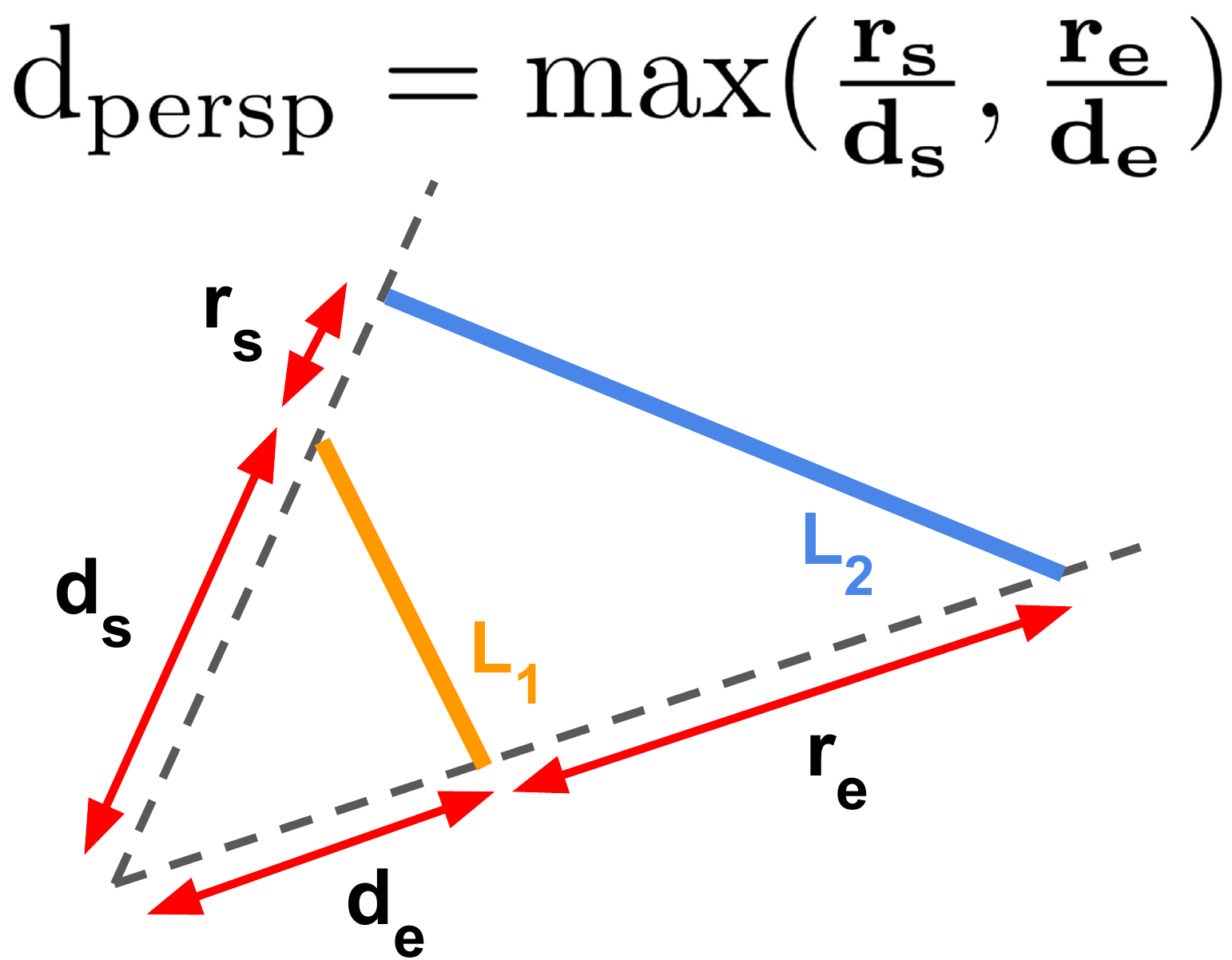} &
\includegraphics[width=0.33\linewidth, height=50pt]{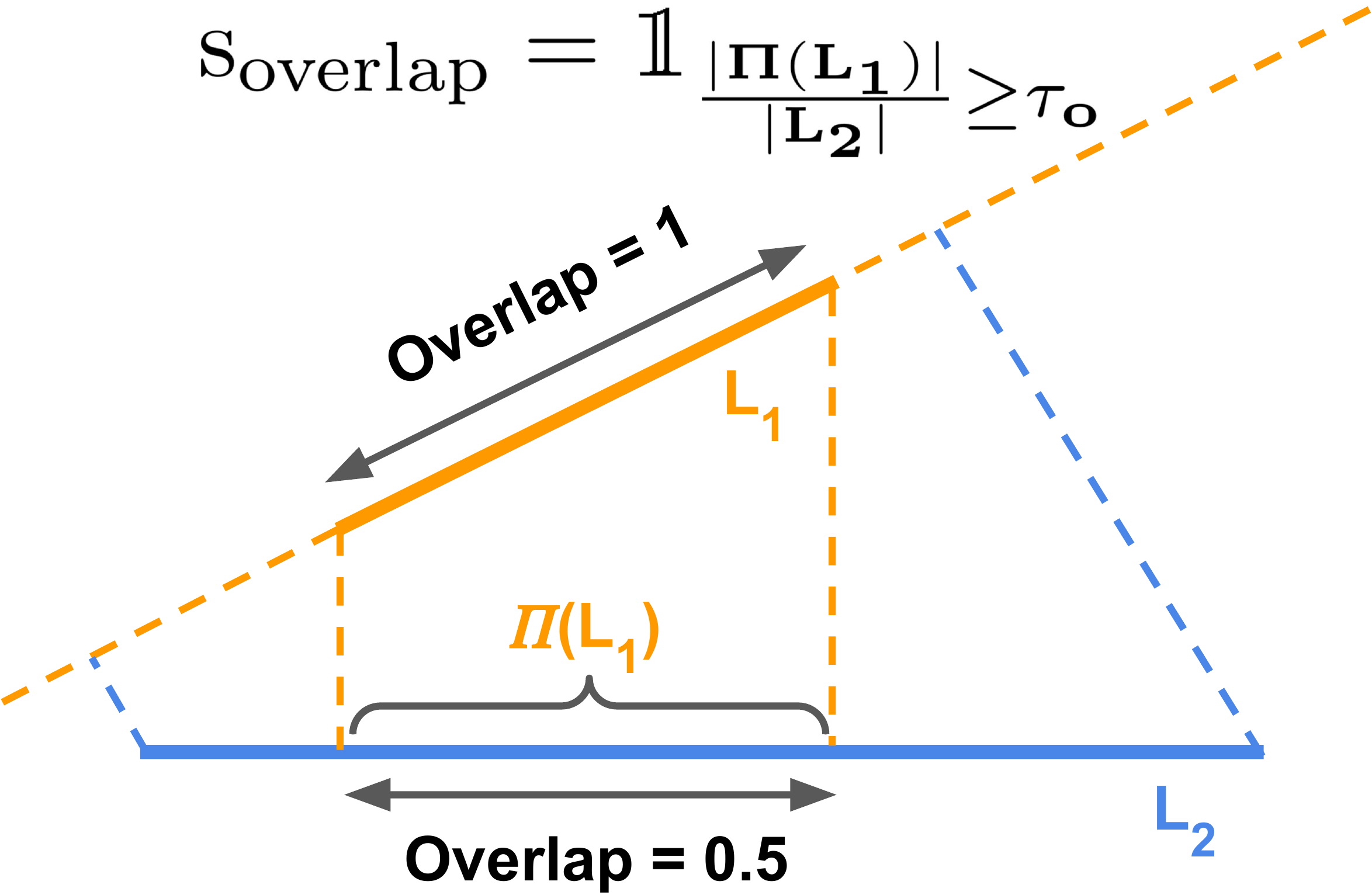} & 
\includegraphics[width=0.33\linewidth, height=50pt]{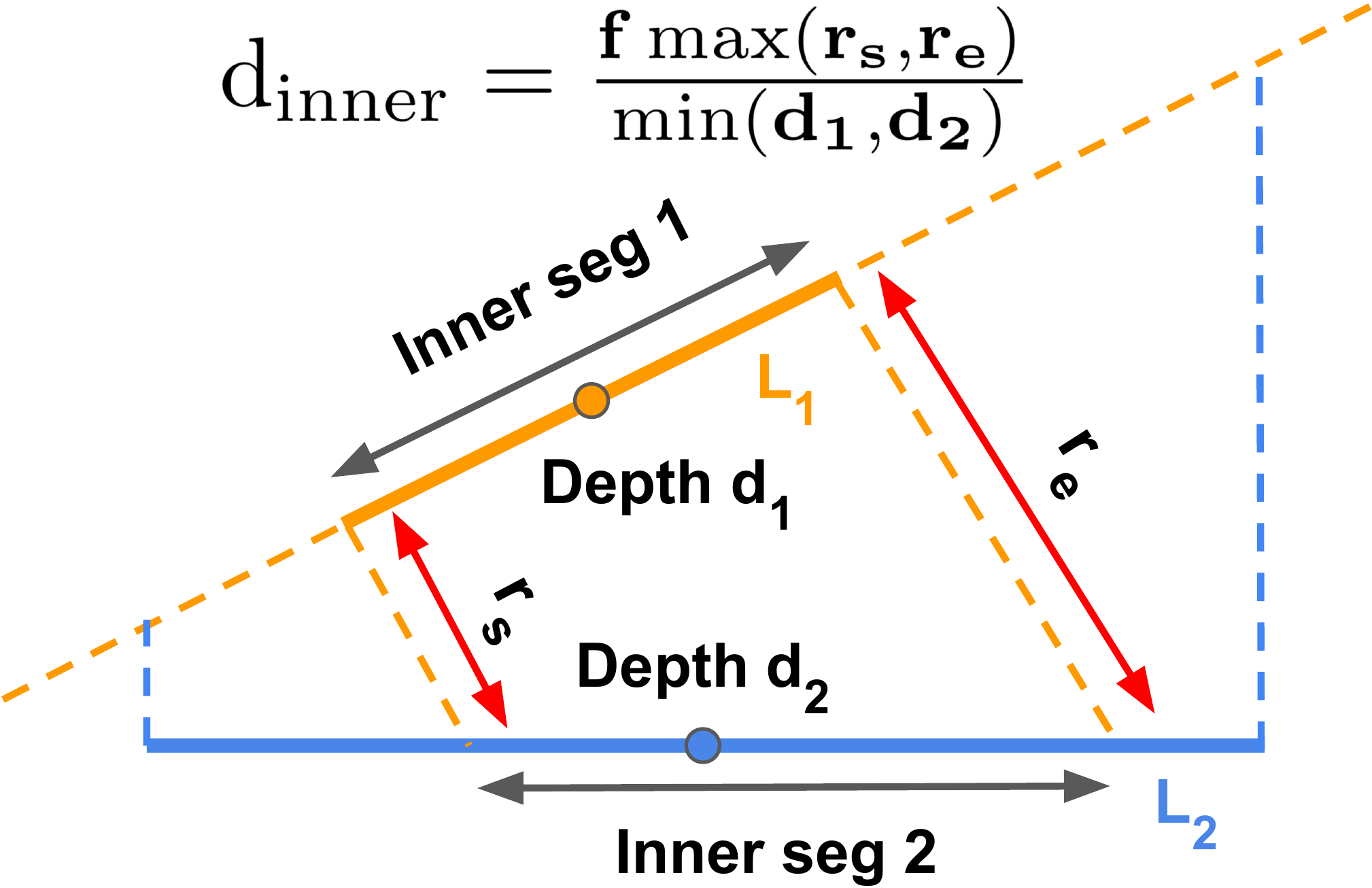} \\
(a) Perspective distance & (b) Overlap score & (c) InnerSeg distance
\end{tabular}
\centering
\caption{\textbf{Scoring methods.} We propose three novel line scoring measures that are scale-invariant and handle different line lengths.}
\label{fig::scoring}
\end{figure}

At this point, each 2D line segment $l$ in image $I$ is associated with a set $\mathcal{K}$ of 3D line segment proposals (stemming from the top $K$ line matches and various triangulations) for each neighboring image $J$. We describe in the following how we select the best 3D line proposal for each 2D line segment, and associate these lines into tracks. For each of these steps, we leverage different scoring methods quantifying the distance between two 3D line segments $(L_1, L_2)$. These distances are usually computed symmetrically and averaged, and can be obtained both in 3D and in 2D by projecting each 3D line into the other view. We start by presenting two classic ones, and then define our three novel line distances (one for 3D proposal selection and two for track building).

\begin{itemize}\itemsep0pt
    \item \textit{Angular distance}: angle between $L_1$ and $L_2$.
    \item \textit{Perpendicular distance}: maximum orthogonal distance of the endpoints of $L_1$ to the infinite line spanned by $L_2$.
\end{itemize}

\noindent
\textbf{3D Proposal Selection.}
To select best 3D candidate for each 2D line, we score each proposal $L_i$ by measuring its consistency with the others. Here we introduce a new distance:
\begin{itemize}\itemsep0pt
    \item \textit{Perspective distance}: assuming the endpoints of $L_1$ and $L_2$ are on the same rays as in \cref{fig::scoring}(a), the distance is defined as the endpoint distances, divided by the ray depths $d_s, d_e$ of the endpoints of $L_1$ in image 1. This score can filter out ill-posed triangulations (refer to Sec. F.3 in supp. for detailed discussions), while remaining scale-invariant.
\end{itemize}
This new distance, together with the \textit{angular distance} in 2D and 3D, and the \textit{perpendicular distance} in 2D, have different scales. In order to aggregate them together, we associate a scaling factor $\tau_r$ to each distance $r$ and get a normalized score $s_n = e^{-(r/\tau_r)^2} \in (0, 1]$. Denoting by $\mathcal{S}$ the set of all the corresponding normalized scores and $\mathbbm{1}$ the indicator function, the score between $L_1$ and $L_2$ becomes
\begin{equation}
    \label{eq:score_aggregation}
    s(L_1, L_2) = \min_{s_n \in \mathcal{S}} ( s_n \cdot \mathbbm{1}_{s_n \geq 0.5} ) \in \{0\}\cup[0.5,1] .
\end{equation}
Now equipped with unique score per line pair, we can consider all the neighboring 3D line candidates $L_j^k$ coming from the neighboring image $J$ and proposal $k$. The consistency score is defined by summing the best score from each image:
\begin{equation}
    s_c (L_i) = \sum_{J \in \mathcal{N}_I} \max_{k \in \mathcal{K}} s(L_i, L_J^k) ,
\end{equation}
where $\mathcal{N}_I$ is the set of neighboring images of $I$. The best 3D line candidate for each 2D line segment $l$ is then selected as the proposal with the highest score: $L = \argmax_{L_i} s_c (L_i)$. If the score is less than 1.0, i.e. the best candidate has less than two supports from neighboring views, we ignore this 2D line segment in the subsequent track building process.

\noindent
\textbf{Track Building.}
At this point, each 2D segment has been assigned a unique 3D line (its best 3D line candidate). The goal of this step is to gather these 2D segments into line tracks. For this, we form a graph where the 2D segments are nodes and all initial line matches are edges. We aim to prune edges in the graph such that the connected 2D segments share similar 3D assignments. We propose two new line scoring measures that can cope with different endpoint configurations and variable scales across images. 
\begin{itemize}\itemsep0pt
    \item \textit{Overlap score}: we project $L_1$ orthogonally onto $L_2$, clip the projected endpoints to the endpoints of $L_2$ if they fall outside of $L_2$ to get segment $\Pi(L_1)$, and compare the ratio of lengths to a threshold $\tau_o$: $\mathbbm{1}_{\frac{|\Pi(L_1)|}{|L_2|} \geq \tau_o}$ (see \cref{fig::scoring}(b)).
    \item \textit{InnerSeg distance}: the endpoints of $L_1$ are perpendicularly unprojected to $L_2$. If they fall outside of $L_2$, we clip them to the closest endpoint of $L_2$. By doing this in both directions, we can define two \textit{inner segments} (see \cref{fig::scoring}(c)), and the \textit{InnerSeg} distance as the maximum distance between their endpoints. To make this measure scale-invariant, we additionally divide it by a scale factor $\sigma = \frac{\min (d_1, d_2)}{f}$, where $d_j$ is the depth of the mid-point of $L_j$ in image $J$ and $f$ is the focal length. This encodes how far the mid-point can move in 3D before reaching 1 pixel error in the image (detailed in Sec. F.3 in supp.).
\end{itemize}
We then convert the \textit{InnerSeg distance} computed in 3D to a normalized score as in the previous paragraph, and combine it with the \textit{overlap score} in 2D and 3D and previous scores using \eqref{eq:score_aggregation}.
Given these pairwise scores of 3D lines, we can now prune edges whose score is below a threshold $t_f=0.5$. The connected components of the resulting graph yield the line tracks, ignoring components with less than 3 nodes.

For each track, we then re-estimate a single 3D line segment. Using the set of endpoints from the 3D assignments of all nodes in the track, we apply Principal Component Analysis (PCA) and use the principal eigenvector and mean 3D point to estimate the infinite 3D line. We then project all endpoints on this infinite line to get the new 3D endpoints.

\subsection{Joint Optimization of Lines and Structures} \label{sec:refinement}

Finally, we perform non-linear refinement on the acquired 3D lines with their track information. The straightforward approach is to perform geometric refinement on the reprojection error. With the 2D point-line association available, we can formulate a joint optimization problem by including additional structural information. The energy to minimize can be written as follows:

\begin{equation}
    E = \sum_pE_P(p) + \sum_lE_L(l) + \sum_{(p, l)}E_{PL}(p, l),
\end{equation}
where $E_P$ and $E_L$ are the data terms, and $E_{PL}$ encodes the 3D association between lines and points / VPs. In particular, $E_P$ is the 2D point reprojection error as in regular bundle adjustment \cite{schonberger2016structure}. The association energy is softly weighted (as discussed later) and optimized with robust Huber loss \cite{ceres}. Each line is converted into a 4-DoF infinite line with Pl\"ucker coordinate \cite{bartoli2005structure} for optimization and converted back to line segments by unprojecting its 2D supports. Each vanishing point is parameterized with a 3-dimensional homogeneous vector. Refer to Sec. A in supp. for details on efficient computation with minimal parameterization.

\noindent
\textbf{Geometric Refinement.}
The data term of each line track is also defined on its 2D reprojections. In particular, we measure the 2D perpendicular distance weighted by the angle consistency, which we robustly equip with Cauchy loss \cite{ceres}:
\begin{equation}
    E_L(l) = \sum_k w_{\angle}^2(L_k, \ell_k)~\cdot~e_{\text{perp}}^2(L_k, \ell_k) ,
\end{equation}
where $e_{\text{perp}}$ is the perpendicular distance, $L_k$ is the 2D projection of the 3D segment, $\ell_k$ are the 2D line segments, and $w_{\angle}$ is the exponential of one minus the cosine of the 2D angle between the projected and the observed line.

\noindent
\textbf{Soft Association between Lines and Points.}
For each pair of 3D line and 3D point with their track information, we can estimate how likely they are spatially associated by traversing the 2D association graph (described in Sec. \ref{sec::association_2d}) of their supports. Specifically, we count the number of associations among the 2D supports of the line track and point track, and keep pairs with at least three 2D associations. The 3D association energy $E_{PL}$, defined on the surviving pairs, is formulated as the 3D point-line distance weighted by the number of 2D associations on their supports.

\noindent
\textbf{Soft Association between Lines and VPs.}
Same as the point case, we can also build a soft association problem between lines and VPs. First, we acquire 3D VP tracks by transitively propagating line correspondences from the 3D line tracks. Then, we count the number of associations among the 2D supports for each pair of 3D line and VP track. The 3D line-VP association energy is defined as the sine of the direction angle between the 3D line and the VP, implicitly enforcing parallelism. Furthermore, we add regularizations to the nearly orthogonal VP pairs to enforce orthogonality of different line groups. Refer to Sec. C in supp. for details.

\section{Experiments}

\label{sec::exp}
\noindent
\textbf{Implementation Details. }
Our whole library is implemented in C++ with Python bindings~\cite{pybind11}. The triangulation and scoring can be run in parallel for each node, enabling scalability to large datasets. We use $n_v = 20$ visual neighbors and keep the top $K = 10$ line matches. We provide all values of thresholds and scaling factors in Sec. F.2 in supp.

\subsection{Line Mapping}

\begin{table}[tb]
\begin{center}
\scriptsize
\setlength{\tabcolsep}{3pt}
\begin{tabular}{clccccccc}
\toprule
Line type & Method & R1 & R5 & R10 & P1 & P5 & P10 & \# supports \\
\midrule
\multirow{3}{*}{\makecell{LSD\\\cite{von2008lsd}}} & L3D++ \cite{hofer2017efficient} & 37.0 & 153.1 & 218.8 & 53.1 & 80.8 & \textbf{90.6} & (14.8 / 16.8) \\
& ELSR \cite{wei2022elsr} & 13.9 & 59.7 & 96.5 & 55.4 & 72.6 & 82.2 & (N/A / N/A) \\
& Ours & \textbf{48.6} & \textbf{185.2} & \textbf{251.3} & \textbf{60.1} & \textbf{82.4} & 90.0 & (\textbf{16.4} / \textbf{20.5}) \\
\midrule
\multirow{2}{*}{\makecell{SOLD2\\ \cite{pautrat2021sold2}}} & L3D++ \cite{hofer2017efficient} & 36.9 & 107.5 & 132.8 & 67.2 & \textbf{86.8} & \textbf{93.2} & (13.2 / 20.4) \\
& Ours & \textbf{54.3} & \textbf{151.1} & \textbf{191.2} & \textbf{69.8} & 84.6 & 90.0 & (\textbf{16.5} / \textbf{38.7}) \\
\bottomrule
\end{tabular}
\caption{\textbf{Line reconstruction on Hypersim \cite{roberts:2021}} with LSD \cite{von2008lsd} and SOLD2 \cite{pautrat2021sold2} lines. $R\tau$ and $P\tau$ are reported at 1mm, 5mm, 10 mm along with the average number of supporting images/lines.}
\label{tab::main-hypersim}
\end{center}
\end{table}

\begin{table}[tb]
\begin{center}
\scriptsize
\setlength{\tabcolsep}{3pt}
\begin{tabular}{lccccccc}
\toprule
Method & R5 & R10 & R50 & P5 & P10 & P50 & \# supports \\
\midrule
L3D++ \cite{hofer2017efficient} & 373.7 & 831.6 & 2783.6 & 40.6 & 54.5 & 85.9 & (8.8 / 9.3) \\
ELSR \cite{wei2022elsr} & 139.2 & 322.5 & 1308.0 & 38.5 & 48.0 & 74.5 & (N/A / N/A) \\
Ours (line-only) & 472.1 & 1058.8 & 3720.7 & \textbf{46.8} & \textbf{58.4} & \textbf{86.1} & (10.3 / 11.8) \\
Ours & \textbf{508.3} & \textbf{1154.5} & \textbf{4179.5} & 46.0 & 56.9 & 83.7 & (\textbf{10.4} / \textbf{12.0}) \\
\bottomrule
\end{tabular}
\caption{\textbf{Line reconstruction on \textit{train} split of \textit{Tanks and Temples} \cite{Knapitsch2017}} with LSD \cite{von2008lsd} lines. $R\tau$ and $P\tau$ are reported at 5mm, 10mm, 50mm along with the average number of supporting images/lines.}
\label{tab::main-tnt}
\end{center}
\end{table}

To validate the effectiveness of our system, we set up an evaluation benchmark to quantify the quality of the reconstructed 3D line maps. As there are no ground truth (GT) 3D lines, we evaluate the 3D line mapping with either GT mesh models or point clouds. We use the following metrics:
\begin{itemize}[noitemsep,nolistsep]
    \item \textit{Length recall} (in meters) at $\tau$ ($R
\tau$): sum of the lengths of the line portions within $\tau$ mm from the GT model.
    \item \textit{Inlier percentage} at $\tau$ ($P
\tau$): the percentage of tracks that are within $\tau$ mm from the GT model.
    \item \textit{Average supports}: average number of image supports and 2D line supports across all line tracks.
\end{itemize}

In the following, we compare our system with two state-of-the-art methods as baselines: L3D++ \cite{hofer2017efficient} and ELSR \cite{wei2022elsr}, using two line detectors: the traditional LSD detector \cite{von2008lsd} and the learning-based SOLD2 \cite{pautrat2021sold2}. For ELSR \cite{wei2022elsr}, we convert the input into VisualSfM \cite{wu2011visualsfm} format and use code\footnote{\texttt{https://skyearth.org/publication/project/ELSR/}} from the authors (only supporting LSD \cite{von2008lsd}).

Our first evaluation is run on the first eight scenes of the Hypersim dataset \cite{roberts:2021}, composed of 100 images each, and is reported in \cref{tab::main-hypersim}.
For both detectors, we reconstruct much more complete line maps with better or comparable precision than the competitors, while also exhibiting significantly higher quality of track information. This abundant track association is beneficial particularly for line-based applications such as visual localization \cite{gao2022pose}.
After discussing with the authors of ELSR, it seems that their method does not achieve satisfactory results due to a lack of point and plane features.

\begin{figure}[tb]
\scriptsize
\setlength\tabcolsep{2pt} 
\begin{tabular}{cccccccc}
{\includegraphics[trim={0 0 0 0}, clip, width=0.14\linewidth, height=10pt]{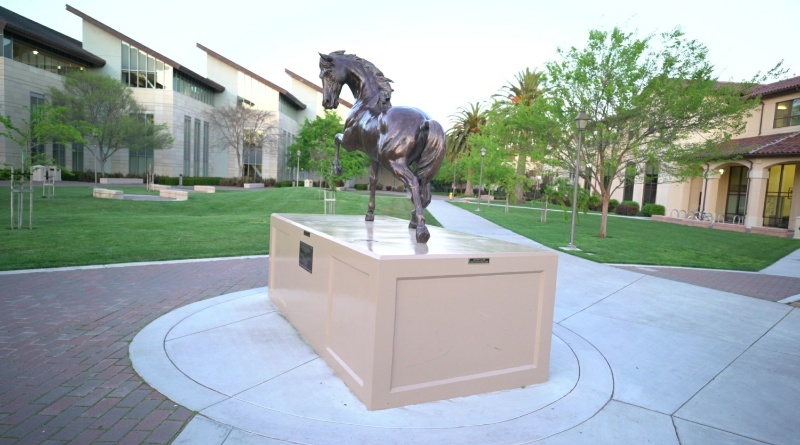}} &
{\includegraphics[trim={0 0 0 0}, clip, width=0.14\linewidth, height=10pt]{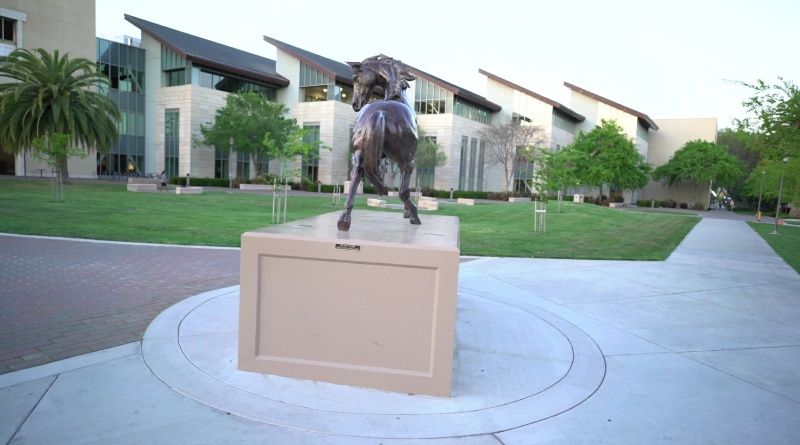}} &
{\includegraphics[trim={0 0 0 0}, clip, width=0.14\linewidth, height=10pt]{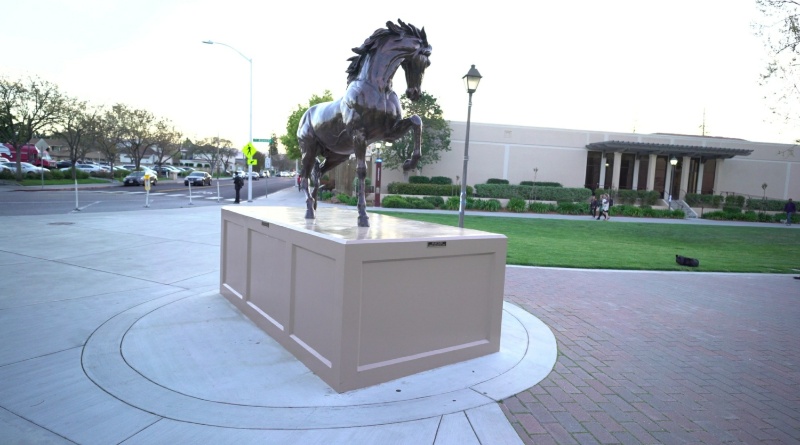}} &
{\includegraphics[trim={0 0 0 0}, clip, width=0.14\linewidth, height=10pt]{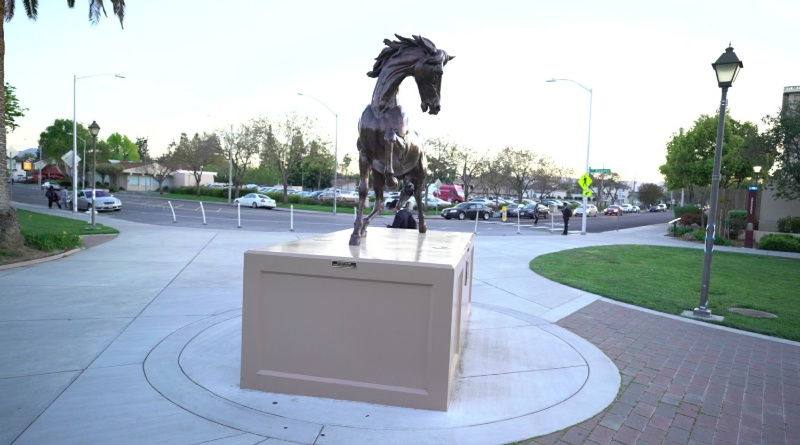}} & & &
{\includegraphics[trim={0 0 0 0}, clip, width=0.14\linewidth, height=10pt]{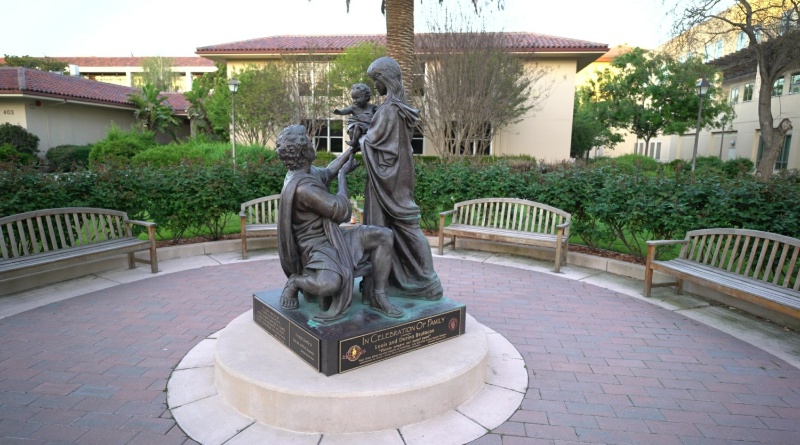}} & 
{\includegraphics[trim={0 0 0 0}, clip, width=0.14\linewidth, height=10pt]{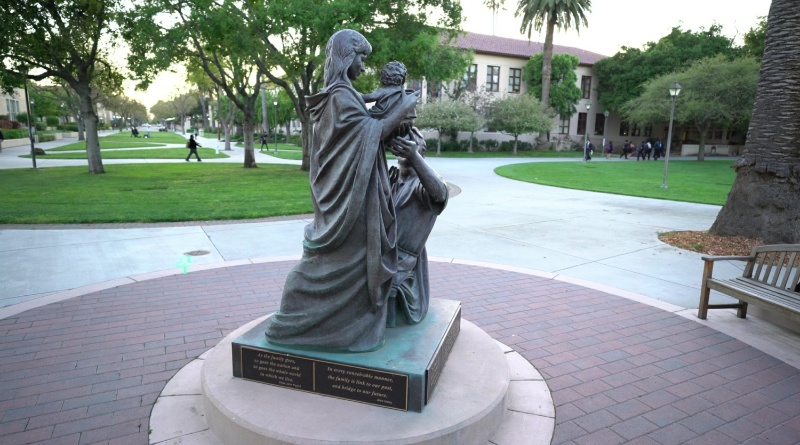}} \\
\end{tabular}
\begin{tabular}{ccc}
{\includegraphics[trim={0 0 0 100}, clip, width=0.34\linewidth, height=53pt]{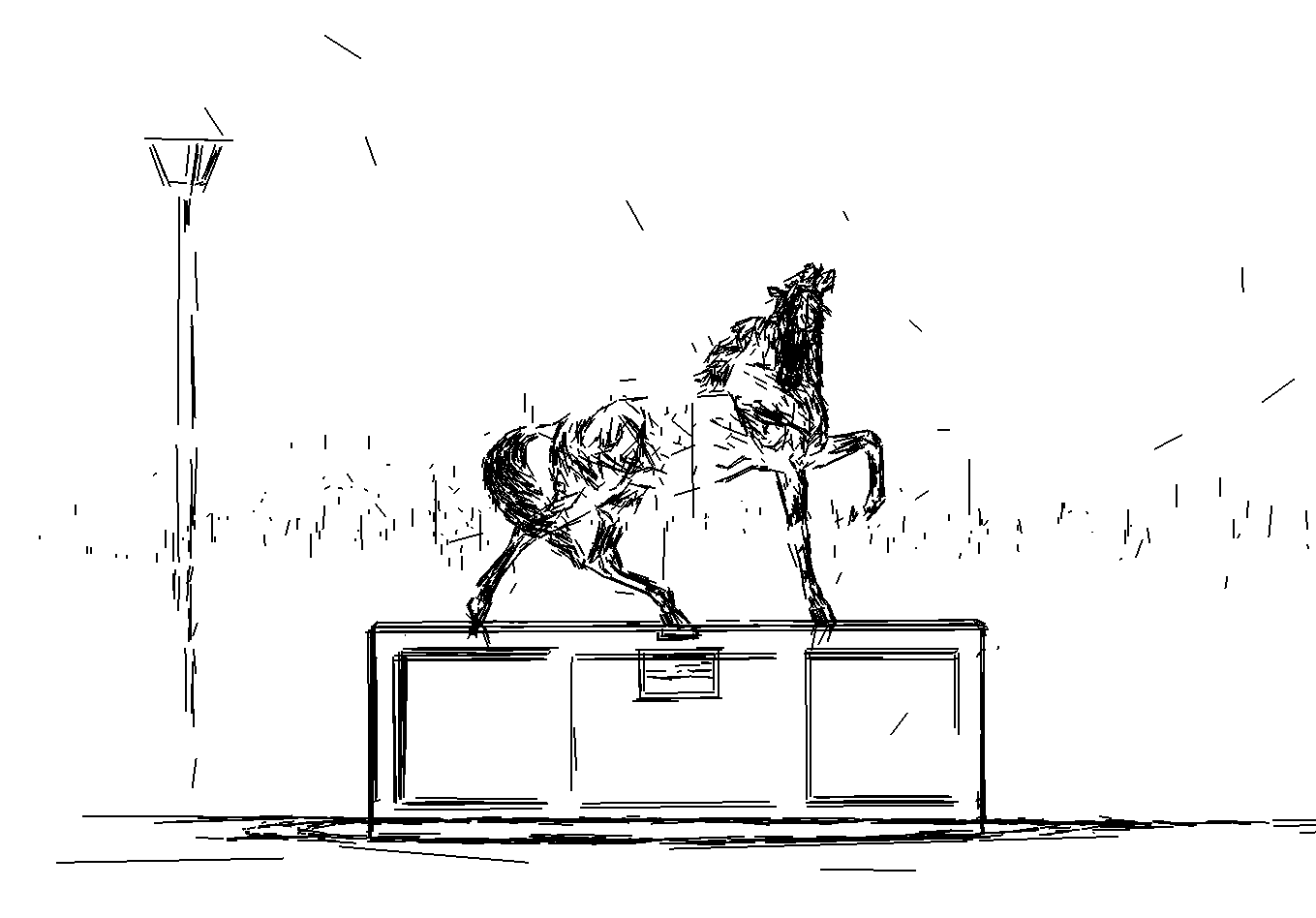}} & 
{\includegraphics[trim={100 0 200 0}, clip, width=0.26\linewidth, height=53pt]{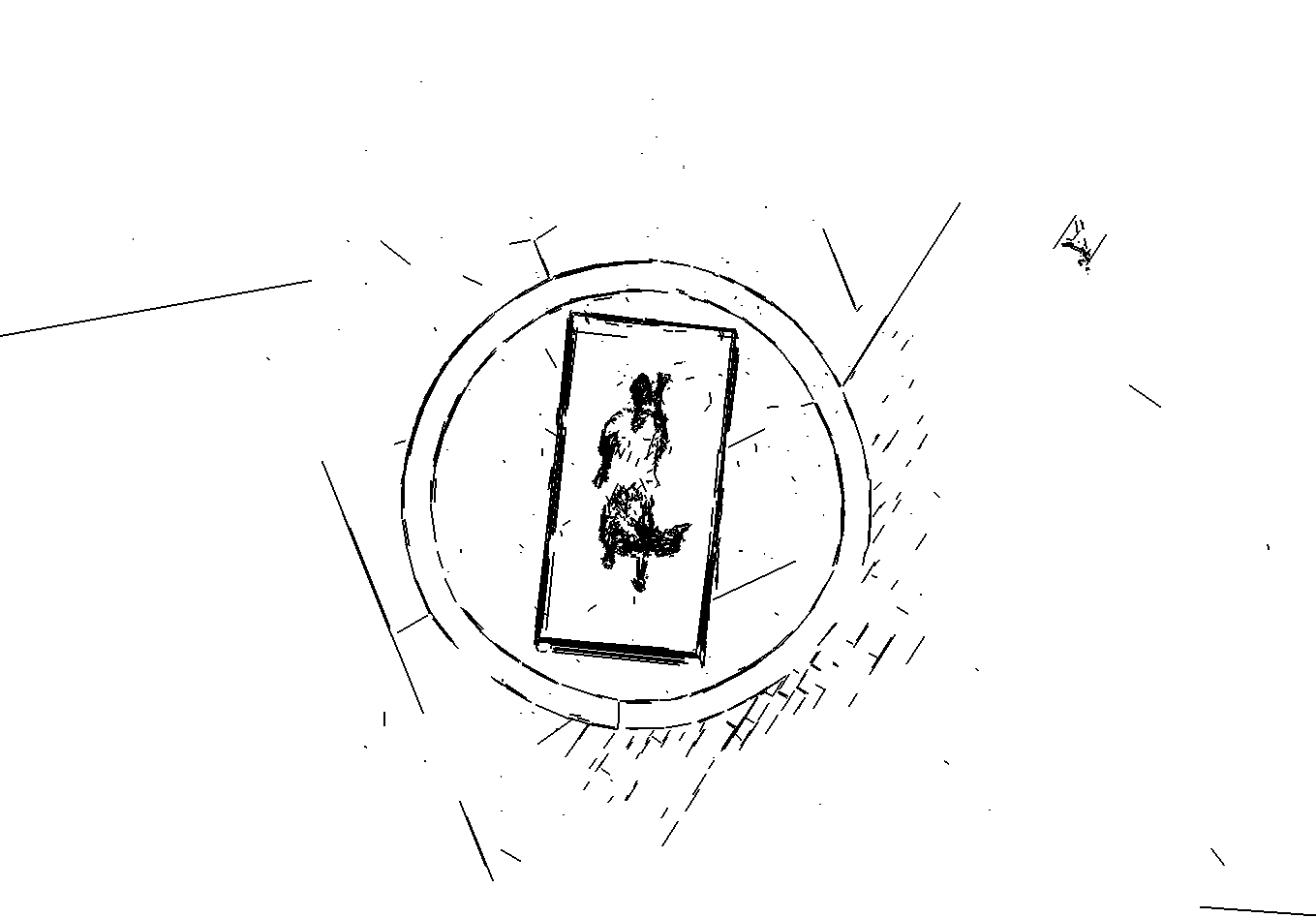}} &
{\includegraphics[trim={0 0 0 40}, clip, width=0.35\linewidth, height=53pt]{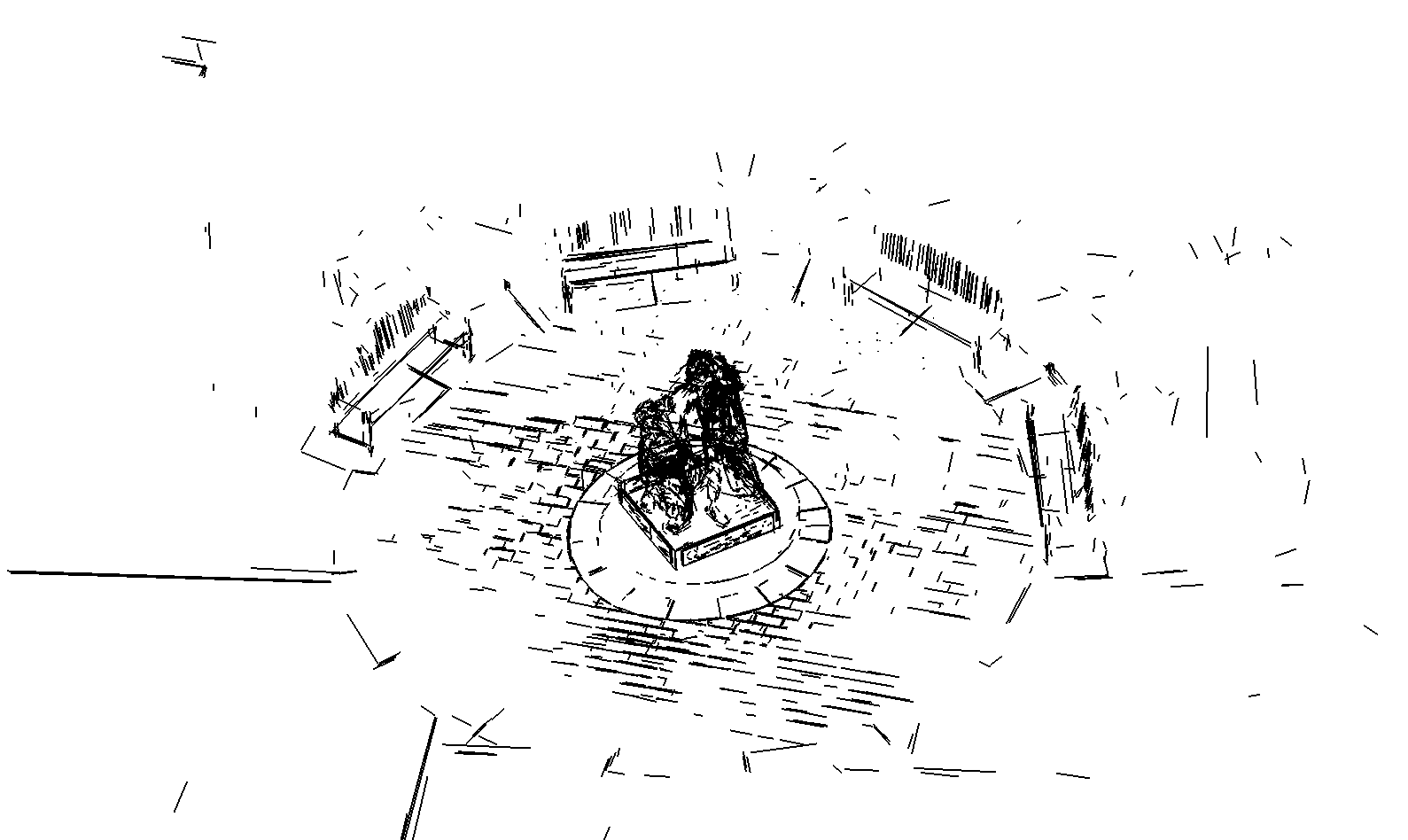}} \\
{\includegraphics[trim={0 0 0 100}, clip, width=0.34\linewidth, height=53pt]{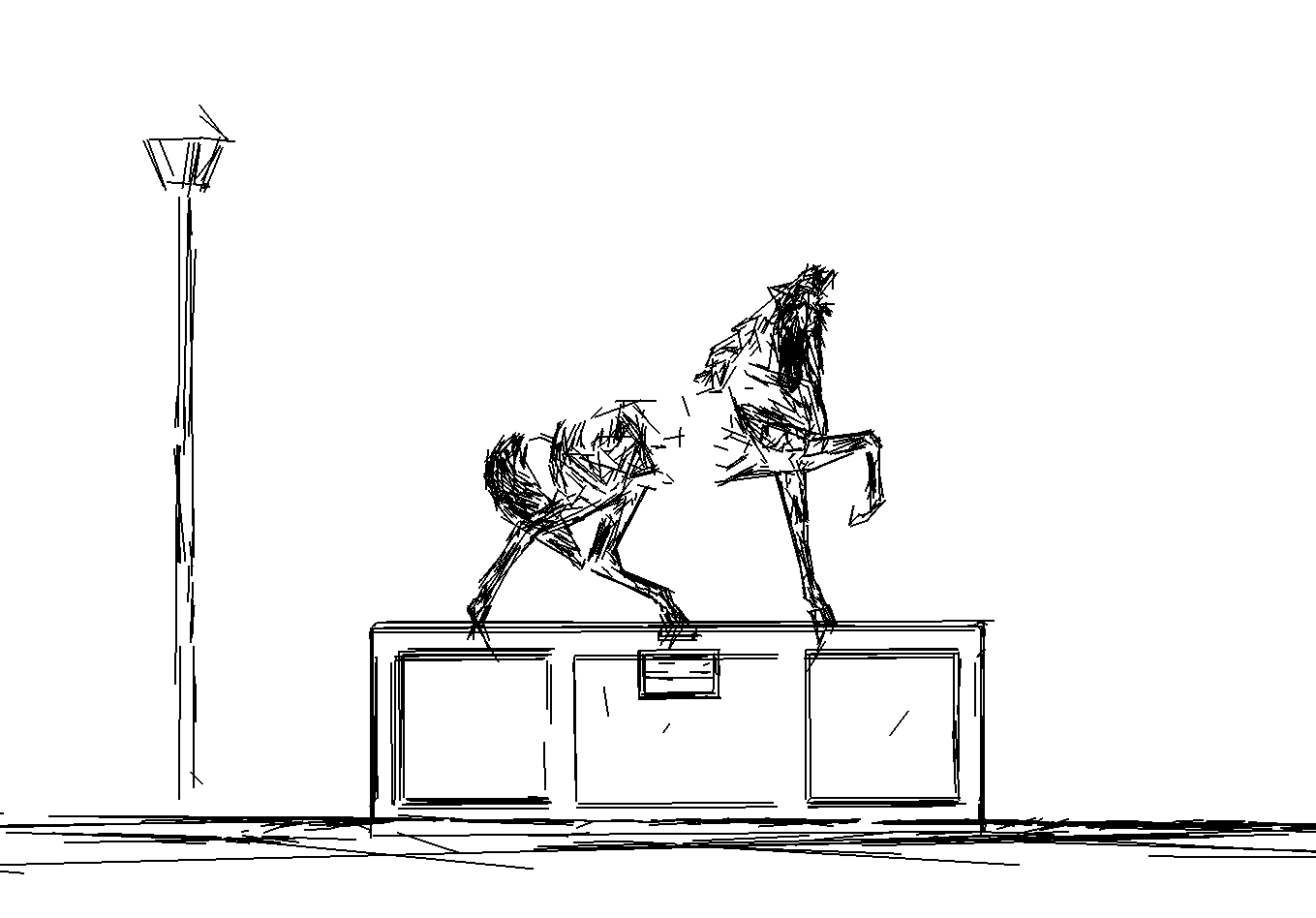}} & 
{\includegraphics[trim={100 0 200 0}, clip, width=0.26\linewidth, height=53pt]{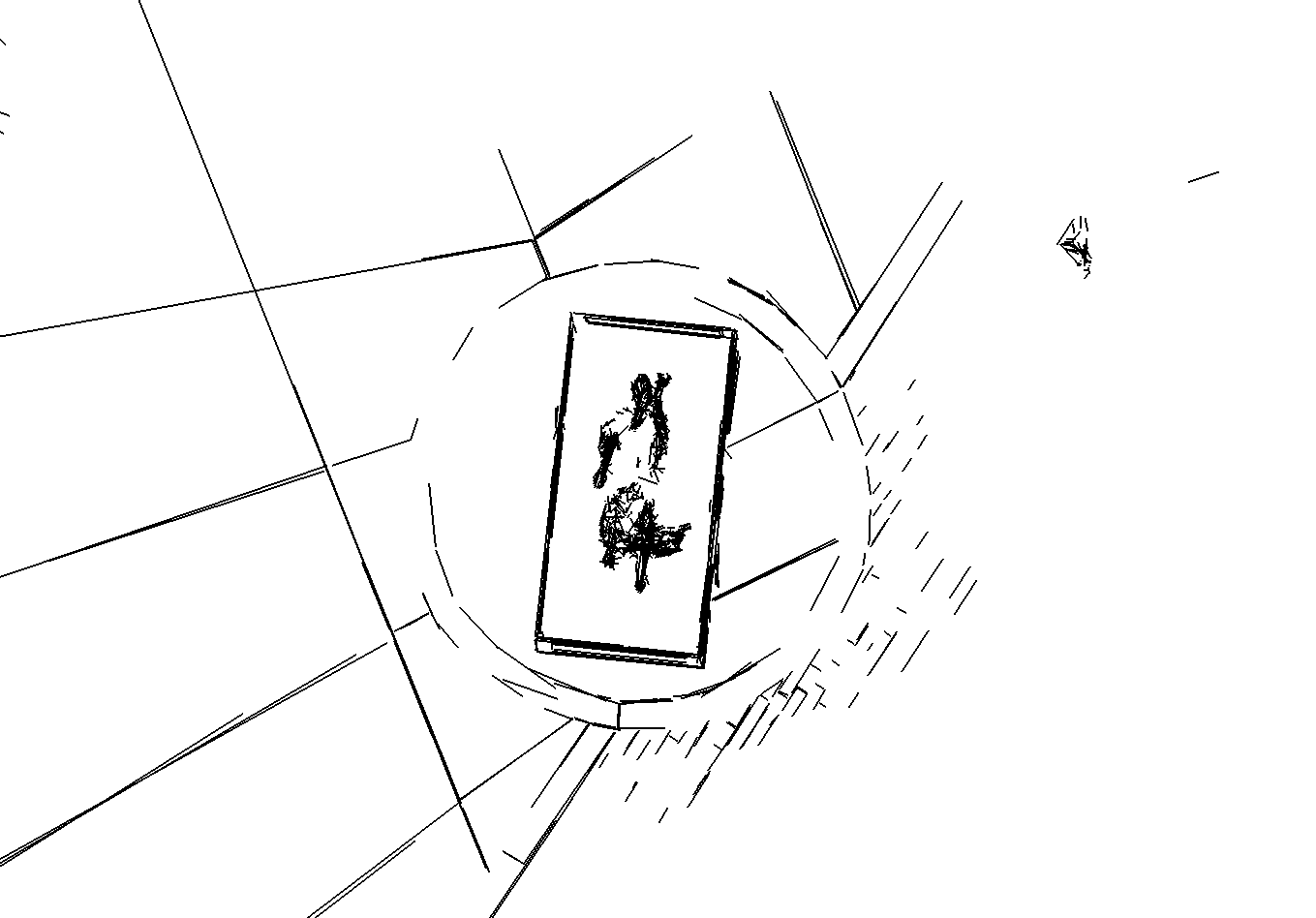}} &
{\includegraphics[trim={0 0 0 40}, clip, width=0.35\linewidth, height=53pt]{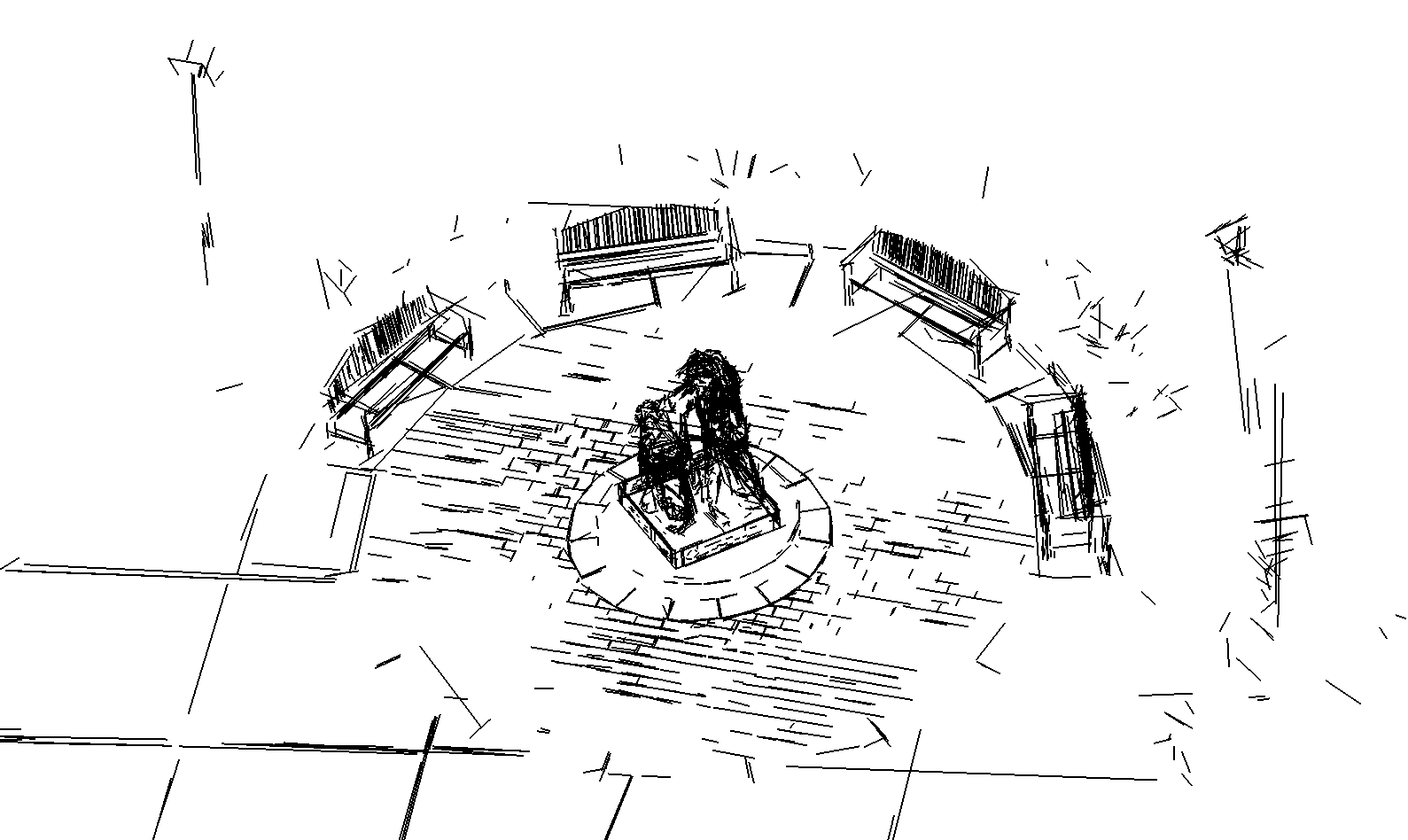}}
\\
\end{tabular}
\centering
\caption{\textbf{Top row:} L3D++ \cite{hofer2017efficient}. \textbf{Bottom row:} Ours. Both systems are run on \textit{Horse} and \textit{Family} from \cite{Knapitsch2017}. We show two different views on the main scene of \textit{Horse}.}
\label{fig::tnt_comparison}
\end{figure}

\begin{figure}[tb]
\scriptsize
\setlength\tabcolsep{2pt} 
\begin{tabular}{cccccc}
{\includegraphics[width=0.15\linewidth, height=10pt]{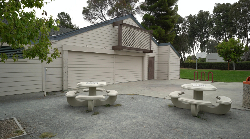}} &
{\includegraphics[width=0.15\linewidth, height=10pt]{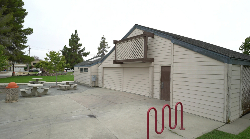}} &
{\includegraphics[width=0.15\linewidth, height=10pt]{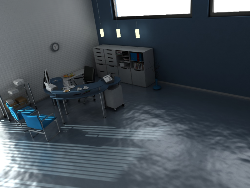}} &
{\includegraphics[width=0.15\linewidth, height=10pt]{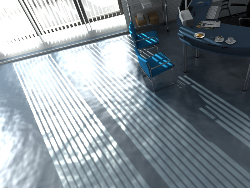}} &
{\includegraphics[width=0.15\linewidth, height=10pt]{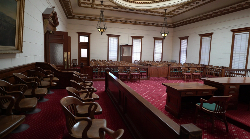}} &
{\includegraphics[width=0.15\linewidth, height=10pt]{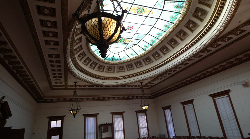}}
\\
\multicolumn{2}{c}{\includegraphics[trim={0 0 0 0}, clip, width=0.31\linewidth, height=50pt]{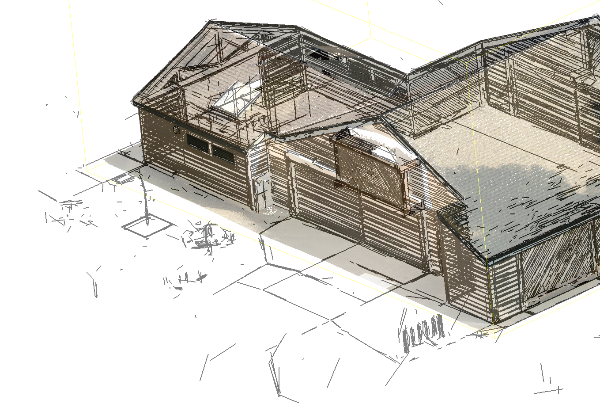}} & 
\multicolumn{2}{c}{\includegraphics[trim={30 30 30 30}, clip, width=0.31\linewidth, height=50pt]{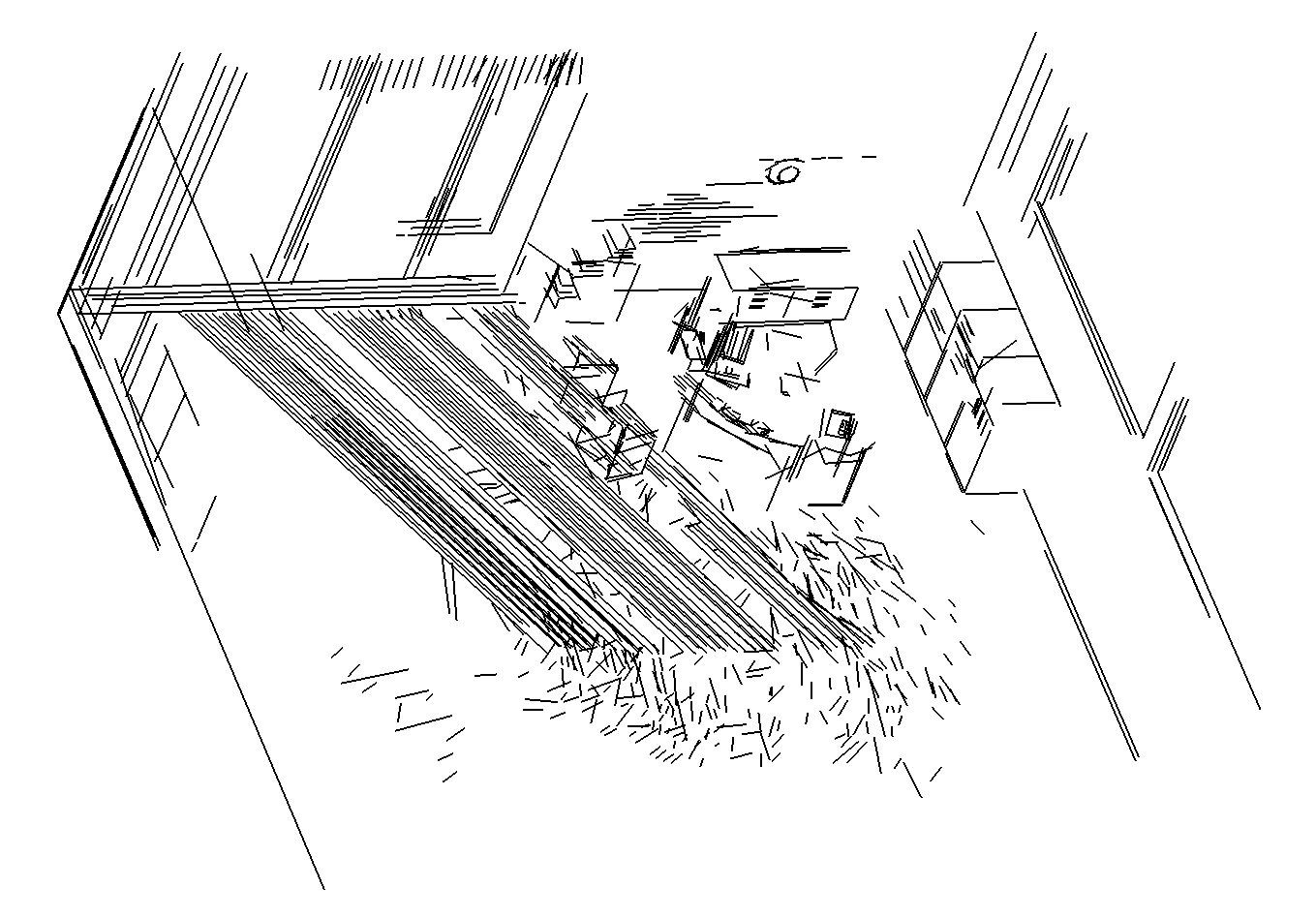}} &
\multicolumn{2}{c}{\includegraphics[trim={80 60 80 100}, clip, width=0.31\linewidth, height=50pt]{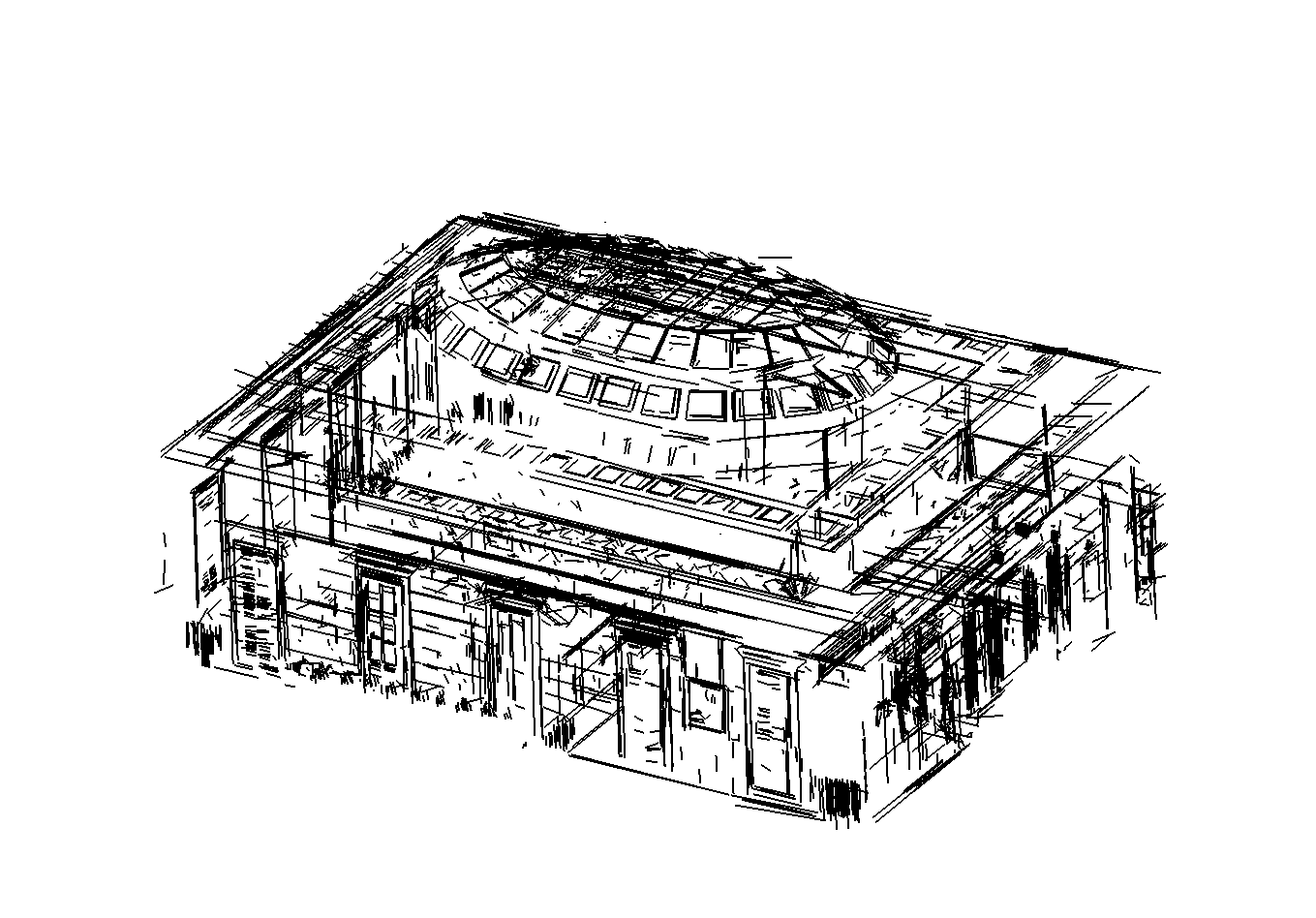}}
\\
\multicolumn{2}{c}{\textit{Barn} (410 images)} &
\multicolumn{2}{c}{\textit{ai\_001\_003} (100 images)} &
\multicolumn{2}{c}{\textit{Courtroom} (301 images)} \\
\end{tabular}
\centering
\caption{\textbf{Qualitative results on Hypersim \cite{roberts:2021} and \textit{Tanks and Temples} \cite{Knapitsch2017}.} On \textit{Barn} we jointly visualize our results and the aligned ground truth point cloud.}
\label{fig::qualitative}
\end{figure}

We further evaluate all three methods on the \textit{train} split of the \textit{Tanks and Temples} dataset~\cite{Knapitsch2017} without \textit{Ignatius} as it has no line structures. As SOLD2 \cite{pautrat2021sold2} is trained for indoor images, we only use LSD \cite{von2008lsd}. Since the provided point cloud was cleaned to focus only on the main subject, we compute its bounding box, extend it by one meter, and only evaluate lines inside this region. This prevents incorrectly penalizing correct lines that are far away from the main scene, which our method is particularly good at thanks to our scale-invariant design (refer to Sec. G in supp.). \cref{tab::main-tnt} shows the results, where our methods significantly improve the mapping quality across the board. \cref{fig::tnt_comparison} shows qualitative comparison between our method and L3D++ \cite{hofer2017efficient}. Our results exhibit better completeness, have less noisy lines that are flying around, and achieve significantly more robust reconstructions of subtle details (e.g. on the ground). More examples of our produced line maps are shown in \cref{fig::qualitative}.

\begin{figure}[tb]
\scriptsize
\setlength\tabcolsep{2pt} 
\begin{tabular}{cccc}
{\includegraphics[trim={480 150 450 200}, clip, width=0.23\linewidth, height=50pt]{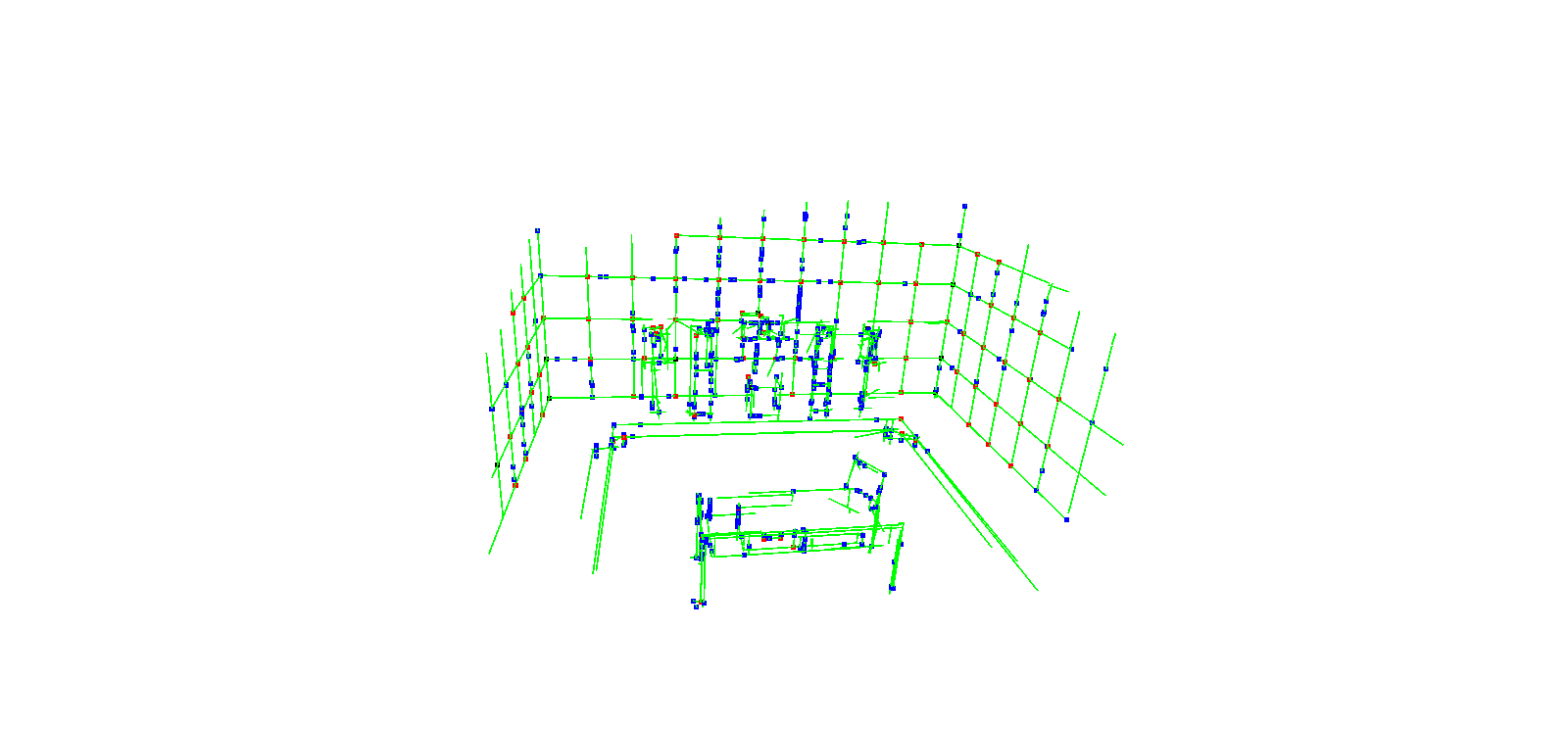}} & 
{\includegraphics[trim={440 120 420 100}, clip, width=0.23\linewidth, height=50pt]{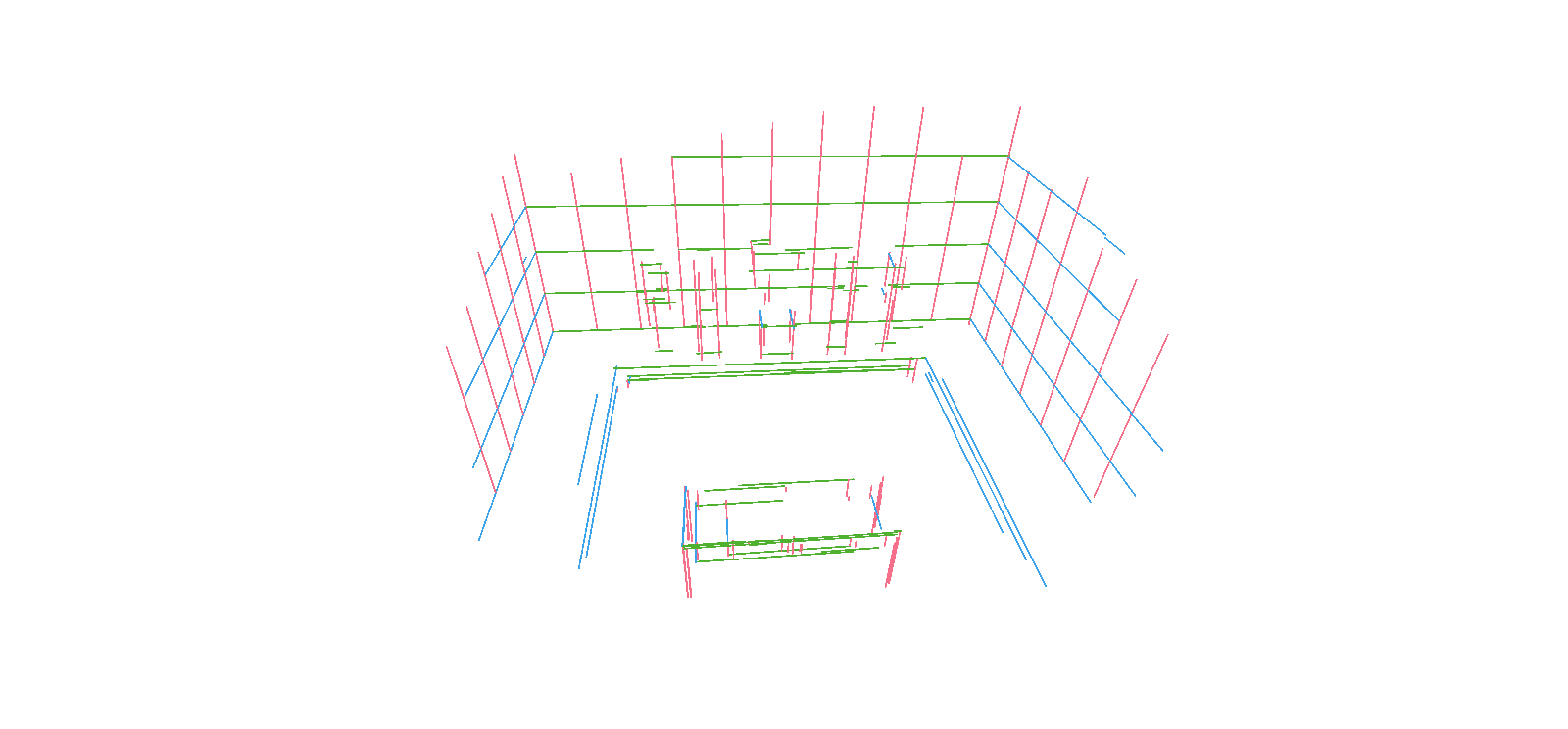}} &
{\includegraphics[trim={500 100 550 120}, clip, width=0.23\linewidth, height=50pt]{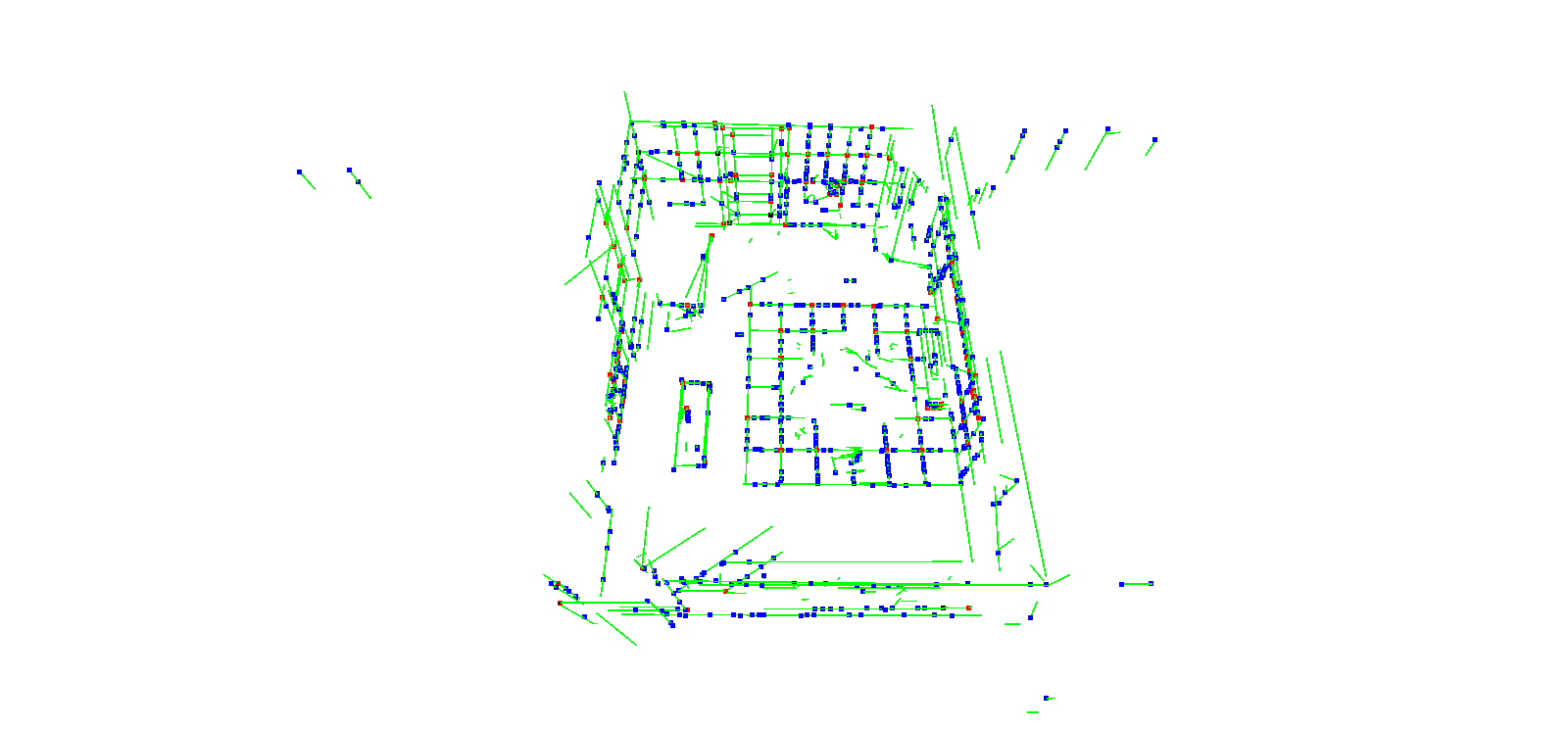}} &
{\includegraphics[trim={640 150 500 200}, clip, width=0.23\linewidth, height=50pt]{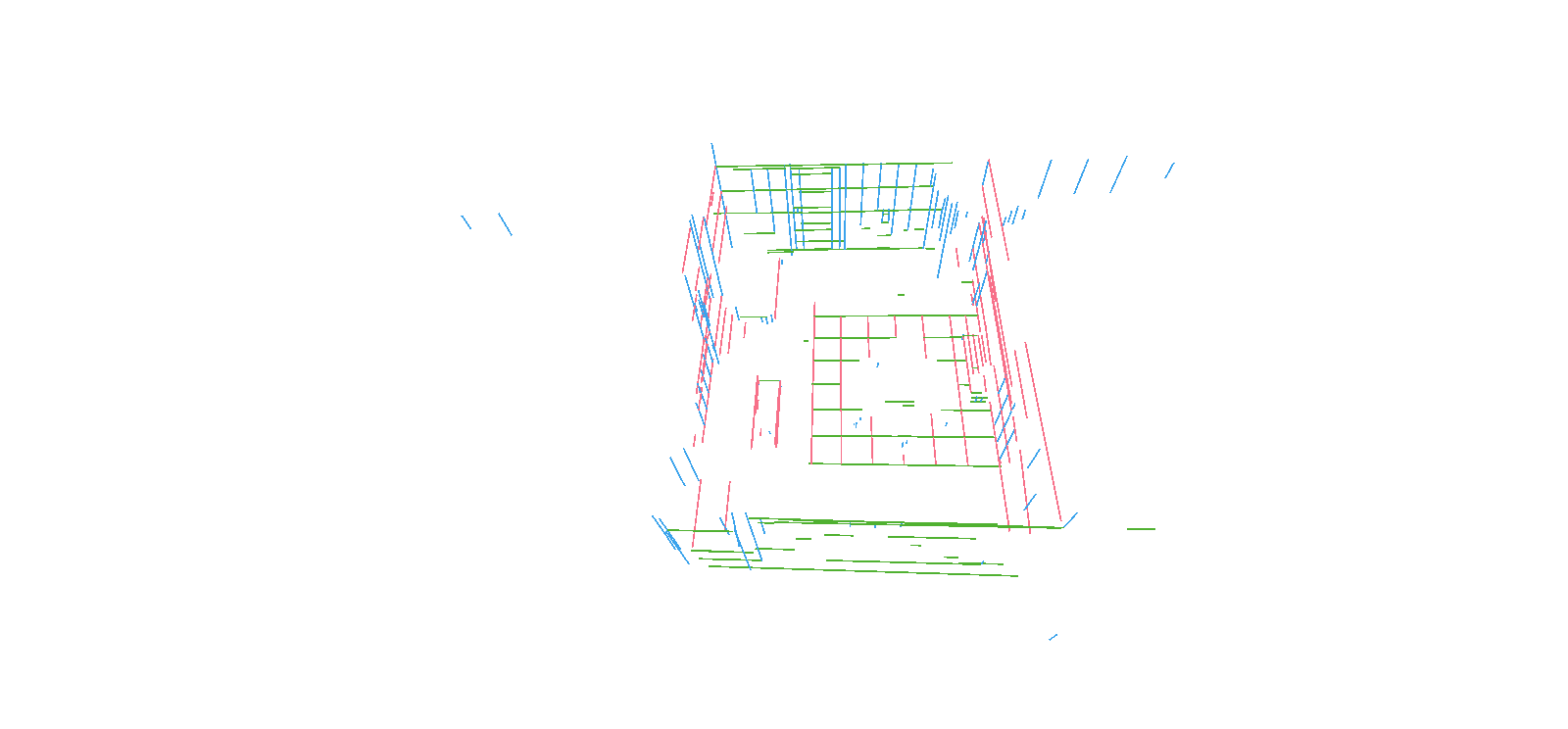}}
\end{tabular}
\centering
\caption{\textbf{Qualitative results} of the recovered line-point and line-VP association graphs (visualized similarly as in \cref{fig::teaser}). }
\label{fig::qualitative_structures}
\end{figure}

As an additional output of our system, junction structures and line-line relations such as parallelism and orthogonality are discovered, as shown in \cref{fig::qualitative_structures}. This directly comes from the line-point and line-VP soft associations of \cref{sec:refinement}. From the recovered structures, we can clearly perceive the scene and easily recognize the main Manhattan directions \cite{coughlan2000manhattan}.

\begin{figure*}[tb]
\centering
\begin{minipage}{.38\linewidth}
{\includegraphics[trim={200 20 150 100}, clip, width=0.98\linewidth, height=64pt]{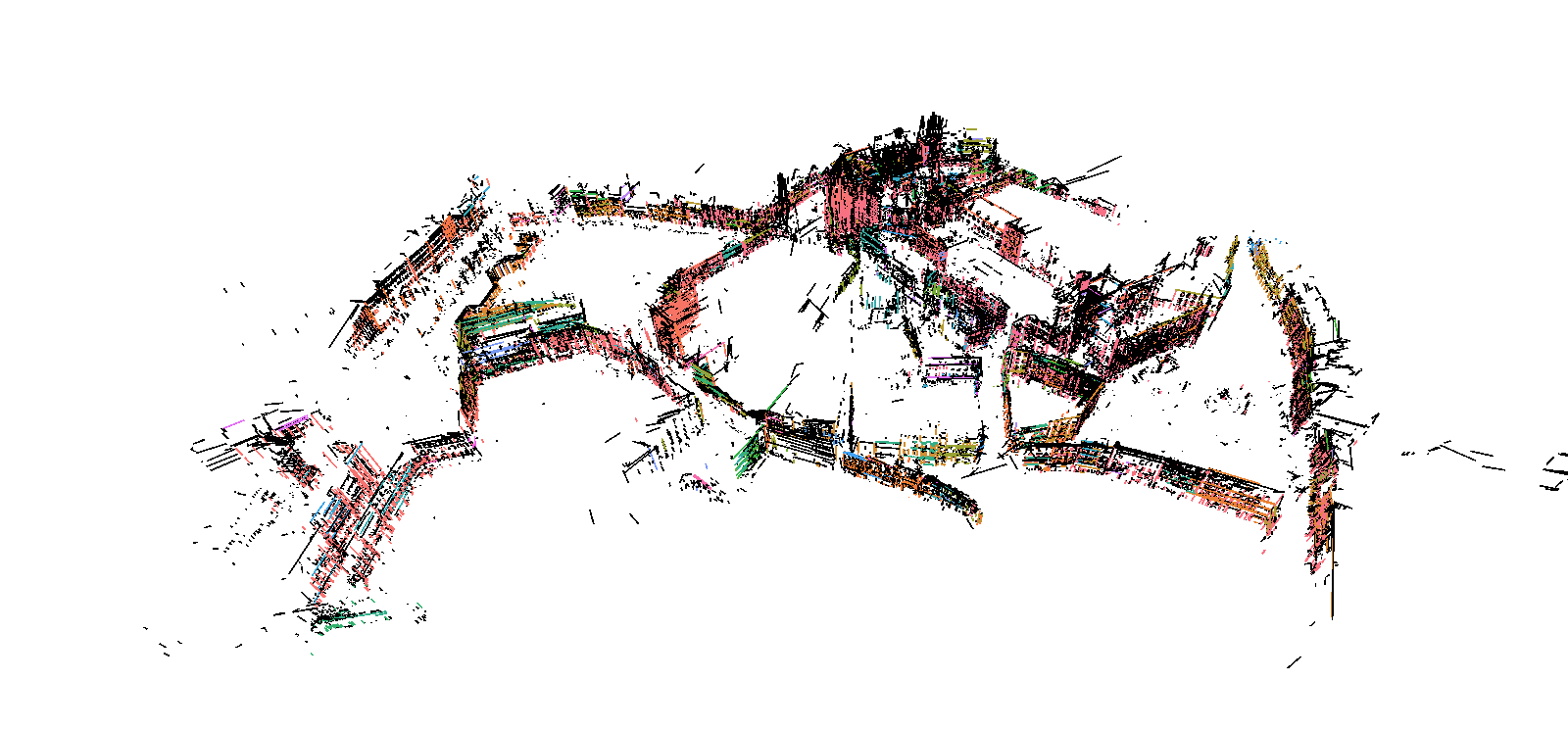}}
\end{minipage}
\begin{minipage}{.58\linewidth}
\setlength\tabcolsep{2pt} 
\begin{tabular}{ccccc}
\begin{minipage}{.19\linewidth}
\renewcommand{\arraystretch}{0.4}
\begin{tabular}{cc}
\includegraphics[width=0.24\linewidth, height=12pt]{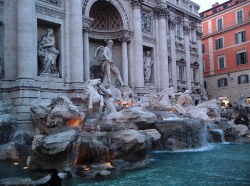} &
\raisebox{-0.55\height}[0pt][0pt]{\includegraphics[width=0.7\linewidth]{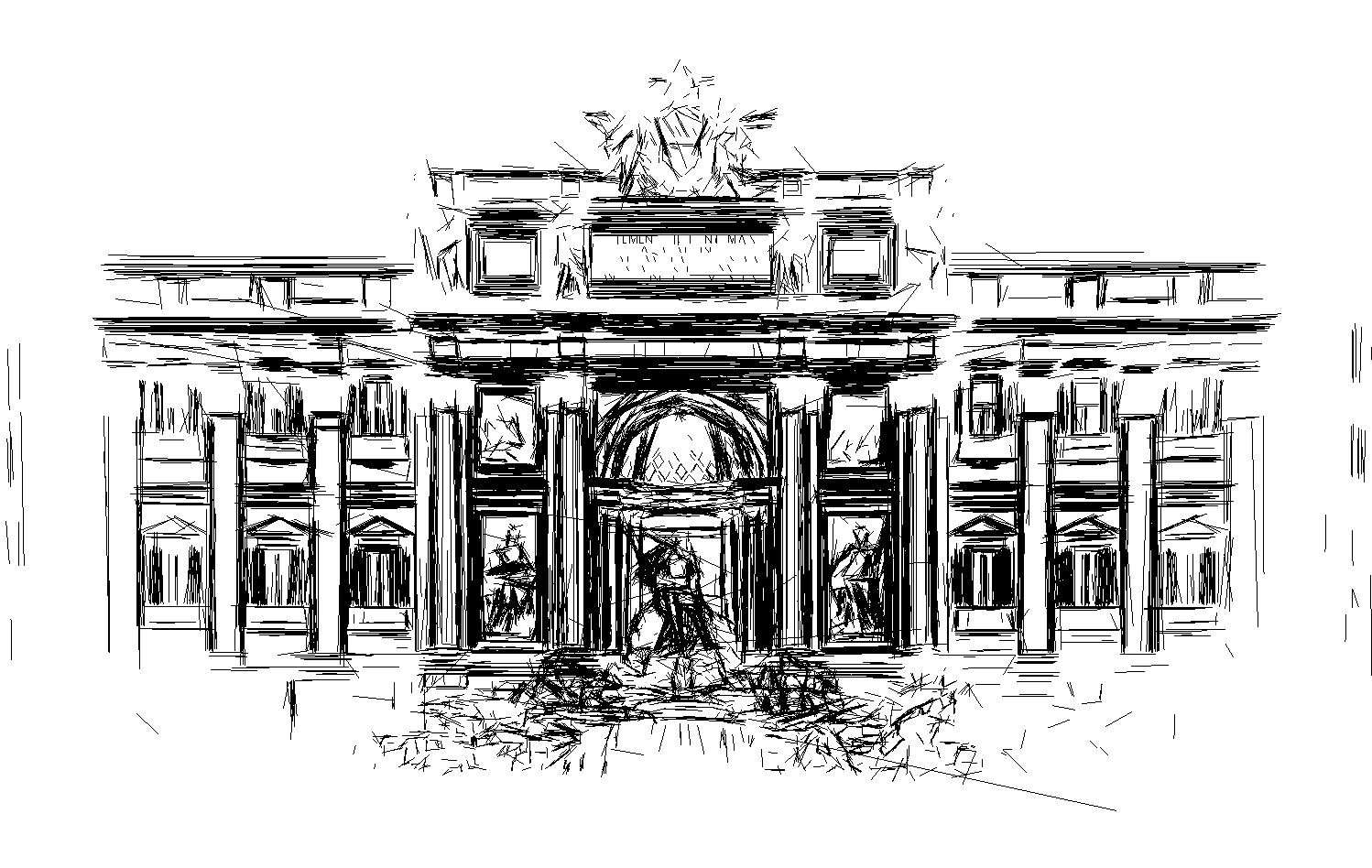}} \\
\includegraphics[width=0.24\linewidth, height=12pt]{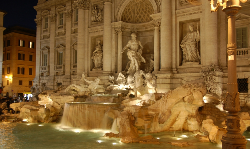} &
\end{tabular}
\end{minipage} 
&
\begin{minipage}{.19\linewidth}
\renewcommand{\arraystretch}{0.4}
\begin{tabular}{cc}
\includegraphics[width=0.24\linewidth, height=12pt]{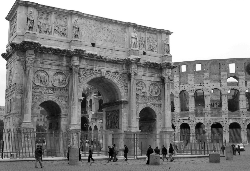} &
\raisebox{-0.55\height}[0pt][0pt]{{\includegraphics[width=0.7\linewidth]{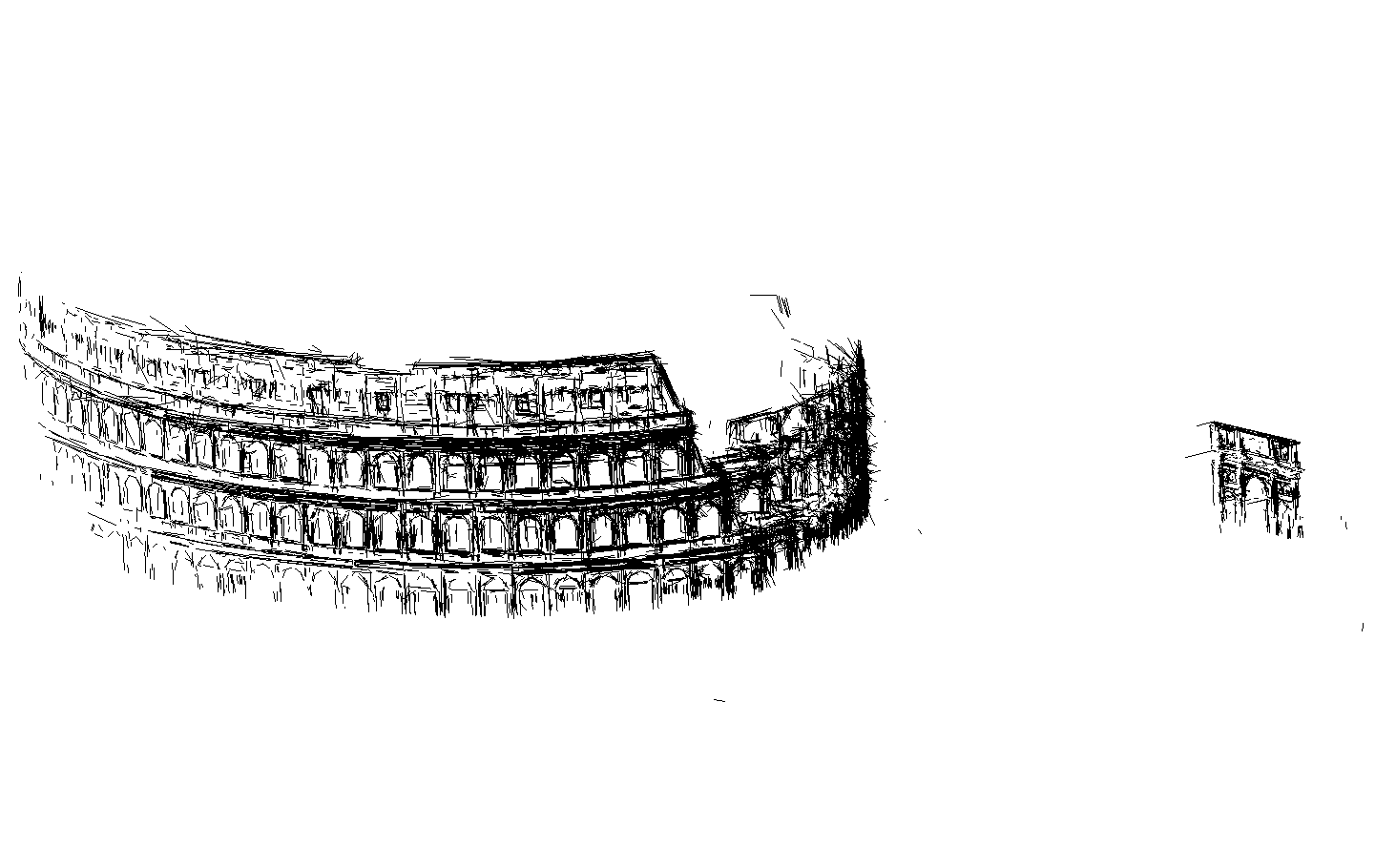}}} \\
\includegraphics[width=0.24\linewidth, height=12pt]{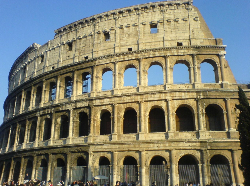} \\
\end{tabular}
\end{minipage} 
&
\begin{minipage}{.19\linewidth}
\renewcommand{\arraystretch}{0.4}
\begin{tabular}{cc}
\includegraphics[width=0.24\linewidth, height=12pt]{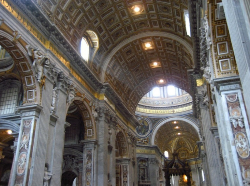} &
\raisebox{-0.55\height}[0pt][0pt]{{\includegraphics[trim={30 30 30 30}, clip, width=0.7\linewidth]{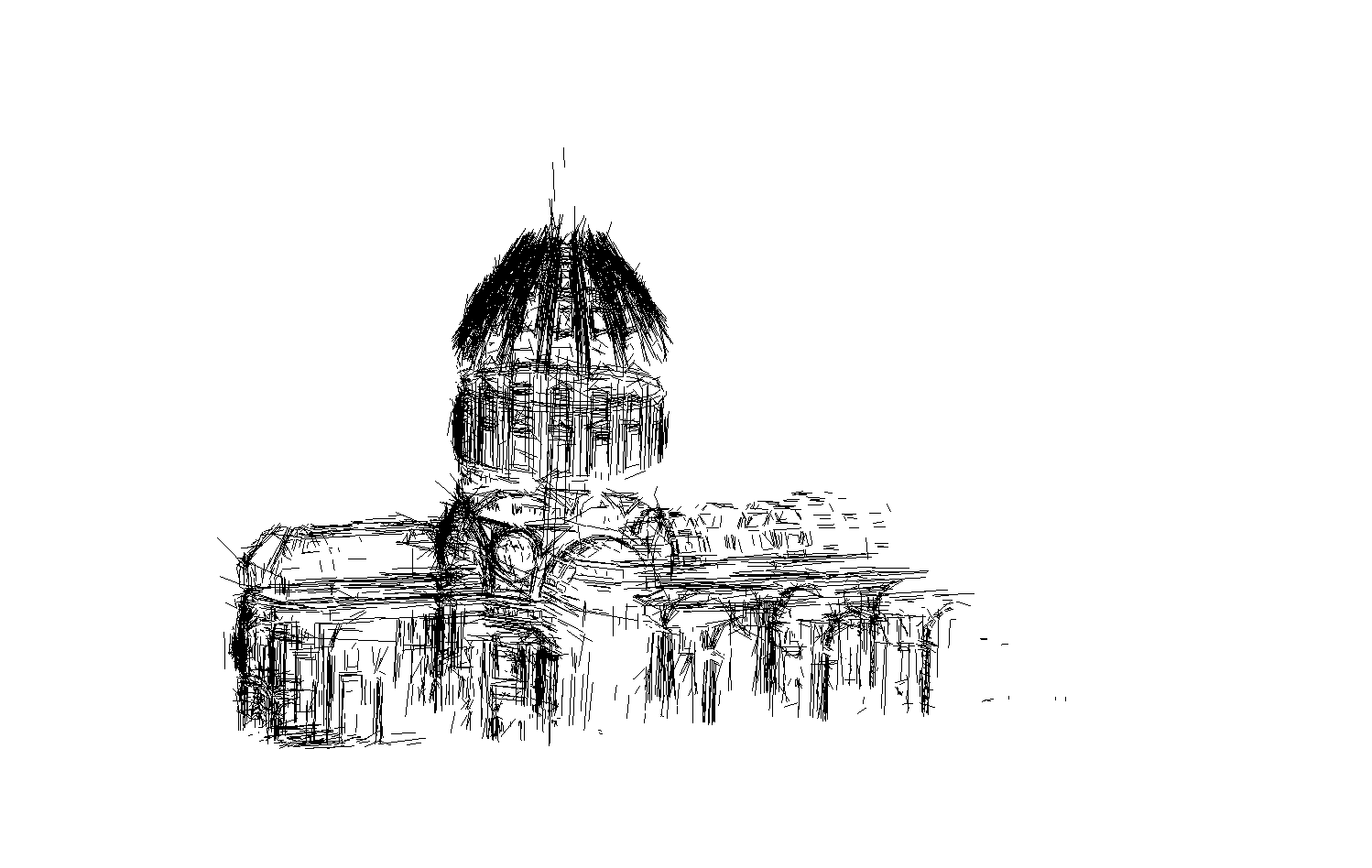}}} \\
\includegraphics[width=0.24\linewidth, height=12pt]{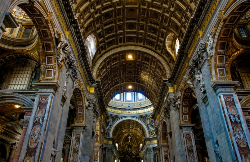} 
\end{tabular}
\end{minipage} 
&
\begin{minipage}{.19\linewidth}
\renewcommand{\arraystretch}{0.4}
\begin{tabular}{cc}
\includegraphics[width=0.24\linewidth, height=12pt]{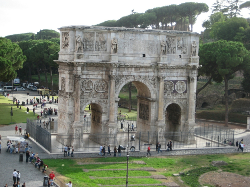} &
\raisebox{-0.55\height}[0pt][0pt]{{\includegraphics[width=0.7\linewidth]{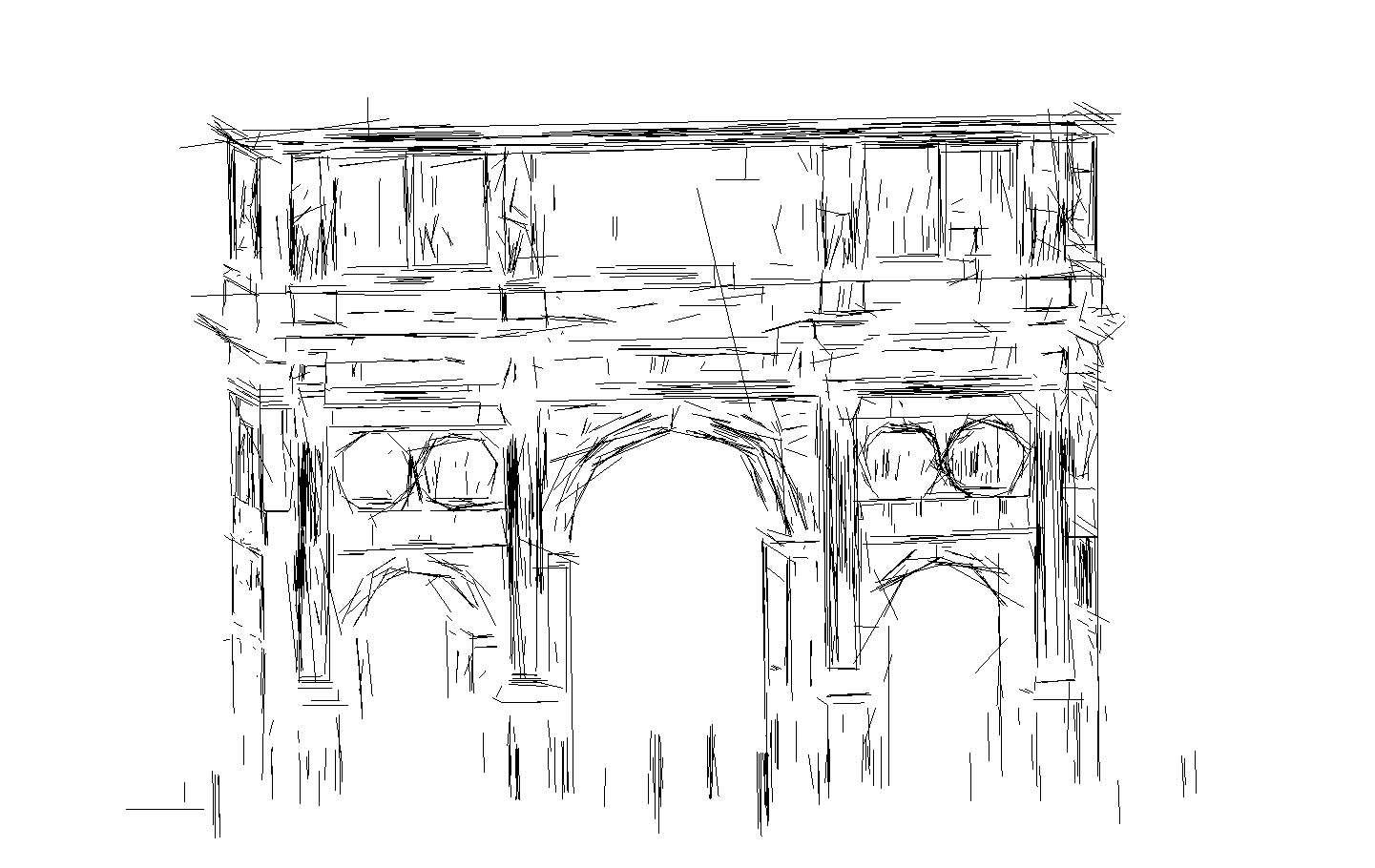}}} \\
\includegraphics[width=0.24\linewidth, height=12pt]{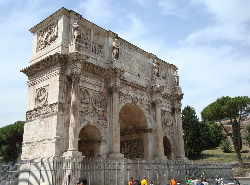} &
\end{tabular}
\end{minipage} 
&
\begin{minipage}{.19\linewidth}
\renewcommand{\arraystretch}{0.4}
\begin{tabular}{cc}
\includegraphics[width=0.24\linewidth, height=12pt]{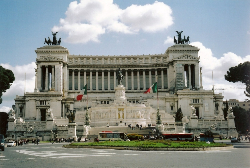} &
\raisebox{-0.55\height}[0pt][0pt]{{\includegraphics[width=0.7\linewidth]{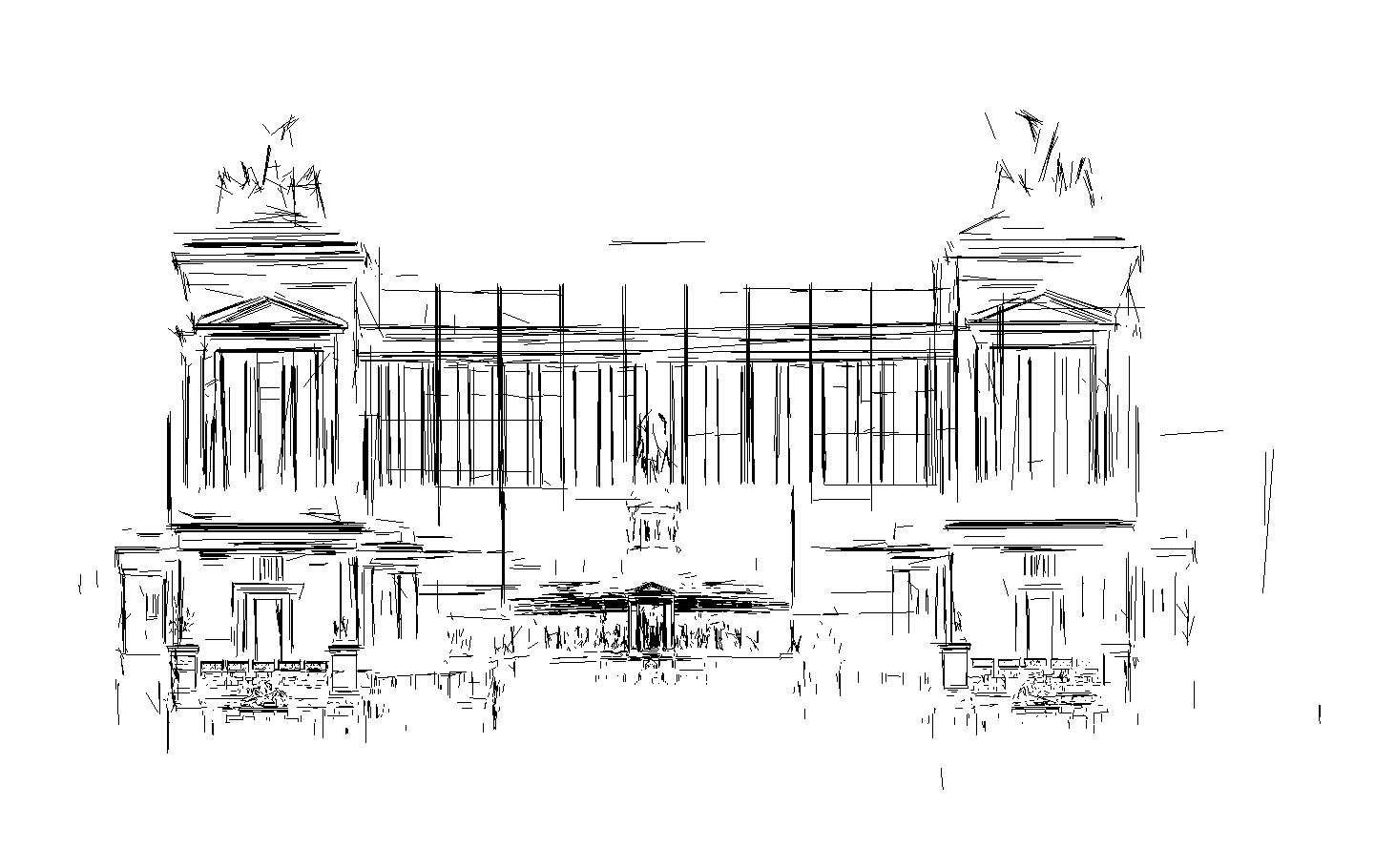}}} \\
\includegraphics[width=0.24\linewidth, height=12pt]{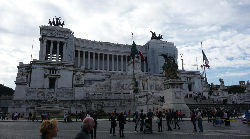} &
\end{tabular}
\end{minipage} 

\\
\noalign{\vskip 1mm}   
\begin{minipage}{.19\linewidth}
\renewcommand{\arraystretch}{0.4}
\begin{tabular}{cc}
\includegraphics[width=0.24\linewidth, height=12pt]{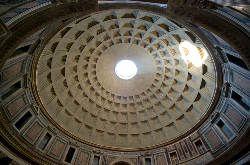} &
\raisebox{-0.55\height}[0pt][0pt]{{\includegraphics[width=0.7\linewidth]{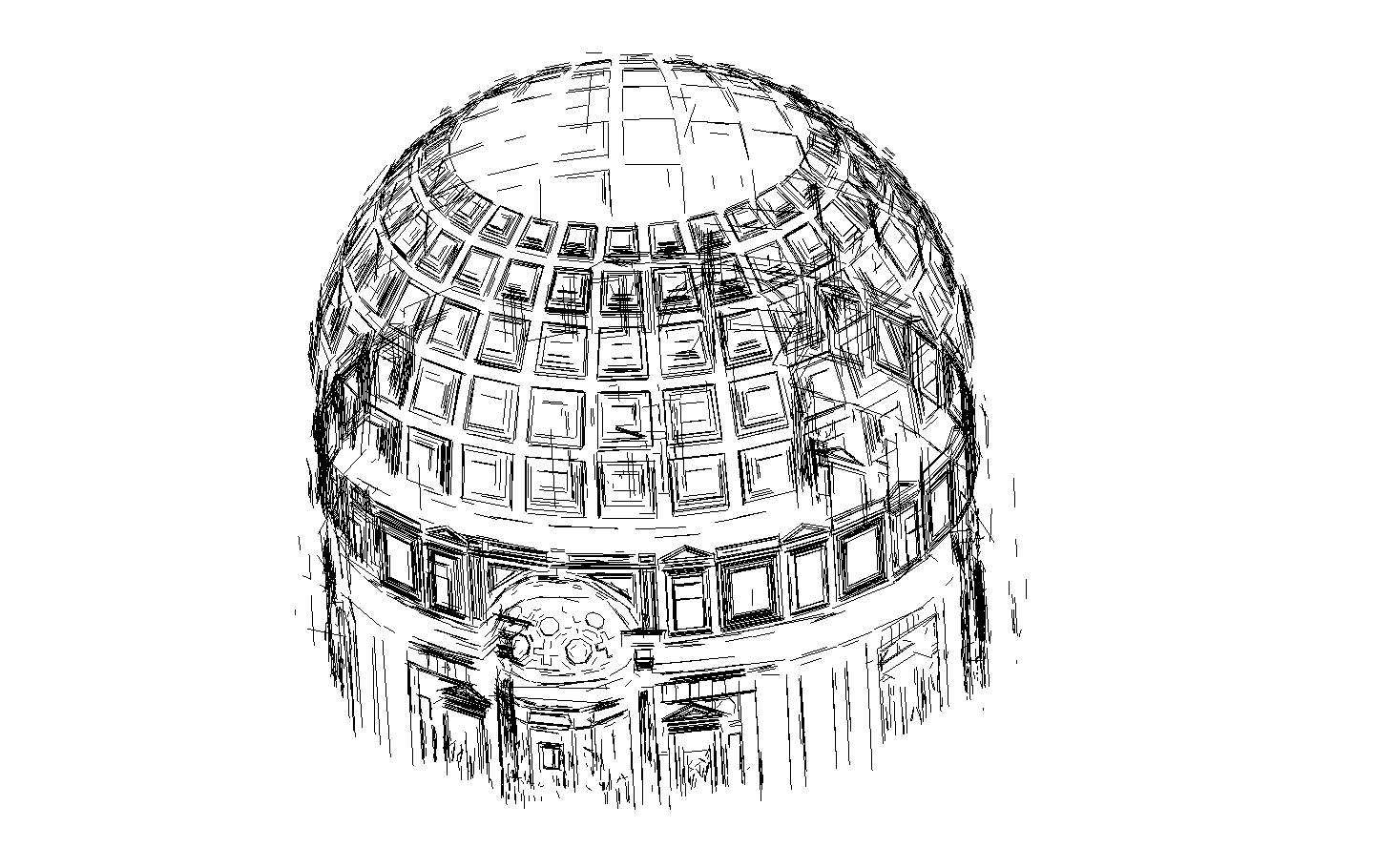}}} \\
\includegraphics[width=0.24\linewidth, height=12pt]{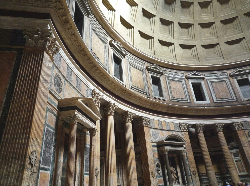} &
\end{tabular}
\end{minipage} 
&
\begin{minipage}{.19\linewidth}
\renewcommand{\arraystretch}{0.4}
\begin{tabular}{cc}
\includegraphics[width=0.24\linewidth, height=12pt]{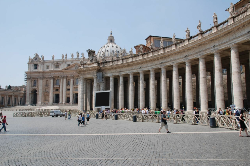} &
\raisebox{-0.55\height}[0pt][0pt]{{\includegraphics[width=0.7\linewidth]{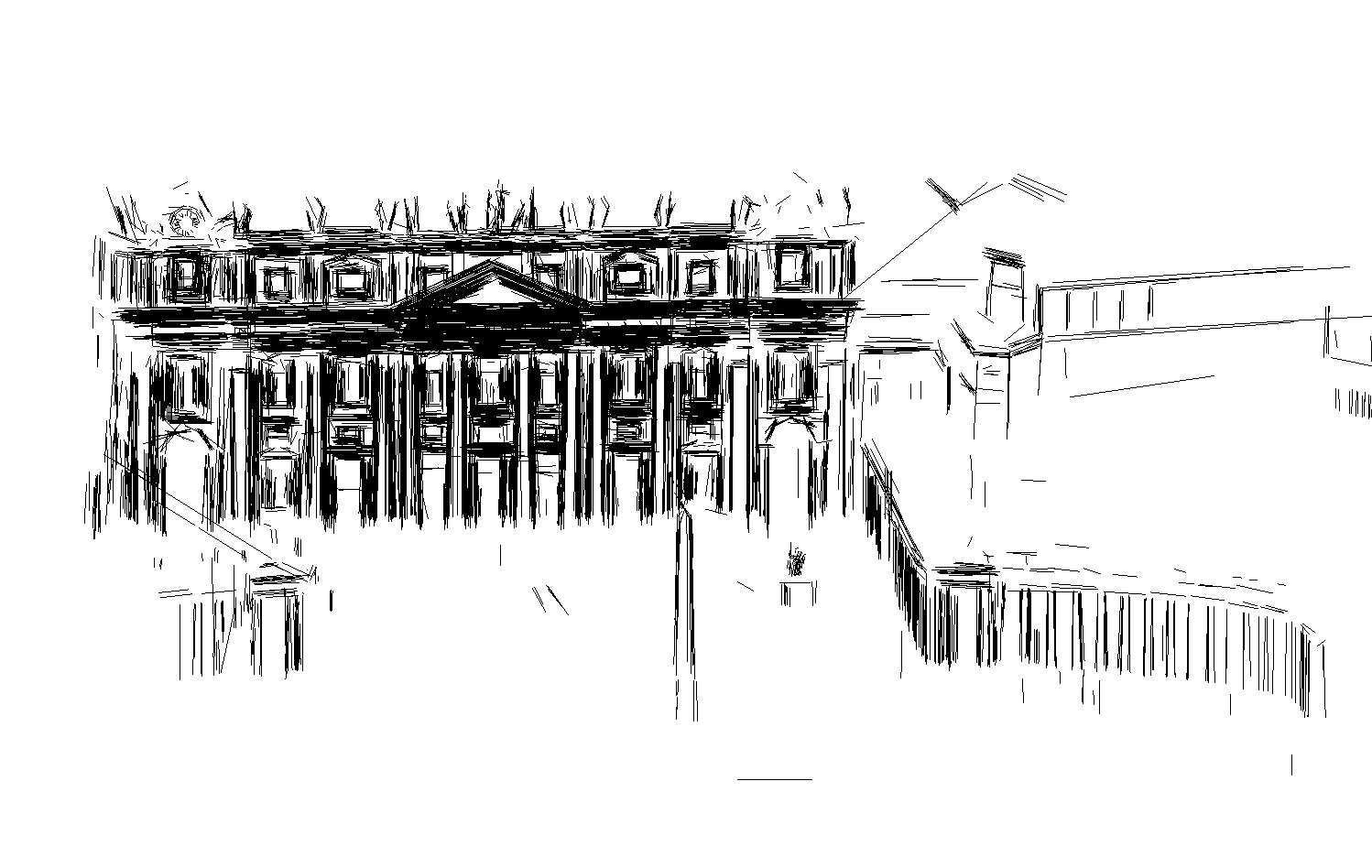}}} \\
\includegraphics[width=0.24\linewidth, height=12pt]{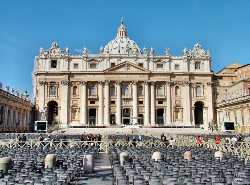} &
\end{tabular}
\end{minipage} 
&
\begin{minipage}{.19\linewidth}
\renewcommand{\arraystretch}{0.4}
\begin{tabular}{cc}
\includegraphics[width=0.24\linewidth, height=12pt]{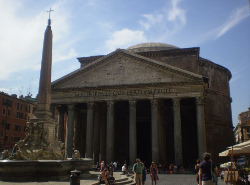} &
\raisebox{-0.55\height}[0pt][0pt]{{\includegraphics[trim={30 30 30 30}, clip, width=0.7\linewidth]{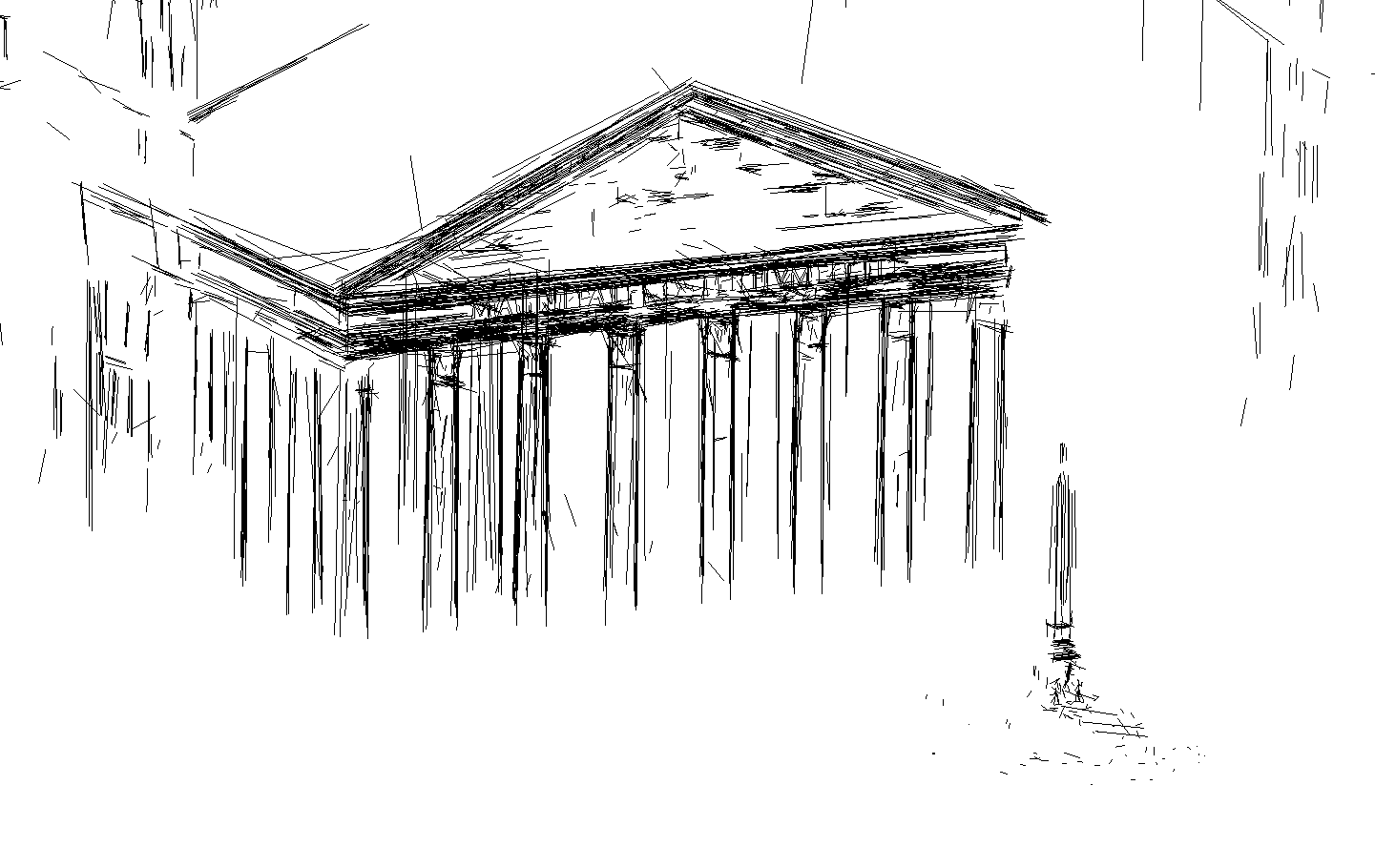}}} \\
\includegraphics[width=0.24\linewidth, height=12pt]{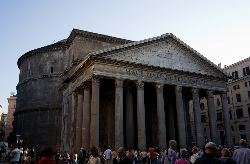} &
\end{tabular}
\end{minipage} 
&
\begin{minipage}{.19\linewidth}
\renewcommand{\arraystretch}{0.4}
\begin{tabular}{cc}
\includegraphics[width=0.24\linewidth, height=12pt]{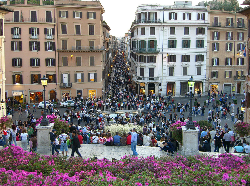} &
\raisebox{-0.55\height}[0pt][0pt]{{\includegraphics[width=0.7\linewidth]{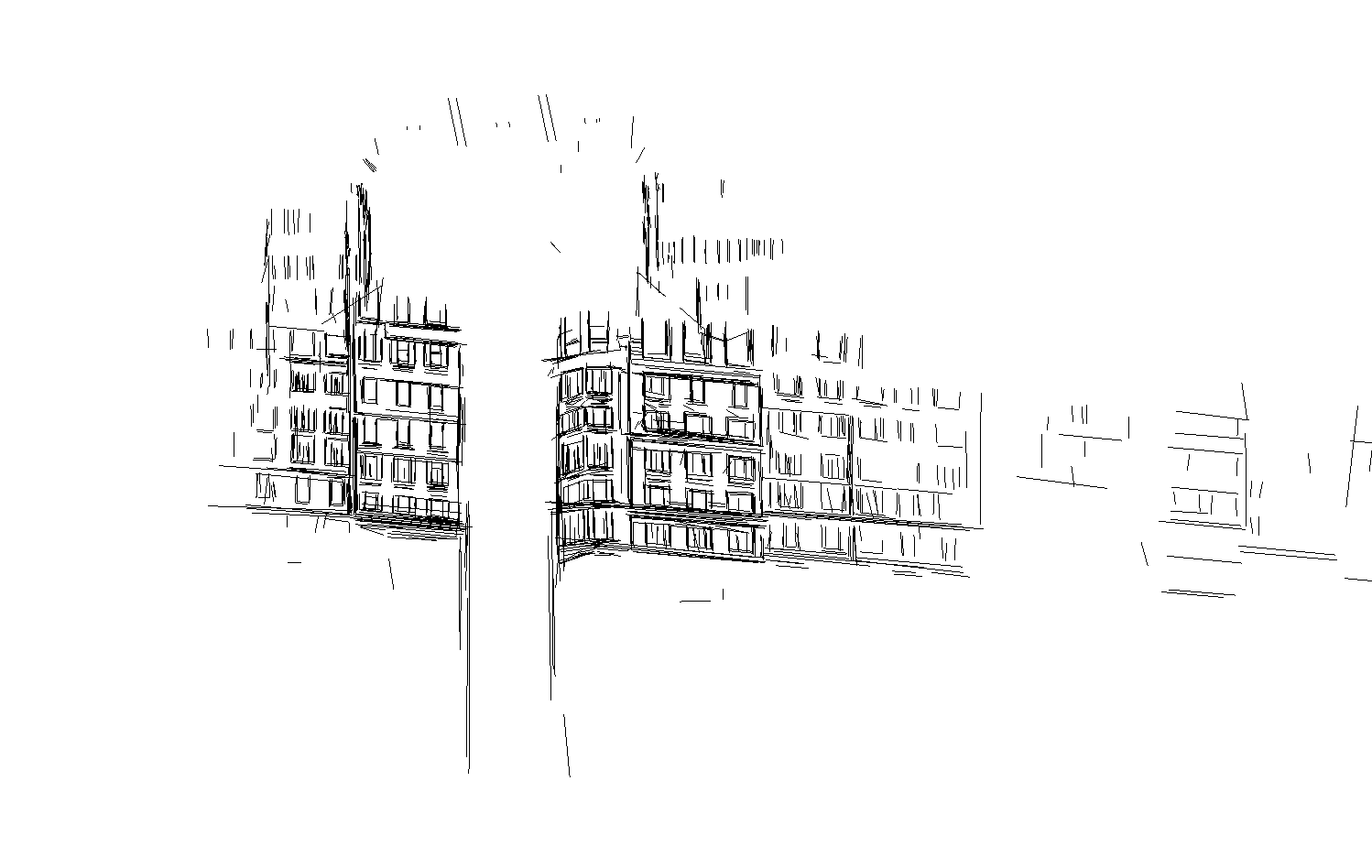}}} \\
\includegraphics[width=0.24\linewidth, height=12pt]{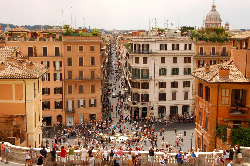} &
\end{tabular}
\end{minipage} 
&
\begin{minipage}{.19\linewidth}
\renewcommand{\arraystretch}{0.4}
\begin{tabular}{cc}
\includegraphics[width=0.24\linewidth, height=12pt]{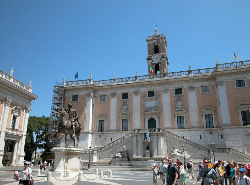} &
\raisebox{-0.55\height}[0pt][0pt]{{\includegraphics[width=0.7\linewidth]{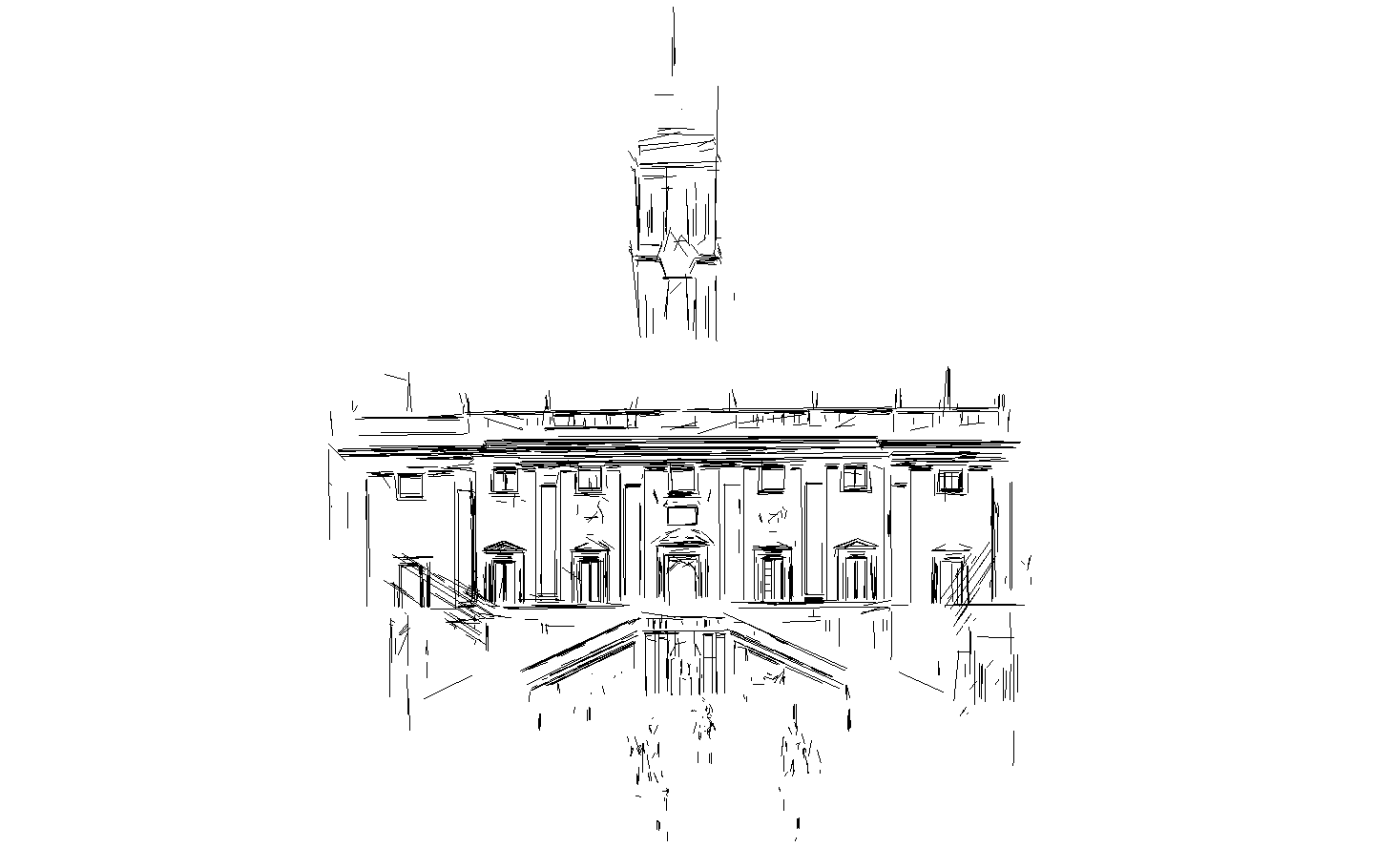}}} \\
\includegraphics[width=0.24\linewidth, height=12pt]{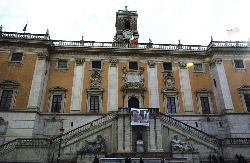} &
\end{tabular}
\end{minipage} 
\end{tabular}

\end{minipage}

\caption{\textbf{Scalability to large-scale datasets:} Aachen (6,697 images) \cite{sattler2018benchmarking} and Rome (16,179 images) \cite{snavely2006photo,snavely2008modeling,agarwal2011building}. For Aachen \cite{sattler2018benchmarking}, parallel lines from the line-VP association graph are colored the same. For Rome \cite{snavely2006photo,snavely2008modeling,agarwal2011building}, we visualize 10 representative components individually.}
\label{fig::scalibility}
\end{figure*}

To demonstrate the scalability of the proposed system, we also run our method on two large-scale datasets: Aachen (6,697 images) \cite{sattler2012image,sattler2018benchmarking} and Rome city (16,179 images) \cite{snavely2006photo,snavely2008modeling,agarwal2011building}. \cref{fig::scalibility} shows that our method produces reliable line maps with clear structures. Note that the camera poses from Bundler \cite{snavely2006photo} on Rome city are far from perfect, while our mapping still works reasonably well. The efficiency bottleneck is in line detection and matching (we use SOLD2 \cite{pautrat2021sold2} descriptors), while the rest of the mapping takes only $\sim$10 minutes on Aachen \cite{sattler2012image,sattler2018benchmarking}.
The time complexity of our system is nearly linear with the number of images. 

\subsection{More Insights and Ablation Studies}

\begin{figure}[tb]
\centering
\setlength\tabcolsep{2pt} 
{\includegraphics[width=0.45\columnwidth, height=60pt]{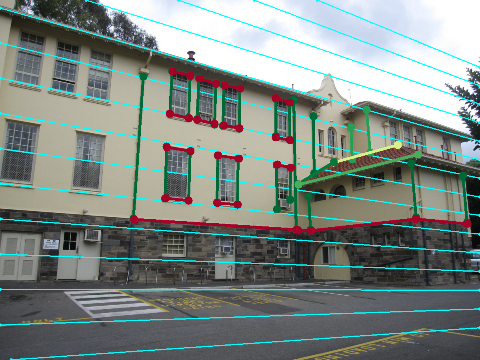}}
 \raisebox{-0.25cm}{\input{figs/synthetic}}

\vspace{-0.35cm}
\caption{\textbf{Uncertainty in line triangulation} measured by the largest eigenvalue of the covariance (Sec. D in supp.). \textbf{Left:} Each segment is colored by the uncertainty in the triangulation. Lines that align with the epipolar lines (shown in blue) exhibit higher (red) uncertainty. \textbf{Right:} We perform a small synthetic experiment to illustrate this. The graph shows the uncertainty for line triangulation as the lines approach the degenerate state. We compare with point-based triangulation assuming that endpoints are consistent.}
\label{fig::synthetic_tests}
\end{figure}

\begin{table}[tb]
\begin{center}
\scriptsize
\setlength{\tabcolsep}{3pt}
\begin{tabular}{clccccccc}
\toprule
Line type & Triangulation & R1 & R5 & R10 & P1 & P5 & P10 & \# supports \\
\midrule
\multirow{2}{*}{\makecell{LSD\\\cite{von2008lsd}}} & Endpoints & 27.6 & 101.4 & 138.0 & 58.2 & \textbf{83.5} & \textbf{92.1} & (13.0 / 13.2) \\
& Line & \textbf{48.3} & \textbf{187.0} & \textbf{257.4} & \textbf{59.2} & 81.9 & 89.8 & (\textbf{15.8} / \textbf{19.1}) \\
\midrule
\multirow{2}{*}{\makecell{SOLD2\\ \cite{pautrat2021sold2}}} & Endpoints & 27.3 & 82.8 & 106.5 & 68.2 & 84.5 & 90.9 & (12.3 / 19.9) \\
& Line & \textbf{50.8} & \textbf{143.5} & \textbf{180.8} & \textbf{74.4} & \textbf{86.9} & \textbf{91.2} & (\textbf{15.1} / \textbf{32.2}) \\
\bottomrule
\end{tabular}
\caption{\textbf{Comparison between endpoint and line triangulation} on Hypersim \cite{roberts:2021}. While being more stable at triangulation, the endpoints are often unmatched between line pairs. }
\label{tab::ablations_endpoints}
\end{center}
\end{table}

\begin{table}[tb]
\begin{center}
\scriptsize
\setlength{\tabcolsep}{3pt}
\begin{tabular}{cccc|ccccccc}
\toprule
Line & M1 & M2 & M3 & R1 & R5 & R10 & P1 & P5 & P10 & \# supports \\
\midrule
$\checkmark$ & & & & 50.8 & 143.5 & 180.8 & 74.4 & 86.9 & 91.2 & (15.1 / 32.2) \\
& $\checkmark$ & & & 24.9 & 72.5 & 95.8 & 65.9 & 81.2 & 88.5 & (11.3 / 15.7) \\
& $\checkmark$ & $\checkmark$ & & 37.7 & 116.8 & 152.6 & 71.0 & 84.2 & 89.7 & (13.8 / 25.8) \\
$\checkmark$ & $\checkmark$ & & & 51.5 & 146.9 & 185.4 & 71.7 & 85.4 & 90.1 & (14.9 / 31.2) \\
$\checkmark$ & $\checkmark$ & $\checkmark$ & & 51.3 & 146.4 & 186.4 & 73.4 & 85.7 & 90.5 & (15.8 / 35.6) \\
$\checkmark$ & $\checkmark$ & $\checkmark$ & $\checkmark$ & 51.4 & 145.4 & 184.9 & 74.1 & 86.1 & 90.6 & (16.5 / 38.7) \\
\bottomrule
\end{tabular}
\caption{\textbf{Ablation study} on different types of triangulation proposals (defined in Sec. \ref{sec::point-guided-triangulation}) on Hypersim \cite{roberts:2021} with SOLD2 \cite{pautrat2021sold2}. }
\label{tab::ablations_triangulation}
\end{center}
\end{table}

\noindent
\textbf{Line Triangulation.} To study the stability of the triangulation, we perform a small test on a stereo pair from AdelaideRMF \cite{AdelaideRMF} on the uncertainty (measured by the largest singular value of the covariance) of the triangulated 3D segments. We further run a synthetic experiment by generating random lines on a plane orthogonal to the stereo pair, and plot the uncertainty of point and line triangulations with respect to the angle of the lines with the baseline (refer to Sec. D in supp. for details). The results in \cref{fig::synthetic_tests} show that when the matched line is nearly parallel to the epipolar line, the line triangulation becomes degenerate with exploding uncertainty, while triangulating the endpoints is significantly more stable. Thus, combining points and VPs from the 2D association is beneficial to improve the stability of the proposals. However, the endpoints are generally not consistent across line matches in practice and need to be complemented with line-line triangulation. This can be verified in \cref{tab::ablations_endpoints} where the performance significantly drops when we change line triangulation into endpoint triangulation.

We further ablate our four types of triangulation for generating proposals. Results in \cref{tab::ablations_triangulation} show that integrating points and VPs enhance the 3D line maps, in particular significantly improving the track quality. Another surprising fact is that the third line in the table, relying only on points and line + point triangulation, already achieves better results than the prior baselines in \cref{tab::main-hypersim}. Employing all four types of proposals obtains the best trade-off.

\begin{table}[tb]
\begin{center}
\scriptsize
\setlength{\tabcolsep}{3pt}
\begin{tabular}{clccccc}
\toprule
Line type & Method & R1 & R5 & P1 & P5 & \# supports \\
\midrule
\multirow{5}{*}{\makecell{LSD\\\cite{von2008lsd}}} & L3D++ \cite{hofer2017efficient} & 37.0 & 153.1 & 53.1 & 80.8 & (14.8 / 16.8) \\
& Ours (line) w/  \cite{hofer2017efficient} scoring & \textbf{48.6} & 186.0 & 56.5 & 80.6 & (14.4 / 16.8) \\
& Ours (line) w/  \cite{hofer2017efficient} merging & 41.2 & 158.2 & \textbf{59.6} & \textbf{82.5} & (15.6 / 16.7) \\
& Ours (line) w/ exhaustive & 46.7 & 177.2 & 57.6 & 80.9 & (\textbf{16.8} / \textbf{20.8}) \\
& Ours (line) & 48.3 & \textbf{187.0} & 59.2 & 81.9 & (15.8 / 19.1) \\
\midrule
\multirow{5}{*}{\makecell{SOLD2\\ \cite{pautrat2021sold2}}} & L3D++ \cite{hofer2017efficient} & 36.9 & 107.5 & 67.2 & 86.8 & (13.2 / 20.4) \\
& Ours (line) w/  \cite{hofer2017efficient} scoring & 45.8 & 133.2 & 72.6 & 85.9 & (15.0 / 31.1) \\
& Ours (line) w/ \cite{hofer2017efficient} merging & 37.7 & 113.4 & 70.5 & 84.5 & (13.3 / 23.9) \\
& Ours (line) w/ exhaustive & 48.9 & 139.7 & 72.9 & 85.7 & (\textbf{16.2} / \textbf{36.9}) \\
& Ours (line) & \textbf{50.8} & \textbf{143.5} & \textbf{74.4} & \textbf{86.9} & (15.1 / 32.2) \\
\bottomrule
\end{tabular}
\caption{\textbf{Studies on different components of our method} with only line-line proposals against L3D++ \cite{hofer2017efficient}.}
\vspace{-5pt}
\label{tab::ablations_matcher}
\end{center}
\end{table}

\noindent
\textbf{Scoring and Track Building.}
We first study the effects of using exhaustive line matching as in L3D++ \cite{hofer2017efficient}. To enable direct comparison we only use line triangulation proposals. Results are shown in \cref{tab::ablations_matcher}. While there are more proposals generated from the exhaustive matches, both the recall and precision decrease by a noticeable margin. This is probably due to the large number of wrong proposals misleading the scoring process. Nevertheless, our method with exhaustive matches still works significantly better than L3D++ \cite{hofer2017efficient}. To further study the effects of the proposed distance measurements at scoring and track building (merging), we re-implement the ones proposed in L3D++ \cite{hofer2017efficient} and perform direct comparison. Both our scoring and track building are significantly better, especially when equipped with SOLD2 \cite{pautrat2021sold2} which produces more structured lines. 

\noindent
\textbf{Joint Optimization.}
\begin{table}[tb]
\begin{center}
\scriptsize
\setlength{\tabcolsep}{3pt}
\begin{tabular}{lccccccc}
\toprule
Method & R1 & R5 & R10 & P1 & P5 & P10 & \# supports \\
\midrule
Line-only w/o refine & 43.5 & 135.8 & 180.1 & 75.1 & 87.2 & 92.2 & (15.1 / 32.2) \\
Line-only w/ geom alone & 50.8 & 143.5 & 180.8 & 74.4 & 86.9 & 91.2 & (15.1 / 32.2) \\
\midrule
w/o refine & 46.5 & 146.0 & 189.7 & \textbf{76.8} & \textbf{88.9} & \textbf{93.3} & (\textbf{16.5} / \textbf{38.7}) \\
w/ geom alone & 51.4 & 145.4 & 184.9 & 74.1 & 86.1 & 90.6 & (\textbf{16.5} / \textbf{38.7}) \\
\midrule
w/ joint optimization & \textbf{54.3} & \textbf{151.1} & \textbf{191.2} & 69.8 & 84.6 & 90.0 & (\textbf{16.5} / \textbf{38.7}) \\
\bottomrule
\end{tabular}
\caption{\textbf{Line refinement} on Hypersim \cite{roberts:2021} with SOLD2 \cite{pautrat2021sold2}.}
\label{tab::refine-hypersim}
\end{center}
\end{table}

Finally, we ablate the proposed joint optimization in our pipeline. First, we remove the point-line association and only apply the geometric residuals (reprojection error). Results in \cref{tab::refine-hypersim} show that the geometric refinement improves significantly when the proposals solely come from line triangulation. However, when adding additional proposals from points and VPs, it contributes marginally and even misleads some lines that are generated from points and VPs but poorly conditioned for lines (R10 decreases). When integrated with joint optimization with soft association, the recall is further improved noticeably, while sacrificing a bit on the precision. It is worth pointing out that the joint optimization also enables the byproduct of junction structures and line-line relations (e.g. in \cref{fig::qualitative_structures}).

\subsection{Applications}

\begin{table}[tb]
    \centering
    \scriptsize
    \setlength{\tabcolsep}{5pt}
    \begin{tabular}{lccc}
        \toprule
        Dataset & HLoc\footnotemark~\cite{sarlin2019coarse,hloc} & PtLine \cite{gao2022pose} & Ours \\
        \midrule
        Cambridge~\cite{kendall2017geometric} & 7.0 / 0.13 / 44.0 & 7.4 / 0.13 / 43.5 & \textbf{6.7} / \textbf{0.12} / \textbf{46.1} \\
        7Scenes~\cite{7scenes} & 3.3 / 1.08 / 73.0 & 3.3 / 1.09 / 72.7 & \textbf{3.0} / \textbf{1.00} / \textbf{78.0} \\
        \bottomrule
    \end{tabular}\textbf{}
    \caption{\textbf{Visual localization on Cambridge~\cite{kendall2015posenet} and 7Scenes~\cite{7scenes}}. We report the median translation and rotation errors in cm and degrees, and the pose accuracy (\%) at 5 cm / 5 deg threshold. All metrics are averaged across all scenes of each dataset.}
    \label{tab:localization}
\end{table}

\footnotetext{Up to the date of submission, the COLMAP model \cite{schonberger2016structure} used by HLoc \cite{sarlin2019coarse,hloc} does not consider radial distortion from the VisualSfM \cite{wu2011visualsfm} model. So our results are better than the original ones.}

\begin{figure}[tb]
\begin{minipage}{.46\linewidth}
\scriptsize
\setlength\tabcolsep{1.5pt} 
\begin{tabular}{cc}
{\includegraphics[width=0.47\linewidth, height=56pt]{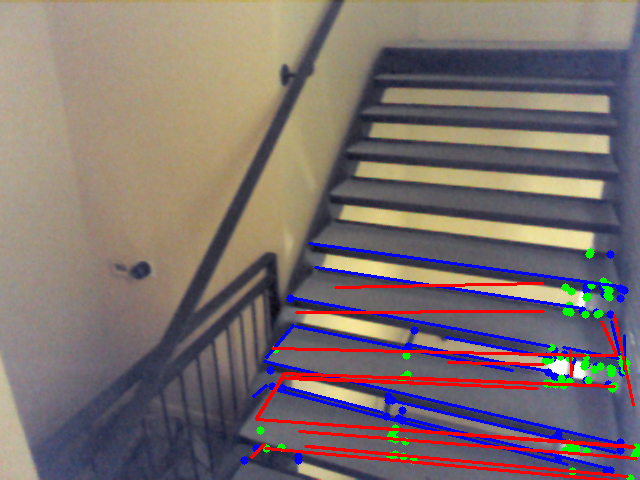}} &
{\includegraphics[width=0.47\linewidth, height=56pt]{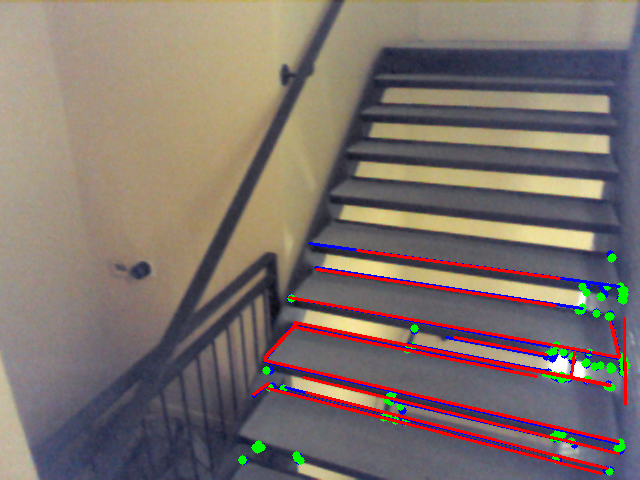}} \\
HLoc~\cite{sarlin2019coarse,hloc} & Ours w/ LIMAP
\end{tabular}
\end{minipage}
\begin{minipage}{.52\linewidth}
\centering
\scriptsize
\setlength{\tabcolsep}{2pt}
\begin{tabular}{lcc}
    \toprule
    & (T / R) err. $\downarrow$ & Acc. $\uparrow$  \\
    \midrule
    HLoc~\cite{sarlin2019coarse} & 5.2 / 1.46 & 46.8 \\
    HLoc~\cite{sarlin2019coarse} w/ depth & 4.7 / 1.25 & 53.4 \\
    \midrule
    PtLine~\cite{gao2022pose} & 4.8 / 1.33 & 51.9 \\
    Ours w/ L3D++~\cite{hofer2017efficient} & 4.1 / 1.14 & 60.8 \\
    Ours w/ LIMAP & \textbf{3.7} / \textbf{1.02} & \textbf{71.1} \\
    \bottomrule
\end{tabular}
\end{minipage}
\caption{\textbf{Line-assisted Visual localization} on \textit{Stairs} from 7Scenes ~\cite{7scenes}. Blue: 2D points/lines; Green/Red: Projected 3D points/lines. }
\label{fig::stairs}
\end{figure}

\noindent
\textbf{Line-Assisted Visual Localization.}
We build a hybrid visual localization with both points and lines on top of the acquired 3D line maps. Specifically, we first build point maps as in HLoc \cite{sarlin2019coarse,hloc} and line maps with our proposed method. Then, we match points and lines respectively and get 2D-3D correspondences from the track information in the 3D maps. Given these correspondences, we combine four minimal solvers \cite{kukelova2016efficient,persson2018lambda,zhou2018stable}: P3P, P2P1LL, P1P2LL, P3LL from PoseLib \cite{PoseLib}, together in a hybrid RANSAC framework \cite{camposeco2018hybrid,Sattler2019Github} with local optimization \cite{chum2003locally,Lebeda2012BMVC} to get the final 6-DoF pose (refer to Sec. H in supp. for details). This also enables direct comparison since only using P3P \cite{persson2018lambda} corresponds to the point-alone baseline similar to HLoc \cite{sarlin2019coarse,hloc}. We also compare with the post-refinement of PtLine \cite{gao2022pose} that optimizes over the initial point-alone predictions.

Results in \cref{tab:localization} show that our localization system achieves consistently better results than the point-alone baseline both indoors \cite{7scenes} and outdoors \cite{kendall2017geometric}, validating the effectiveness of employing 3D line maps for visual localization. 
In \cref{fig::stairs} we show more detailed results from the \textit{Stairs} scene from 7Scenes~\cite{7scenes} as it is one of the most challenging ones. Integrating lines significantly benefits the alignment of the reprojected structures, improving the pose accuracy from 46.8 to 71.1. Also, with our localization pipeline, using the map built from our proposed method is better than from L3D++ \cite{hofer2017efficient} by a noticeable margin, again demonstrating the advantages of our proposed line mapping system. 
Refer to Sec. H in supp. for more results on InLoc~\cite{taira2018inloc}.

\noindent
\textbf{Refining Structure-from-Motion.}
\begin{table}[tb]
    \centering
    \scriptsize
    \setlength{\tabcolsep}{4.5pt}
    \begin{tabular}{lcccc}
        \toprule
        & Med. error $\downarrow$ & AUC @ (1$^{\circ}$ / 3$^{\circ}$ / 5$^{\circ}$) $\uparrow$ \\
        \midrule
        COLMAP \cite{schonberger2016structure} & 0.188 & 77.3 / 89.0 / 91.6 \\ 
        COLMAP \cite{schonberger2016structure} + LIMAP refinement & \textbf{0.146} & \textbf{82.9} / \textbf{91.2} / \textbf{93.0} \\
        \bottomrule
    \end{tabular}
    \caption{\textbf{Joint bundle adjustment of points and lines} on Hypersim \cite{roberts:2021}. Relative pose errors are measured on all image pairs.}
    \label{tab:sfm_hypersim}
\end{table}
With the acquired 3D line maps built from a roughly correct point-based structure-from-motion model, e.g, COLMAP \cite{schonberger2016structure}, we can use the 3D lines with their track information to refine the input camera poses with joint optimization of points and lines. To verify this, we run COLMAP \cite{schonberger2016structure} with SuperPoint \cite{detone2018superpoint} on the first eight scenes of Hypersim \cite{roberts:2021}, run the proposed line mapping on top of it, and perform joint bundle adjustment to refine poses and intrinsics.
We report the relative pose evaluation of all image pairs~\cite{jin2021image}.
\cref{tab:sfm_hypersim} shows that the joint point-line refinement consistently benefits the accuracy of the camera poses, in particular improving AUC@1$^\circ$ by 5.6.

\section{Conclusion}

\label{sec::conclusion}

In this paper, we introduce LIMAP: a library for robust 3D line mapping from multi-view imagery. Extensive experiments show that our method, by improving all stages of the reconstruction pipeline, produces significantly more complete 3D lines, with much higher quality of track association. As a byproduct, the method can also recover 3D association graphs between lines and points / VPs. We further show the usefulness of 3D line maps on visual localization and bundle adjustment. Future directions include incremental / real-time structure mapping, distinguishing structural lines from textural lines for wireframe modeling, and exploiting higher-level structures and relations for downstream applications.

\noindent
\textbf{Acknowledgements.}
We sincerely thank the reviewers for their constructive feedback. We are grateful to Philipp Lindenberger, Paul-Edouard Sarlin, Johannes Sch\"onberger for their open-source projects, Iago Su\'arez, Hengkai Guo, Daniel Barath, Siyan Dong, Martin Oswald, Jing Ren, Iro Armeni for helpful discussions, Marcel Geppert, Daniel Thul, Peizhuo Li for technical support on Blender. Special thanks to Mihai Dusmanu and Wang Zhao for proof-reading. Viktor Larsson received funding by the strategic research project ELLIIT.

\section*{Appendix}
\appendix
\appendix
This document provides a list of supplementary materials that accompany the main paper. The content is organized as follows:

\begin{itemize}
    \item In Section \ref{sec::supp_plucker}, we introduce the notation, parameterization, and transformations of Pl\"ucker coordinates that are used throughout the system, in particular at joint optimization.
    \item In Section \ref{sec::supp_triangulation}, we provide detailed derivations for different types of triangulations discussed in the main paper.
    \item In Section \ref{sec::supp_association}, we provide details on constructing the association graph among lines and points / vanishing points, and further introduce some examples of its extensions to higher-level applications.
    \item In Section \ref{sec::supp_covariance}, we provide details on computing the covariance used in the main paper for endpoint triangulation and algebraic line triangulation, and also give details on the setup of the corresponding synthetic tests.
    \item In Section \ref{sec::supp_fitnmerge}, we extend the discussion in the main paper to show how our system can be easily extended to map lines with available depth maps, and further present relevant experimental results on mapping and localization.
    \item In Section \ref{sec::supp_implementation_details}, more implementation details are provided on datasets, hyperparameters, detailed distance measurements and the two baseline methods \cite{hofer2017efficient,wei2022elsr} in the experiments.
    \item In Section \ref{sec::results_mapping}, we present more experimental results and additional analysis on our 3D line mapping system.
    \item In Section \ref{sec::supp_localization}, we provide details on our proposed line-assisted visual localization system, along with more results and comparisons against the baseline method \cite{gao2022pose}, including point-line localization results on InLoc dataset \cite{taira2018inloc}.
    \item In Section \ref{sec::supp_sfm_refinement}, we present more results on refining point-based structure-from-motion with our proposed line mapping system.
    \item In Section \ref{sec::supp_mvs}, we show preliminary results on how to adapt the acquired 3D line maps into the PatchMatch Stereo pipeline \cite{schonberger2016pixelwise} to improve the completeness of dense reconstruction.
    \item In Section \ref{sec::supp_featuremetric}, we present how to extend featuremetric optimization over the acquired line tracks to improve the pixelwise alignment with deep features.
    \item Finally, in Section \ref{sec::supp_limitations}, we expand the conclusions in the main paper and discuss more on the limitations and future work.
\end{itemize}

\section{Background: Pl\"ucker Coordinate}
\label{sec::supp_plucker}
Here we revisit how to represent an infinite line with its Pl\"ucker coordinate \cite{hartley2003multiple}. We first present the definition and its 4 DoF minimal parameterization \cite{bartoli2005structure}. Then, we show how to apply geometric operations on top of it.

\subsection{Definition}
A 3D line segment is compactly encoded with its two 3D endpoints $\vec{p}_s$ and $\vec{p}_e$, which exhibit six degrees of freedom. Its corresponding infinite 3D line, however, has only 4 degrees of freedom, as both points can be moving along the line direction. Thus, representing an infinite 3D line with two 3D points is not a compact representation in the sense that two coordinates can be both feasible and correspond to the same infinite 3D line. 
The Pl\"ucker coordinate \cite{hodge1947methods,mason2001mechanics,bartoli2005structure} $(\vec{d}, \vec{m})$ is a compact representation for an infinite 3D line, where $\vec{d}$ is the normalized direction of the 3D line, and $\vec{m}$ is the moment that is invariant to any point $\vec{p}$ along the line:
\begin{equation}
    \vec{d} = \frac{\vec{p}_e - \vec{p}_s}{\lVert \vec{p}_e - \vec{p}_s \rVert}
\end{equation}
\begin{equation}
\label{eq::plucker-moment}
    \vec{m} = \vec{p} \times \vec{d} = \vec{p}_s \times \vec{d} = \vec{p}_e \times \vec{d}.
\end{equation}

The property in \eqref{eq::plucker-moment} is due to the fact that $(\vec{p} - \vec{p}_s) \times \vec{d}=\vec{0}$, where $\vec{p}$ is any point along the line. The coordinate is convenient in the sense that we can directly perform transformations and projections efficiently on top of it, which will be presented in the following subsections.

\subsection{Minimal Parameterization}
We first discuss here how to minimally parameterize a Pl\"ucker coordinate $(\vec{d}, \vec{m})$ in a
non-linear optimization, e.g. our joint optimization scheme. The minimal parameterization was initially discussed in \cite{bartoli2005structure} as the orthonormal representation. A Pl\"ucker coordinate can be minimally represented with:

\begin{equation}
    (\vec{U}, \vec{W}) \in SO(3) \times SO(2).
\end{equation}
This results in the minimal $3 + 1 = 4$ degrees of freedom for the infinite 3D line. Specifically, since $\vec{d}$ is orthogonal to $\vec{m}$, we can represent the coordinate $(\vec{d}, \vec{m})$ with:

\begin{equation}
    \vec{U} = (\vec{d} \ \ \frac{\vec{m}}{\lVert \vec{m} \rVert} \ \ \frac{\vec{d} \times \vec{m}}{\lVert \vec{d} \times \vec{m} \rVert}) \in SO(3),
\end{equation}

\begin{equation}
    \vec{W} = 
    \begin{pmatrix}
    w_1 & w_2 \\
    -w_2 & w_1
    \end{pmatrix} \in SO(2),
\end{equation}

\begin{equation}
    w_1 = \frac{1}{\sqrt{1 + \lVert \vec{m} \rVert^2}}, \ \ w_2 = \frac{\lVert \vec{m} \rVert}{\sqrt{1 + \lVert \vec{m} \rVert^2}}.
\end{equation}

Denote $\vec{U}$ as $\vec{U}=(\vec{u}_1 \ \ \vec{u}_2 \ \ \vec{u}_3)$, we can easily recover the original Pl\"ucker coordinate $(\vec{d}, \vec{m})$ with the minimal parameterization by:

\begin{equation}
    \vec{d} = \vec{u}_1, \ \ \vec{m} = \frac{w_2}{w_1}\vec{u_2}
\end{equation}

At optimization, we can parameterize $\vec{U}\in SO(3)$ with a quaternion, and $\vec{V}\in SO(2)$ with a 2-dimensional homogeneous parameterization using Ceres \cite{ceres}.

\subsection{Perspective Projection}

The Pl\"ucker coordinate $(\vec{d}, \vec{m})$ can be written in matrix form. The $4 \times 4$ Pl\"ucker matrix $\vec{L}$ is formulated as:

\begin{equation}
    \vec{L} = 
    \begin{pmatrix}
    [\vec{m}]_\times & \vec{d} \\
    -\vec{d} & 0
    \end{pmatrix}.
\end{equation}

We here directly provide the clean formulation from \cite{hartley2003multiple} for projecting an infinite 3D line with Pl\"ucker matrix $\vec{L}$ perspectively with a $3 \times 4$ projection matrix $\vec{P}$:

\begin{equation}
    [\vec{l}]_\times = \vec{P} \vec{L} \vec{P}^T,
\end{equation}
where $\vec{l}$ is the 3-dimensional homogeneous coordinate for the resulting 2D infinite line, where on the 2D image we have $\vec{l}^T[x, y, 1] = 0$.

\subsection{Point-to-Line Projection}
\label{sec::supp_plucker_pointtoline}
Here we discuss how to project a 3D point $\vec{p}$ onto the infinite 3D line represented with Pl\"ucker coordinate $(\vec{d}, \vec{m})$. Specifically, we can compute the moment $\vec{m}_p$ of the line with respect to the 3D point $\vec{p}$ (rather than the origin):
\begin{equation}
    \vec{m}_p = \vec{m} + \vec{d} \times \vec{p}.
\end{equation}

Then, the projection of the 3D point $\vec{p}_\perp$ on the infinite 3D line can be computed as:
\begin{equation}
    \vec{p}_\perp = \vec{p} + \vec{d} \times \vec{m}_p = \vec{p} + \vec{d} \times (\vec{m} + \vec{d} \times \vec{p}).
\end{equation}

We can use this property to efficiently compute the projection of the 3D point without computing squared distances, which gives us robustness to numerical issues at joint optimization with soft point-line associations.

\subsection{Line-to-Line Projection}
\label{sec::supp_plucker_linetoline}
The line-to-line projection aims to find the point on the line (line 1) with Pl\"ucker coordinate $(\vec{d}_1, \vec{m}_1)$ that is closest to the projected line  (line 2) with Pl\"ucker coordinate $(\vec{d}_2, \vec{m}_2)$. This operation is particularly useful in our system at the following steps:

\begin{itemize}
    \item \textbf{M1. Triangulation with multiple points.} We need to project the infinite 3D line fitted from multiple 3D points onto the camera rays of the two endpoints in the reference image.
    \item \textbf{Endpoint aggregation at joint optimization.} We aim to get a rough estimate of the 3D endpoints on the optimized infinite 3D line, without 3D line proposals. Here we can project the camera rays from the endpoints of each 2D support onto the infinite 3D line.
    \item \textbf{Cheirality test for point-line localization.} We need to test, for each 2D-3D line correspondence, if the 3D line segment has positive depth at the range of the unprojection of its 2D support. Here we can project the camera rays from the endpoints of the 2D supporting line segment onto the infinite 3D line of the 3D line segment to get the ranges where the cheirality test is applied.
\end{itemize}

Specifically, take ${p_{2\rightarrow1}}_\perp$ as the projection of infinite line 2 with $(\vec{d}_2, \vec{m}_2)$ onto infinite line 1 with $(\vec{d}_1, \vec{m}_1)$, the point can be computed with Pl\"ucker coordinate \cite{hartley2003multiple} as follows:

\begin{equation}
    {p_{2\rightarrow1}}_\perp = \frac{-\vec{m}_1 \times (\vec{d}_2 \times (\vec{d}_1 \times \vec{d}_2)) + (\vec{m}_2^T (\vec{d}_1 \times \vec{d}_2))\vec{d}_1}{\lVert \vec{d}_1 \times \vec{d}_2 \rVert^2}
\end{equation}

The other way can be computed similarly by substitution of variables. 

\section{Detailed Derivations for Different Types of Triangulations}
\label{sec::supp_triangulation}
As discussed in the paper, we propose four types of different triangulation methods to generate the 3D proposals for each 2D line segment, including the straightforward algebraic line triangulation, plus three advanced triangulation methods utilizing commonly associated points and a vanishing point direction. To ensure completeness as well as support the later discussion in Section \ref{sec::supp_covariance}, here we will provide detailed derivations for each of the four triangulation methods. 

\subsection{Algebraic Line Triangulation}
We start with the conventional line triangulation with back-projected planes. Because we aim to get the 3D endpoints of the triangulated line segment rather than the infinite line, the algebraic line triangulation is geometrically two ray-plane intersection problems between the camera rays from the endpoints $\vec{x}_1^r$, $\vec{x}_2^r$ of the reference line segment and the back-projected plane from the matched line segment spanned by the camera rays of $\vec{x}_1^m$, $\vec{x}_2^m$ on the target image. Here $\vec{x}_1^r$, $\vec{x}_2^r$, $\vec{x}_1^m$, $\vec{x}_2^m$ are all in homogeneous coordinates normalized by the camera intrinsics.

As in the main paper, assume without loss of generality that the world coordinate system aligns with the reference view, while the camera pose of the matched view is $(R, \vec{t})$. Then the intersection point $\vec{X}_i=\lambda_i \vec{x}_i^r$ ($i = 1, 2$) can be written in the linear combination of the two camera rays of the endpoints from the matched segments:

\begin{equation}
    \vec{X}_i = \lambda_i \vec{x}_i^r = -R^T \vec{t} + \beta_1 R^T\vec{x}_1^m + \beta_2 R^T\vec{x}_2^m.
\end{equation}

This results in a $3 \times 3$ linear system to solve for $[\lambda_i, \beta_1, \beta_2]$ for $i=1,2$, which will be used in the derivation of the covariance in Section \ref{sec::supp_covariance}. By multiplying $R$ and adding $\vec{t}$ in both sides we have:
\begin{equation} \label{eq::line_triangulation_3x3}
    R(\lambda_i \vec{x}_i^r) + \vec{t} = \beta_1 \vec{x}_1^m + \beta_2 \vec{x}_2^m.
\end{equation}
Then, by multiplying $(\vec{x}_1^m \times \vec{x}_2^m)^T$ in both sides:
\begin{equation}
\label{eq::supp_ray_plane_intersection}
    (\vec{x}_1^m \times \vec{x}_2^m)^T(R(\lambda_i\vec{x}_i^r) + \vec{t}) = 0,
\end{equation}

which is similar to Eq. (2) in the main paper. Here the equation can be taken as the point $\vec{X}_i = \lambda_i \vec{x}_i$ satisfying the equation of the back-projected plane.

\subsection{M1. Triangulation with Multiple Points}
This triangulation applies to the case when multiple ($\geq2$) common 3D points are available between the reference line segment and the matched one by traversing the 2D point-line association graphs. Here, we can fit an infinite 3D line by computing the mean and principle direction over all the points, and then project the infinite 3D line onto the two camera rays for $\vec{x}_1^r$ and $\vec{x}_2^r$ with Pl\"ucker coordinates discussed in Section \ref{sec::supp_plucker_linetoline}.

\subsection{M2. Line + Point: Triangulation with a Known 3D Point}
For each commonly shared 3D point between the reference line segment and the matched segment, we can formulate a line triangulation solution using a known 3D point. Compared to algebraic line triangulation, M2 can generate stable endpoints for weakly degenerate cases where one of the two endpoints has degenerate configurations.

As discussed in the paper, we ensure that the endpoints of the generated proposal lie on the camera rays of $\vec{x_1^r}$ and $\vec{x_2^r}$. such that $\vec{X}_i=\lambda_i \vec{x}_i^r$ for $i=1, 2$. Therefore, the problem becomes a constrained least square problem with respect to the ray depths of the two endpoints $\vec{\lambda}=(\lambda_1, \lambda_2)$. The least-square error is the residual from Eq. (2) (or equivalently, \eqref{eq::supp_ray_plane_intersection}), which is quadratic to $\vec{\lambda}$. We can denote the least square residual as $\vec{\lambda}^TA\vec{\lambda} + \vec{b}^T\vec{\lambda}$ without loss of generality.

Since both endpoints lie on the camera rays with $\vec{X}_i=\lambda_i \vec{x}_i^r$, we can convert the problem into a 2D subproblem by applying a global rotation $R^r$ such that $R^r \vec{x}_i^r$ has zero value at the third dimension for $i=1,2$. Next, we will discuss how to acquire a closed-form solution for $\vec{\lambda}$ to this 2D subproblem. 

\subsubsection{Closed-form Solution to the 2D Subproblem}
Let $\vec{p}_1$ and $\vec{p}_2$ be the 2D points (after applying $R^r$ and removing the third dimension). The back-projected 3D endpoints become $\vec{v}_1 = \lambda_1 \vec{p}_1,~\vec{v}_2 = \lambda_2 \vec{p}_2$. Take $\vec{p}_0$ as the 2D projection of the known 3D point on the plane, the constraints for the three points $p_0, v_1, v_2$ to be collinear is:
\begin{equation}
    \text{det}([\vec{v}_1-\vec{p}_0,~\vec{v}_2-\vec{p}_0]) = 0
\end{equation}
This is a quadratic equation in $\lambda_1$ and $\lambda_2$, and can thus be written as
\begin{equation}
    \vec{\lambda}^TQ \vec{\lambda} + \vec{q}^T\vec{\lambda} = 0
\end{equation}
Note that there is no constant term since $\text{det}(\vec{p}_0, \vec{p}_0) = 0$. Combining the least square error $\vec{\lambda}^TA\vec{\lambda} + \vec{b}^T\vec{\lambda}$ and introducing Lagrange multiplier $\mu$ we have:
\begin{equation}
    \mathcal{L}(\vec{\lambda},\mu) = \vec{\lambda}^TA \vec{\lambda} + \vec{b}^T\vec{\lambda} + \mu (\vec{\lambda}^TQ \vec{\lambda} + \vec{q}^T\vec{\lambda}).
\end{equation}
First-order constraints are then
\begin{equation} \label{eq:dL1}
    \frac{\partial \mathcal{L}}{\partial \vec{\lambda}} = 2A\vec{\lambda} + \vec{b} + 2\mu Q\vec{\lambda} + \mu \vec{q} = 0
\end{equation}
\begin{equation} \label{eq:dL2}
    \frac{\partial \mathcal{L}}{\partial \mu} = \vec{\lambda}^TQ \vec{\lambda} + \vec{q}^T\vec{\lambda} = 0
\end{equation}
From \eqref{eq:dL1} we get $\vec{\lambda}$ as a function of $\mu$,
\begin{equation} \label{eq:lambda_from_mu}
    \vec{\lambda}(\mu) = \frac{-1}{2} \left( A + \mu Q \right)^{-1} (\vec{b} + \mu \vec{q}),
\end{equation}
Then inserting into \eqref{eq:dL2} we get
\begin{equation}
    p(\mu) = \vec{\lambda}(\mu)^TQ \vec{\lambda}(\mu) + \vec{q}^T\vec{\lambda}(\mu) 
\end{equation}
which is a rational function in $\mu$. The numerator is a degree quartic polynomial in $\mu$ which can be solved in closed form solution to recover $\mu$. Backsubstituting into \eqref{eq:lambda_from_mu} yields the corresponding ray-depths $\vec{\lambda}$. We substitute into the cost for each of the (up to four) real solutions and take the one which minimizes the cost to get the final triangulation.

\subsection{M3. Line + VP: Triangulation with a Known 3D Direction}
One can also generate stable proposals for the weakly degenerate case when a known 3D direction is available, which can come from the vanishing point estimation. This, similarly, can also be converted into a 2D problem since we assume that the 3D endpoints lie on the two camera rays of the two endpoints $\vec{x_1^r}$ and $\vec{x_2^r}$. Specifically, take $\vec{v}$ as the 3D direction from the vanishing point, we assume that the projection of $\vec{v}$ on the plane spanned by $\vec{x_1^r}$ and $\vec{x_2^r}$ is collinear with the vector between the two endpoints, which results in Eq. (4) in the main paper. Since this equation is linear with $\vec{\lambda}$, the problem becomes a least square problem minimizing $\vec{\lambda}^TA\vec{\lambda} + \vec{b}^T\vec{\lambda}$ with a linear constraint on $\vec{\lambda}$. By introducing the Lagrange multiplier we can easily reduce the problem to a quadratic polynomial, which can be solved in closed form.

\section{Details on Point-line Association}
\label{sec::supp_association}
In this section, we present details on how we build the point-line association graphs initially in 2D and then in 3D, and further show two extensions of the recovered association graphs on generating local plane proposals and identifying the structural layout. 

Points and lines are naturally associated in 3D structures. Most salient points lie on top of the lines and the corner points mostly come from the intersection of two or more lines. However, directly discovering point-line relations in 3D is not an ideal choice because the 3D distance is always at an unknown local scale, which is ambiguous to be tested with a predefined threshold and may result in wrong association. The idea of our approach is to first associate lines with points and vanishing points in 2D, and then employ the association graphs in triangulation and joint optimization, the latter of which results in 3D association graphs as a byproduct output.

\subsection{More Details on 2D Association}
\begin{figure}[tb]
\setlength\tabcolsep{2pt} 
\begin{tabular}{cc}
{\includegraphics[trim={0 0 0 0}, clip, width=0.48\linewidth, height=80pt]{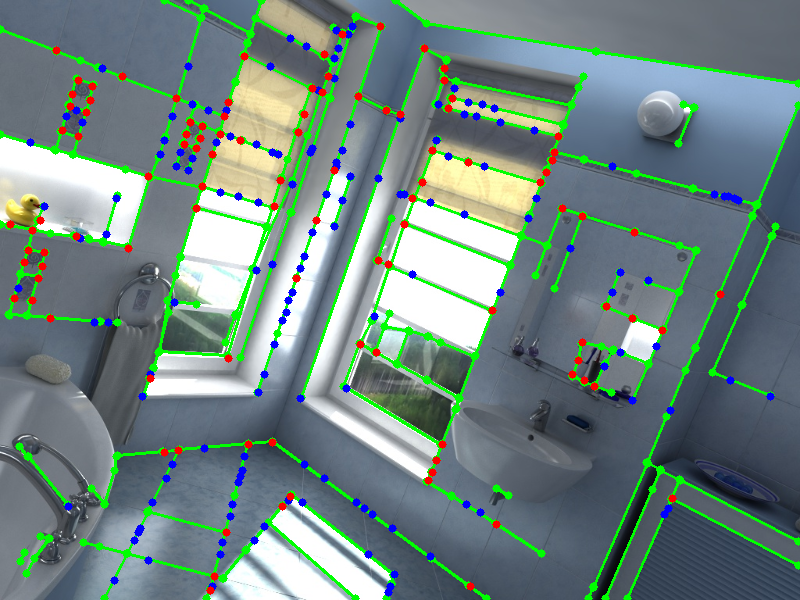}} &
{\includegraphics[trim={0 0 0 0}, clip, width=0.48\linewidth, height=80pt]{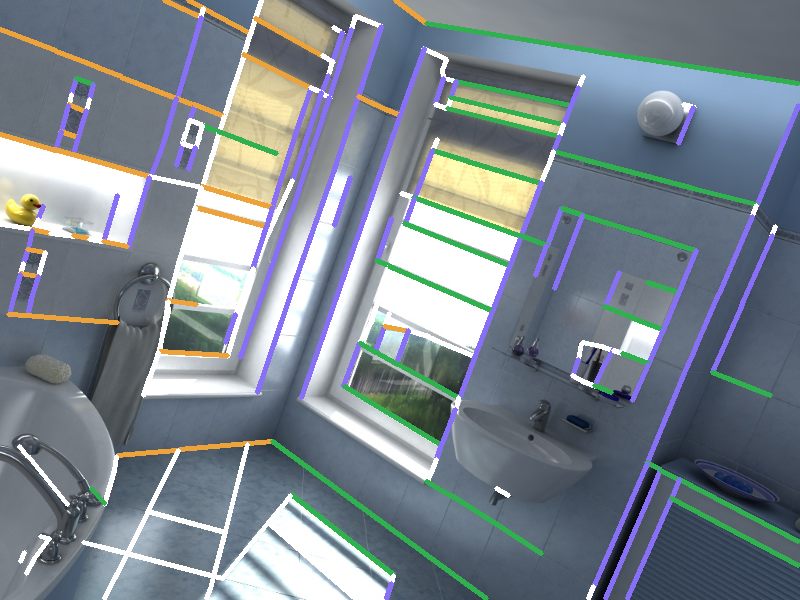}} \\
(a) 2D line-point association & (b) 2D line-VP association \\
\end{tabular}
\centering
\caption{\textbf{Visualization on the 2D association graphs} among lines and points (a) / vanishing points (VPs) (b). (a) 2D line-point association. We only show points that are associated with at least one line. The points with degree 1 are colored blue, while the points with degree $\geq 2$ are colored red. (b) 2D line-VP association. The lines that are associated with the same VP are colored the same.}
\label{fig::association_2d}
\end{figure}

First of all, we aim to recover two association graphs for each image in 2D: a line-point association graph and a line-VP association graph, each of which is a bipartite graph. By recovering the relations we can traverse the graph on each image switching back and forth between lines and points / VPs by walking along the connected edges. This can be useful for the following steps in our pipeline:

\begin{itemize}
    \item \textbf{Point-guided line triangulation}, where we can use the neighboring points and vanishing points to generate additional constraints.
    \item \textbf{Construction of 3D vanishing point tracks}, where we can associate vanishing points from different images using 2D-3D track associations in the line maps.
    \item \textbf{Joint optimization with soft association.} We can traverse the 2D association graph to measure how likely a line track is associated with a point / VP track in 3D by counting the 2D edges among their supports.
\end{itemize}

The advantages of these relational graphs are not limited to the aforementioned examples. Acting as fundamental geometric information for the sparse features, it can be beneficial to most sparse feature-based modules and applications with careful algorithmic designs. In the following parts we present details on how we build the 2D association graphs.

~\\
\noindent
\textbf{2D Line-Point Association.} The line-point association graph is built over the 2D point and line features. We employ SuperPoint \cite{detone2018superpoint} as the point feature extractor as it aims to detect corners on the image. For each pair of a 2D point and a 2D line segment, we measure the distance between them (i.e. the distance between the 2D point and the nearest point on the 2D line segment) and add an edge in the bipartite graph if the distance is less than a pixel threshold, which in our case is set to 2 pixels. In theory, the threshold should depend on the uncertainty of the line detector. An example illustration of the resulting line-point association graph is shown in Fig. \ref{fig::association_2d}(a). 

~\\
\noindent
\textbf{2D Line-VP Association.} The line-VP association graph is naturally built from the 2D vanishing point estimation. In this special bipartite graph, the line has at most one degree, while the vanishing point has at least 5 degrees to be considered valid. In our system, for vanishing point detection we use JLinkage \cite{toldo2008robust}, which aims to detect vanishing points for general parallel lines. One can also employ orthogonal vanishing point detectors \cite{bazin20123} from which the orthogonality constraints can be acquired from 2D, yet it will lose the parallelism information for lines that are not aligned with the three main orthogonal axes. An example illustration of 2D vanishing point estimation and the resulting association is shown in Fig. \ref{fig::association_2d}(b).

\subsection{More Details on 3D Association}
\subsubsection{Constructing 3D Vanishing Point Tracks}
\label{sec::supp_vp_track_construction}
We can make use of the 2D association graph to build 3D vanishing point tracks from the recovered line tracks by transitively propagating the line correspondences. 

Specifically, considering the graph with all detected vanishing points from each image as nodes, we aim to associate them together into a set of 3D vanishing point tracks. We connect two nodes from different images if:
\begin{itemize}
    \item they share at least three common neighboring line tracks on its corresponding 2D line-VP bipartite.
    \item the angle between their vanishing point directions in the global frame is less than 10 degrees. 
\end{itemize}
Once two nodes are connected, we also assign the weight of the edge to be the number of the common neighboring line tracks. Then, we sort the edges with respect to their weights in descending order and apply Kruskal-like VP track construction with the exclusion that each image only contributes one VP in a track, similar to the existing practice on constructing point tracks \cite{dusmanu2020multi}. 

\subsubsection{Joint Optimization}
As discussed in the paper, the joint optimization consists of the energy terms $E_P$, $E_L$, and $E_{PL}$. We optimize the 3D lines, points, and vanishing points jointly. Specifically, each 3D line is converted into an infinite line and parameterized with Pl\"ucker coordinate (4 DoF) as discussed in Section \ref{sec::supp_plucker}, and each 3D vanishing point is parameterized with a 3-dimensional homogeneous vector (2 DoF). The variables of the final problem exhibit $3N_P + 4N_L + 2N_{VP}$ degrees of freedom in total, where $N_P$, $N_L$, and $N_{VP}$ are the number of 3D points, lines, and vanishing points, respectively. In the following, we will discuss the three energy terms in detail.

~\\
\noindent
\textbf{$E_P$. Data term for the point tracks.} This term is defined as the squared reprojection error for each point track, which is exactly the same as in the regular bundle adjustment in COLMAP \cite{schonberger2016structure}. 

~\\
\noindent
\textbf{$E_L$. Data term for the line tracks. } This term is defined as the reprojection error for each line track, termed geometric refinement in the main paper. For the line-only solutions in the experiments, we only employ this term in the final optimization solely over 3D line tracks. The residual is formulated as:

\begin{equation}
    E_L(l) = \sum_k w_{\angle}^2(L_k, \ell_k)~\cdot~e_{\text{perp}}^2(L_k, \ell_k),
\end{equation}
\begin{equation}
    w_{\angle}(L_k, \ell_k) = \exp(\alpha (1 - \cos(\angle(L_k, \ell_k)))),
\end{equation}

where $\alpha$ equals 10.0 in our system. This weighting term $w_{\angle}$ follows the design of L3D++ \cite{hofer2017efficient}, which empirically promotes fast convergence by putting the emphasis to make the 2D direction consistent with the observation. Note that compared to $1 - \cos(\angle(L_k, \ell_k))$ in our formulation, the residual in L3D++ \cite{hofer2017efficient} is directly built on the angle $\angle(L_k, \ell_k)$, which exhibits singularity on the gradient at zero angles and sometimes results in unstable optimization. 

~\\
\noindent
\textbf{$E_{PL}$. 3D Association Term. } This term encourages 3D association among lines and points / vanishing points. Specifically, it consists of three parts: line-point association, line-VP association, and VP orthogonality regularization.

\begin{itemize}
    \item \textbf{3D Line-Point Association.} As discussed in the main paper, the association term is built between each pair of point track and line track that has at least three connected edges (on the 2D line-point association graphs from the corresponding images) among their 2D supports. Each residual is defined as the 3D point-line distance weighted by the number of 2D connections among supports, which can be efficiently computed with Pl\"ucker coordinate as discussed in Section \ref{sec::supp_plucker_pointtoline}.
    \item \textbf{3D Line-VP Association. } As in the point case, the association term is built between each pair of VP tracks (built as in Section \ref{sec::supp_vp_track_construction}) and line track that has at least three connected edges (on the 2D line-point association graphs from the corresponding images) among their 2D supports. Each residual is defined as the sine of the direction angle between the line and the vanishing point, again weighted by the 2D connections among supports.
    \item \textbf{VP Orthogonality Regularization.} Since we do not employ orthogonal vanishing point detection \cite{bazin20123} to ensure the generality of the system, we do not have any orthogonal information from 2D. However, at joint optimization, we can enforce the nearly orthogonal pairs (in practice, when the angle is larger than 87 degrees) of vanishing points to be orthogonal. So we add a regularization residual to these pairs, defined as the cosine of the angle difference between the directions of the vanishing point pair.
\end{itemize}

Note that in the joint optimization we do not have data term for the 3D vanishing point. This means that we only employ the 2D line-VP association graphs and enforce parallelism with soft association, without relying on the actual vanishing point detection on 2D which can be sometimes noisy. In this way, we only enforce the lines that are associated with the same VP to become parallel. Both the line-point and line-VP association residuals are equipped with Huber loss function from Ceres Solver \cite{ceres} to ensure robustness to outlier edges. 

From the joint optimization, we can directly get the 3D association graphs as a byproduct output, by testing the validity of the active line-point / line-VP edges in the soft association problem. The validity check measures the 3D point-line distance and the 3D direction angle ($\leq$ 5 degrees) respectively. For the validity check of the 3D point-line distance, we keep a fixed threshold of 2.0 and re-scale the distance with the minimum uncertainty (defined as the depth divided by the focal length) between the measured 3D point and 3D line to ensure scale invariance. For the output 3D point-line association graph this step removes the outlier edges that are filtered out in the soft association problem at joint optimization. Note that both the resulting 3D association graphs are again bipartite graphs, among lines and points / vanishing points.

\subsection{Extension: Generating Local Plane Proposals}

\begin{figure}[tb]
\begin{tabular}{cc}
{\includegraphics[trim={485 170 480 100}, clip, width=0.48\linewidth, height=100pt]{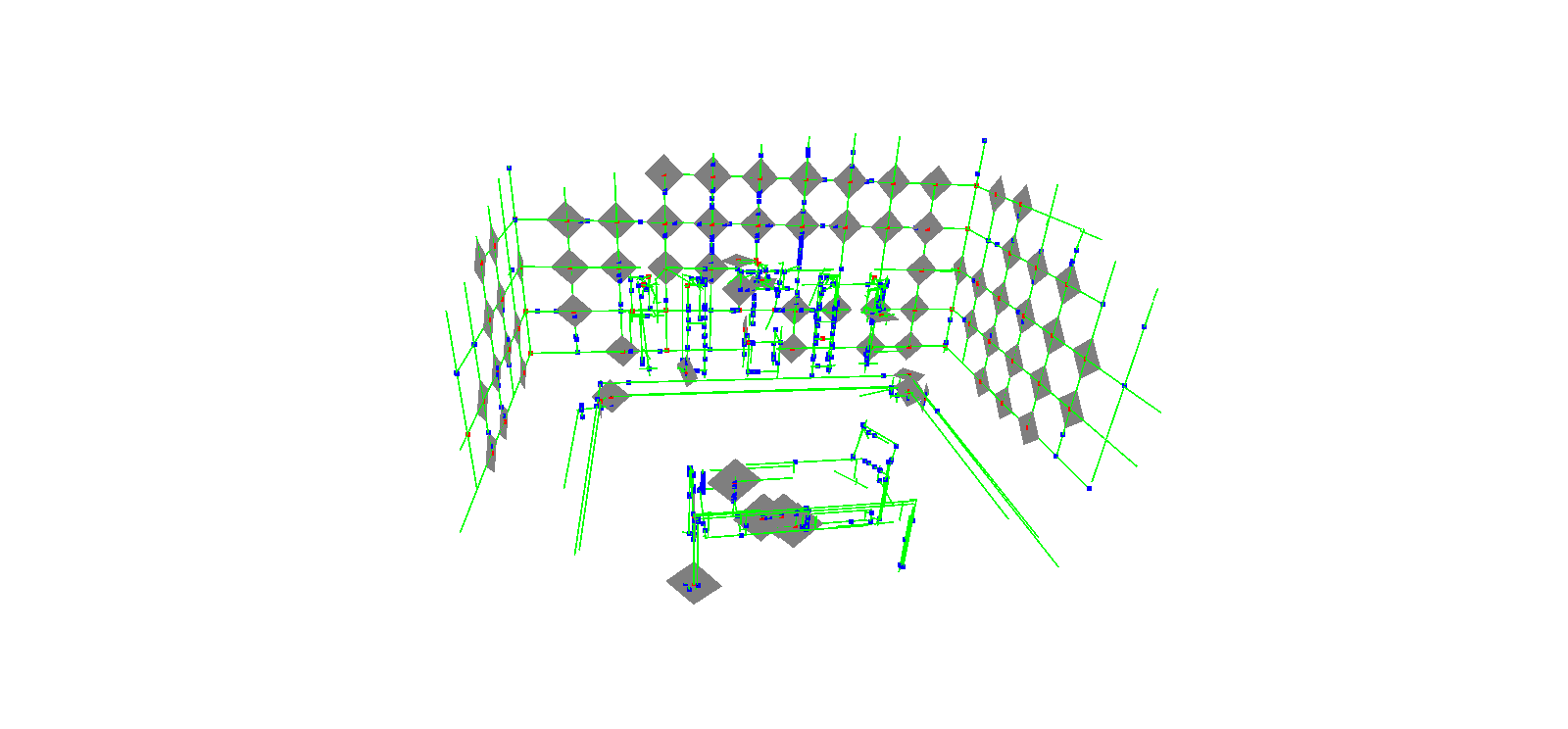}} &
{\includegraphics[trim={220 130 150 100}, clip, width=0.48\linewidth, height=100pt]{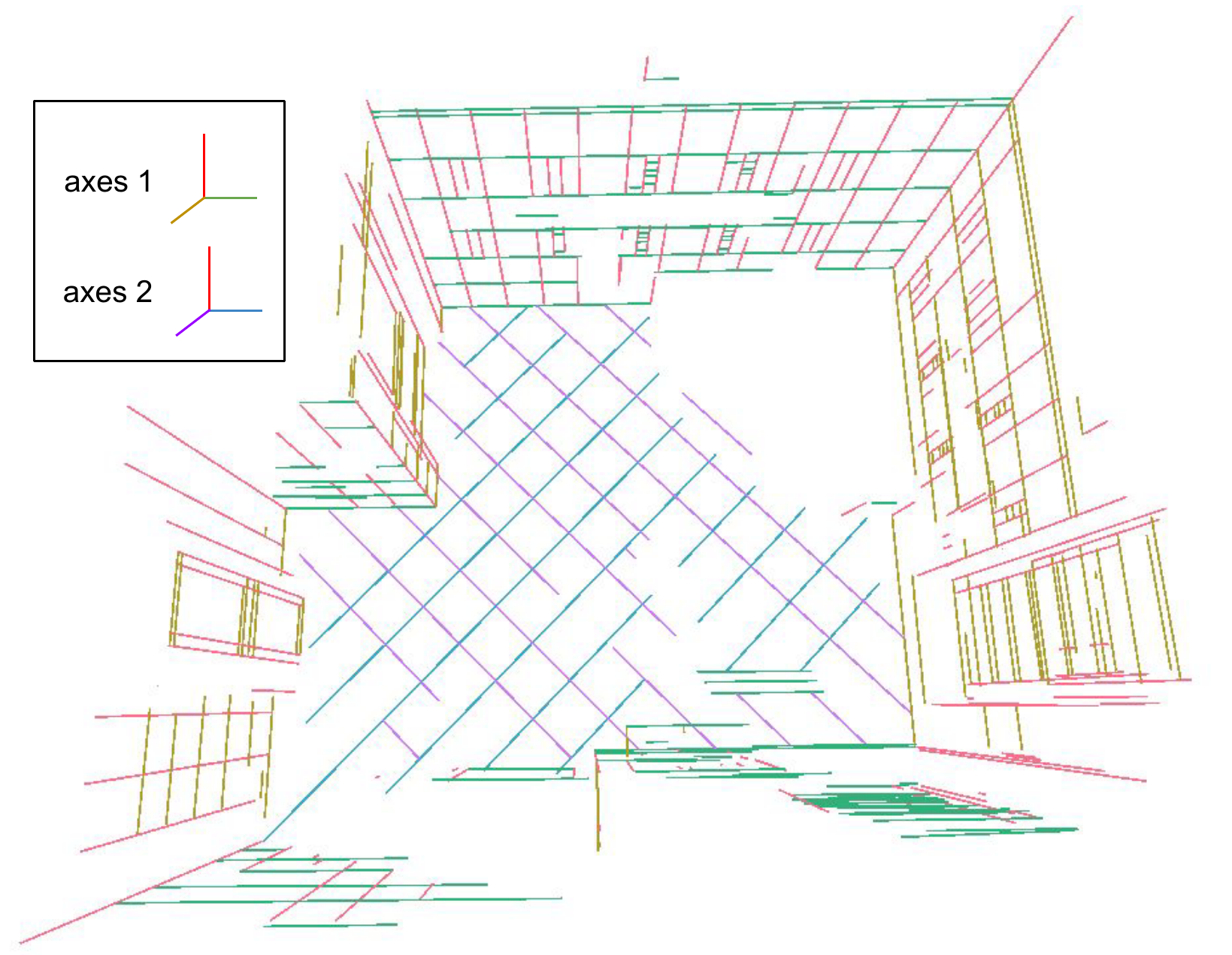}} \\
(a) Local plane proposals & (b) Atlanta World \\
\end{tabular}
\centering
\caption{\textbf{Illustration of the extensions on the recovered 3D association graphs.} (a) Local plane proposals can be directly acquired from the degree-2 junctions. (b) We show the recovered orthogonal axes from joint optimization at the top left of the figure, where two different Manhattan axes \cite{coughlan2000manhattan} are discovered, resulting in an Atlanta World \cite{schindler2004atlanta}. }
\label{fig::supp_association_extensions}
\end{figure}

As a byproduct output of the system, the 3D line-point association graph can be easily extended to benefit high-level problems. We show one most straightforward extension on generating local plane proposals in Fig. \ref{fig::supp_association_extensions}(a). From each degree-2 3D point in the graph, we can compute the plane normal by applying the cross product on its two neighboring 3D lines. From the resulting local planes, one can easily group the plane structures and further recover the scene layout. These planes are also potentially beneficial for visual localization pipelines.

\subsection{Extension: Atlanta World}

From the recovered orthogonality relationship we can easily parse the high-level structure relations in the output 3D maps. Figure \ref{fig::supp_association_extensions}(b) shows an example: From the six recovered orthogonality constraints we can get two groups of orthogonal axes with a shared vertical direction, resulting in an Atlanta World as discussed in \cite{schindler2004atlanta}.

\section{Covariance Derivation and Setup for Synthetic Tests}
\label{sec::supp_covariance}
In this section, we will provide detailed covariance derivations for endpoint triangulation and algebraic line triangulation respectively, which are used to compute the uncertainty (the largest eigenvalue of the covariance) to study the degeneracy problem in the main paper. We also provide details of the setup of the synthetic tests that are discussed in Figure 8 in the main paper.

In this section, we follow the convention of Section \ref{sec::supp_triangulation}, where the world coordinate is aligned with the reference frame without loss of generality, and the camera pose of the matched image is $(R, \vec{t})$. The intrinsic-normalized homogeneous coordinates for the endpoints of the reference segment are $\vec{x}_1^r$ and $\vec{x}_2^r$, and $\vec{x}_1^m$ and $\vec{x}_2^m$ for the endpoints of the matched segment.

\subsection{Background: Covariance Propagation}
Denote the 2D endpoints correspond to $\vec{x}_1^r$, $\vec{x}_2^r$, $\vec{x}_1^m$, and $\vec{x}_2^m$ as $\vec{p}_1^r$, $\vec{p}_2^r$, $\vec{p}_1^m$, and $\vec{p}_2^m$ respectively. The input for the triangulation problem is an 8-dimensional vector $\vec{p}_{2D}$, while the output for the triangulation problem is a 6-dimension vector $\vec{p}_{3D}$, where

\begin{equation}
    \vec{p}_{2D} = 
    \begin{pmatrix}
    \vec{p}_1^r \\
    \vec{p}_2^r \\
    \vec{p}_1^m \\
    \vec{p}_2^m
    \end{pmatrix}, 
    \ \ 
    \vec{p}_{3D} = 
    \begin{pmatrix}
    \vec{X}_1^r \\
    \vec{X}_2^r
    \end{pmatrix}.
\end{equation}

We can assume unit covariance on each endpoint of the line detection and independence across different endpoints. Then the $8 \times 8$ covariance matrix of $\vec{p}_{2D}$ can be written as:

\begin{equation}
    \vec{\Sigma}_{2D} =
    \begin{pmatrix}
    I_2 & 0 & 0 & 0 \\
    0 & I_2 & 0 & 0 \\
    0 & 0 & I_2 & 0 \\
    0 & 0 & 0 & I_2
    \end{pmatrix} = I_8
\end{equation}

Take $f_e$ and $f_l$ be the functions that maps $\vec{p}_{2D}$ to $\vec{p}_{3D}$ with endpoint triangulation and algebraic line triangulation respectively, and $J_e$, $J_l$ be their corresponding $6\times 8$ Jacobian matrices.

With the rule of covariance propagation, we can compute the covariance $\Sigma_{3D}^e$ and $\Sigma_{3D}^l$ for endpoint triangulation and algebraic line triangulation by

\begin{equation}
    \Sigma_{3D}^e = J_e\Sigma_{2D}J_e^T, 
    \ \ 
    \Sigma_{3D}^l = J_l\Sigma_{2D}J_l^T, \ \ .
\end{equation}

The problem of measuring the covariance of the triangulated 3D line segment reduces to the computation of the corresponding Jacobian matrices $J_e$ and $J_l$.

We first provide here the Jacobian $J_d$ of a normalized direction vector $\vec{d}$ with respect to $\vec{x}$, where $\vec{d}=\frac{\vec{x}}{\lVert \vec{x} \rVert}$:

\begin{equation} \label{eq::direction_gradient}
    J_d = \frac{I - \vec{d}\vec{d^T}}{\lVert \vec{x} \rVert}.
\end{equation}

For ease of notation in the following sections, we use $\vec{d}_i^r$ and $\vec{d}_i^m$ ($i=1, 2$) to denote the normalized ray directions correspond to $\vec{x}_i^r$ and $R^T\vec{x}_i^m$ respectively, and $\vec{C}_r=\vec{0}$, $\vec{C}_m=-R^T\vec{t}$ be the corresponding camera center.

\subsection{Endpoint Triangulation} \label{sec::supp_covariance_endpoint_triangulation}
We employ the mid-point triangulation \cite{hartley2003multiple} for computing each of the 3D endpoints respectively. The formulation of midpoint triangulation can be written as follows:

\begin{equation}
    \begin{pmatrix}
    1 & -{\vec{d}_i^r}^T\vec{d}_i^m \\
    -{\vec{d}_i^r}^T\vec{d}_i^m & 1
    \end{pmatrix}
    \begin{pmatrix}
    \lambda_i^r \\
    \lambda_i^m
    \end{pmatrix} = 
    \begin{pmatrix}
    {\vec{d}_i^r}^T(C_m - C_r) \\
    -{\vec{d}_i^m}^T(C_m - C_r)
    \end{pmatrix},
\end{equation}

\begin{equation}
    \vec{X}_i = \frac{1}{2}(\vec{C}_r + \lambda_i^r\vec{d}_i^r + \vec{C}_m + \lambda_i^m\vec{d}_i^m),
\end{equation}

where $\lambda_i^r$ and $\lambda_i^m$ are the ray depths of the rays $\vec{d}_i^r$ and $\vec{d}_i^m$ respectively ($i=1, 2$). 

By using the property on the derivative of matrix inverse:
\begin{equation} \label{eq::matrix_inverse_derivative}
(K^{-1})' = -K^{-1}K'K^{-1},    
\end{equation}
we can compute the derivative of $\vec{X}_i$ with respect to the direction vectors $\vec{d}_i^r$ and $\vec{d}_i^m$. Combining \eqref{eq::direction_gradient} the final $6 \times 8$ Jacobian $J_e$ can be computed with the chain rule.

\subsection{Algebraic Line Triangulation} \label{sec::supp_covariance_line_triangulation}
Similar to the endpoint triangulation, we can also formulate a linear system with respect to the direction vectors for algebraic line triangulation. Specifically, we can rewrite \eqref{eq::line_triangulation_3x3} into the following matrix form over $\vec{d}_i^r$, $\vec{d}_1^m$ and $\vec{d}_2^m$:

\begin{equation}
    \begin{pmatrix}
    \vec{d}_i^r & -\vec{d}_1^m & -\vec{d}_2^m
    \end{pmatrix}
    \begin{pmatrix}
    \lambda_i^r \\
    \beta_1^m \\
    \beta_2^m
    \end{pmatrix} = \vec{C}_m - \vec{C}_r,
\end{equation}

\begin{equation}
    \vec{X}_i = \vec{C}_r + \lambda_i^r \vec{d}_i^r, \ \ i = 1, 2
\end{equation}

Again by using \eqref{eq::direction_gradient} and \eqref{eq::matrix_inverse_derivative} the final $6 \times 8$ Jacobian $J_l$ can be computed accordingly.

\subsection{Setup for Synthetic Tests}

Based on the derived covariance forms, we present study on the degeneracy problem of line triangulation in the main paper (Figure 8). We here discuss the detailed setup for the two experiments.

~\\
\noindent
\textbf{Uncertainty Visualization on AdelaideRMF \cite{AdelaideRMF}. } We take an image pair from AdelaideRMF \cite{AdelaideRMF} and compute its two-view geometry with COLMAP \cite{schonberger2016structure}. Then, we manually annotate 42 line pairs that are perfectly matched. On top of the annotated line matches we perform algebraic line triangulation and compute the uncertainty as the largest eigenvalue of the covariance matrix. We also visualize the epipolar lines on the target image to better illustrate the relation to the degeneracy problem. When visualized in 3D, the lines with low uncertainty are reasonably accurate while the lines that are degenerate (with high covariance) locate ``everywhere" in the 3D space. This also shows that the largest eigenvalue of the covariance matrix can be a good indicator of the reliability of the triangulation. 

~\\
\noindent
\textbf{Synthetic Tests. } We design a synthetic test to further study the stability of the line triangulation. Specifically, we first set up a horizontal plane ($z = 10$) and two stereo cameras that point orthogonal to the plane with a distance of 10.0. The baseline (lying along the x direction) of the stereo pair is 4.0 (so the two cameras locate at $(-2, 0, 0)$ and $(2, 0, 0)$) and the focal lengths of both cameras are 700. Under this setup, the epipolar lines are always horizontally aligned with the stereo baseline. We sample random 3D lines with a fixed direction on the horizontal plane ($z=10$) within a range of $[-1, 1]$ on both x and y directions, and project them onto the two views, resulting in perfect 2D line matches with a fixed angle with the epipolar lines. Then, we perform endpoint triangulation and algebraic line triangulation respectively, and compute its covariance as discussed in Section \ref{sec::supp_covariance_endpoint_triangulation} and \ref{sec::supp_covariance_line_triangulation}. We measure the uncertainty as the largest eigenvalue of the covariance matrix. The median uncertainty is computed for 10000 random lines for each tested angle.

\section{Line Reconstruction given Depth Maps}
\label{sec::supp_fitnmerge}
\subsection{System Details}

As discussed in the paper, when depth maps are available (e.g. from an RGB-D sensor), we can apply a robust fitting to the back-projected 3D points to generate an accurate proposal, which can serve as the best candidate for the 2D line segment in the track building step. 

Specifically, we sample points along the 2D line and collect a set of 3D points by back-projecting the points using the depth maps. Then, we apply 3D line fitting with LO-RANSAC \cite{chum2003locally,Lebeda2012BMVC}. To ensure invariance to the scale changes, the inlier threshold is proportional to the median depth of all the points divided by the focal length, which shares similar spirits with the scale factor $\sigma$ used in the \textit{InnerSeg distance}. By associating those fitted 3D line segments with the same track-building strategy, we can acquire high-quality line tracks that align geometrically with the 3D depth maps.

\begin{table}[tb]
\begin{center}
\scriptsize
\setlength{\tabcolsep}{2.5pt}
\begin{tabular}{clccccccc}
\toprule
Line type & Method & R1 & R5 & R10 & P1 & P5 & P10 & \# supports \\
\midrule
\multirow{2}{*}{\makecell{LSD\\\cite{von2008lsd}}} 
& Ours (line-only) & 48.3 & 187.0 & 257.4 & 59.2 & 81.9 & 89.8 & (15.8 / 19.1) \\
& Ours w/ depth & \textbf{89.7} & \textbf{315.3} & \textbf{330.8} & \textbf{63.0} & \textbf{99.7} & \textbf{100} & (\textbf{16.6} / \textbf{23.3}) \\
\midrule
\multirow{2}{*}{\makecell{SOLD2\\ \cite{pautrat2021sold2}}} 
& Ours (line-only) & 50.8 & 143.5 & 180.8 & 74.4 & 86.9 & 91.2 & (15.1 / 32.2) \\
& Ours w/ depth & \textbf{84.4} & \textbf{252.0} & \textbf{278.2} & \textbf{79.7} & \textbf{99.7} & \textbf{99.9} & (\textbf{16.0} / \textbf{38.4}) \\
\bottomrule
\end{tabular}
\caption{\textbf{Quantitative results of line mapping given depth maps} on Hypersim \cite{roberts:2021}. $R\tau$ and $P\tau$ are reported at 1mm, 5mm, 10 mm along with the average number of supporting images/lines.}
\label{tab::main-hypersim-fitnmerge}
\end{center}
\end{table}

\begin{figure}[tb]
\scriptsize
\setlength\tabcolsep{2pt} 
\begin{tabular}{cccccccc}
{\includegraphics[width=0.15\linewidth, height=15pt]{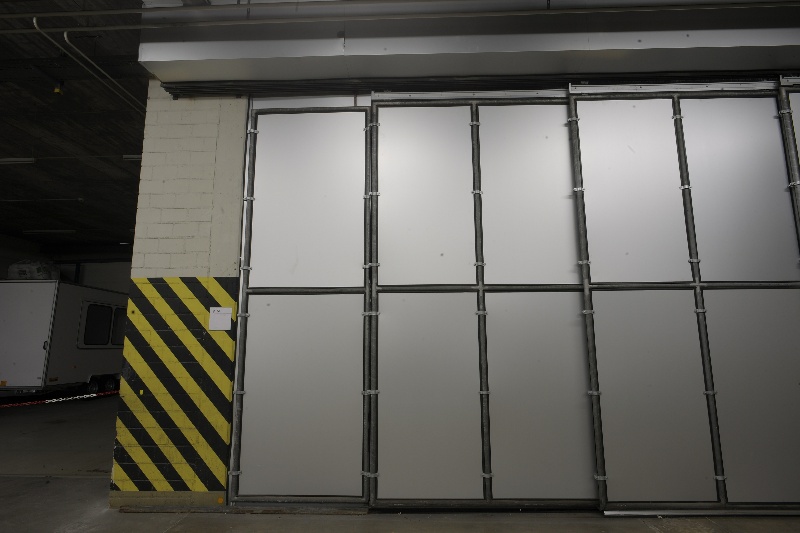}} & 
{\includegraphics[width=0.15\linewidth, height=15pt]{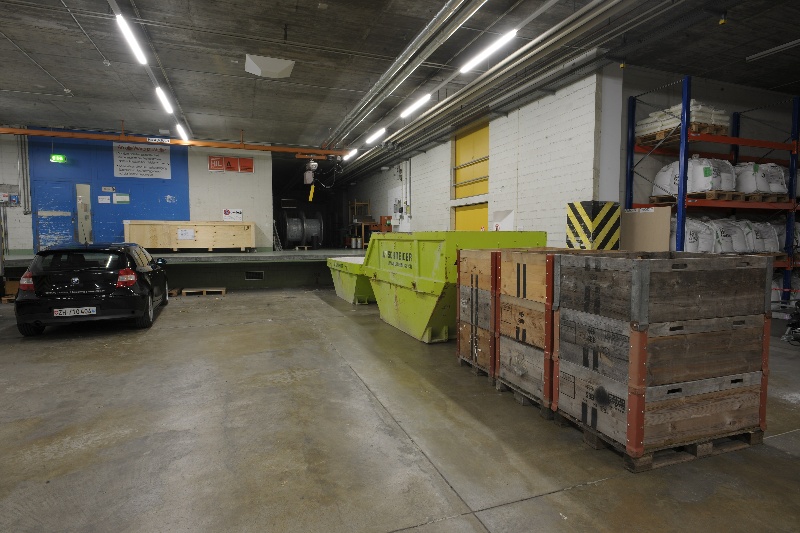}} &
{\includegraphics[width=0.15\linewidth, height=15pt]{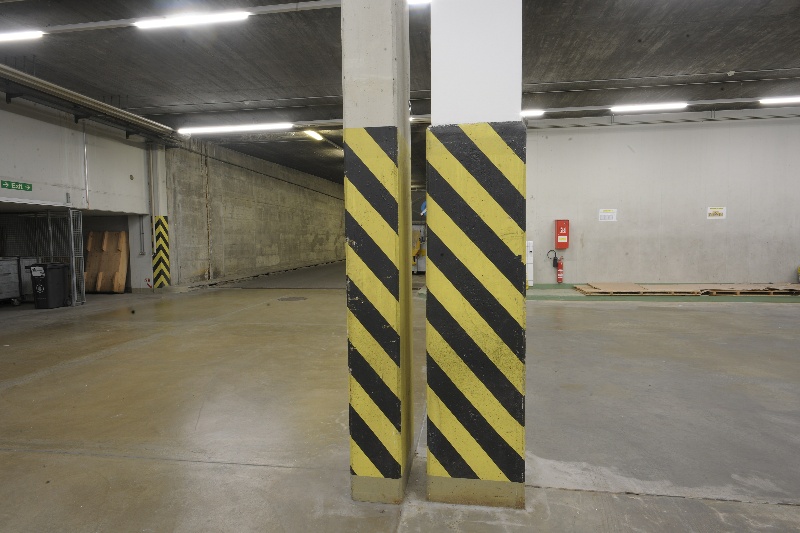}} & & &
{\includegraphics[width=0.15\linewidth, height=15pt]{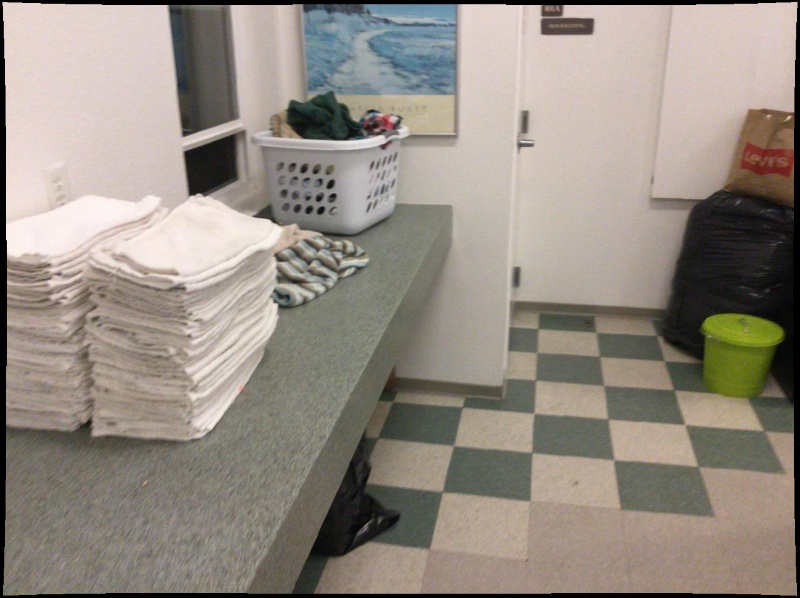}} &
{\includegraphics[width=0.15\linewidth, height=15pt]{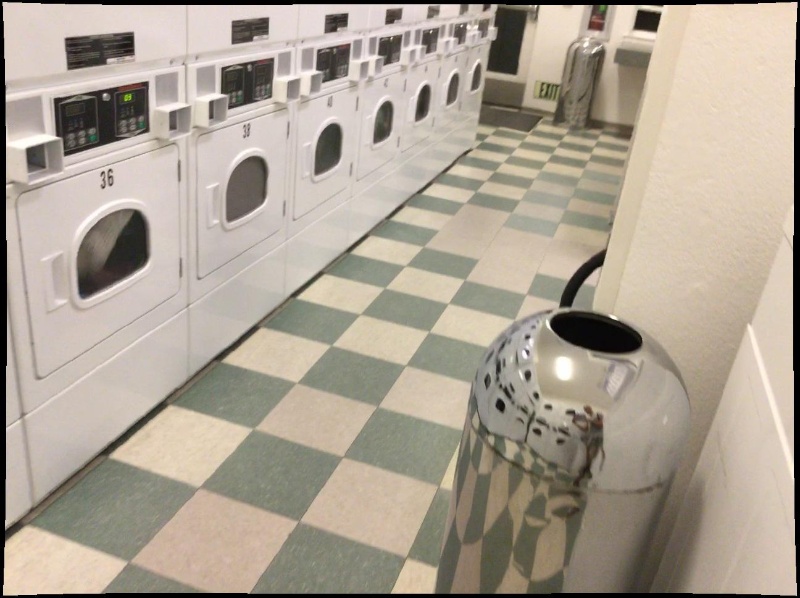}} &
{\includegraphics[width=0.15\linewidth, height=15pt]{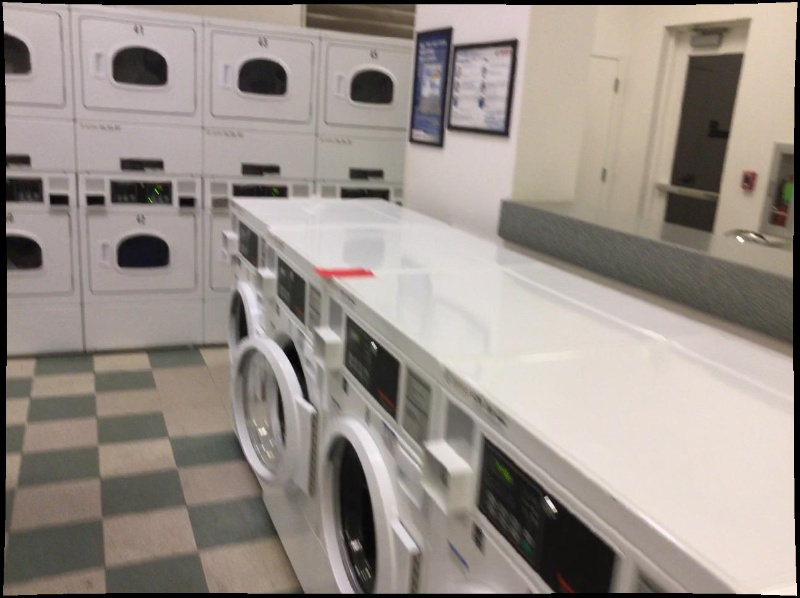}} \\
\end{tabular}
\begin{tabular}{cc}
{\includegraphics[trim={0 0 0 0}, clip, width=0.48\linewidth, height=80pt]{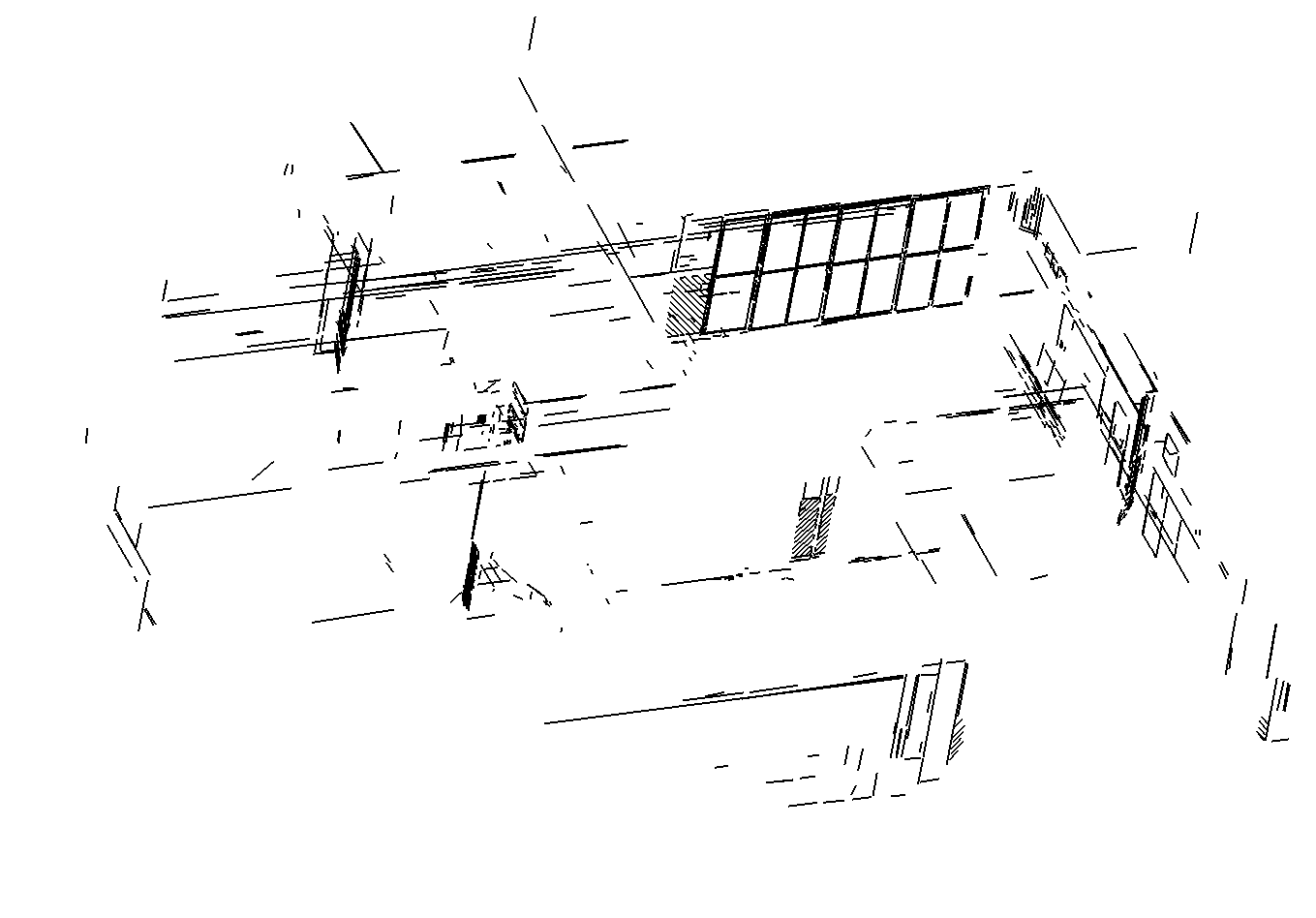}} &
{\includegraphics[trim={0 0 0 0}, clip, width=0.48\linewidth, height=80pt]{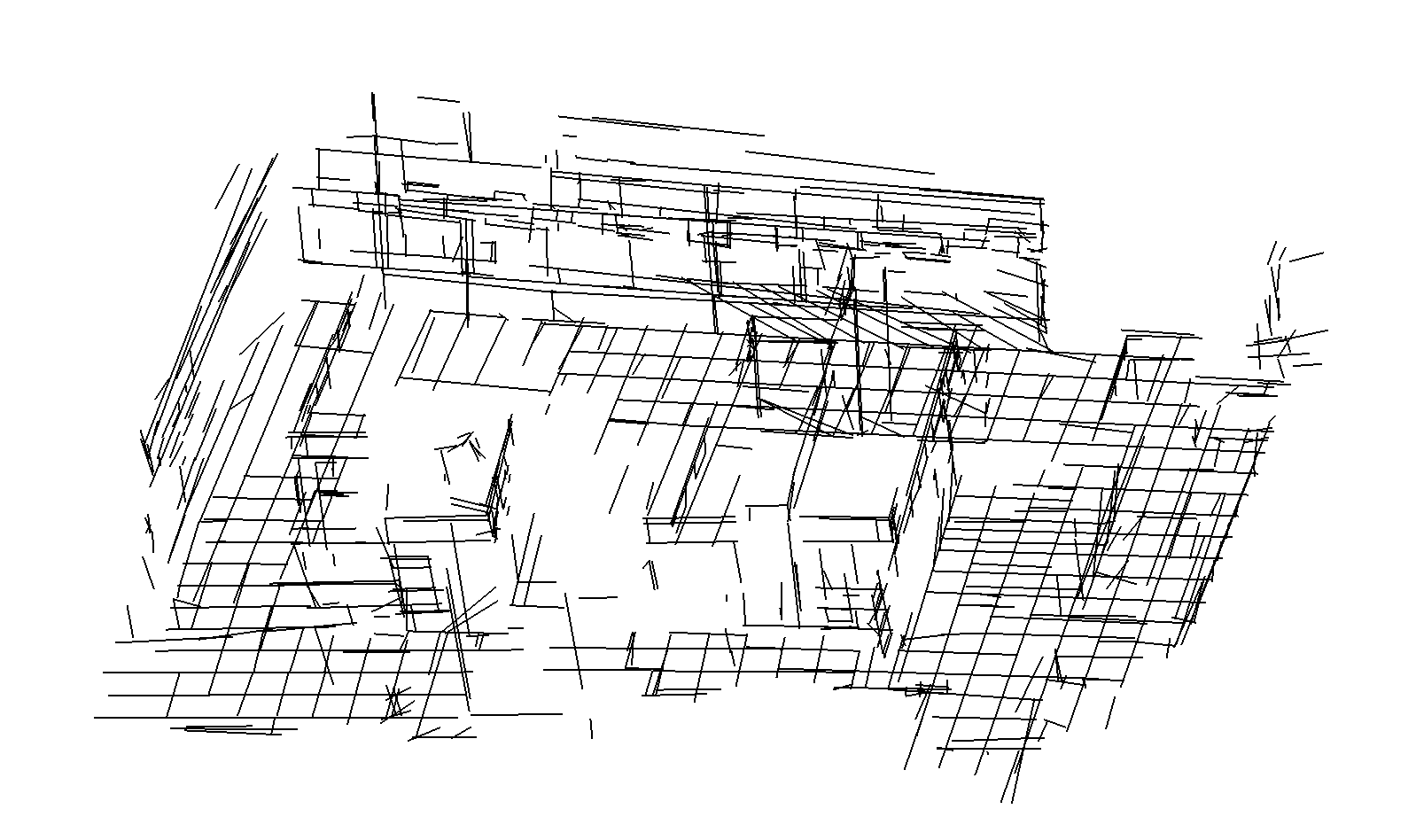}} \\
\end{tabular}
\centering
\caption{\textbf{Line mapping given depth maps.} \textbf{Left:} DSLR sequence of \textit{delivery\_area} (44 images) from ETH3D \cite{schops2017multi,schops2019bad} with LSD \cite{von2008lsd} and LiDAR scanner depth. \textbf{Right:} Laundry (\textit{scene\_0678\_01}, 465 images at 6 FPS) from ScanNet \cite{dai2017scannet} with SOLD2 \cite{pautrat2021sold2} and depth from an RGB-D sensor.}
\label{fig::supp_fitnmerge}
\end{figure}

\begin{table}[tb]
    \centering
    \scriptsize
    \setlength{\tabcolsep}{3pt}
    \begin{tabular}{lccc}
        \toprule
        Scene & HLoc\cite{hloc} w/ Depth & PtLine\cite{gao2022pose} & Ours w/ Depth \\
        \midrule
        Chess & \textbf{2.4} / \textbf{0.81} / 94.8 & \textbf{2.4} / \textbf{0.81} / \textbf{95.0} & \textbf{2.4} / 0.82 / 94.0 \\ 
        Fire & 1.9 / 0.76 / 96.4 & 1.9 / 0.76 / \textbf{96.6} &  \textbf{1.7} / \textbf{0.71} / \textbf{96.6} \\
        Heads & 1.1 / 0.73 / 99.0 & 1.1 / 0.74 / \textbf{99.4} & \textbf{1.0} / \textbf{0.72} / \textbf{99.4} \\
        Office & 2.7 / 0.83 / 83.7 & 2.7 / 0.83 / 83.9 & \textbf{2.6} / \textbf{0.80} / \textbf{84.7} \\
        Pumpkin & 4.1 / \textbf{1.05} / \textbf{61.3} & 4.0 / 1.06 / 60.8 & \textbf{4.0} / \textbf{1.05} / 61.1 \\
        Redkitchen & 3.3 / \textbf{1.12} / 72.1 & \textbf{3.2} / \textbf{1.12} / 72.5 & 3.3 / \textbf{1.12} / \textbf{73.0} \\
        Stairs & 4.7 / 1.25 / 53.4 & 4.3 / 1.16 / 55.9 & \textbf{3.2} / \textbf{0.86} / \textbf{76.0} \\  
        \midrule
        Avg. & 2.9 / 0.94 / 80.1 & 2.8 / 0.93 / 80.6 & \textbf{2.6} / \textbf{0.87} / \textbf{83.5} \\
        \bottomrule
    \end{tabular}
    \caption{\textbf{Visual localization on 7Scenes~\cite{7scenes} with depth maps~\cite{7scenes-depth}.} We report the median translation and rotation error in cm and degrees, as well as the pose accuracy at a 5 cm / 5 deg threshold.}
    \label{tab:7scenes_fitnmerge}
\end{table}

\subsection{Results on Line Mapping}
We show quantitative results on line mapping given depth maps on the first eight scenes of Hypersim \cite{roberts:2021} in Table \ref{tab::main-hypersim-fitnmerge}. As our solution on line mapping given depth maps does not employ points and vanishing points, we show the comparison to our triangulation with only line-line proposals. While still far from perfect, the resulting 3D line maps are significantly better compared to the ones built without depth maps. Nearly all the recovered 3D lines are within 10 millimeters of the ground truth mesh model. The track supports are also significantly richer in comparison. Nevertheless, it is worth noting that our line maps built with the assistance of points and vanishing points can achieve a comparable number of supports with SOLD2 line detector \cite{pautrat2021sold2}. This again demonstrates the advantages of point-guided line triangulation in being able to generate reasonable proposals on degenerate cases, which benefits the track-building process.

We further show qualitative results of line mapping with depth maps in Figure \ref{fig::supp_fitnmerge}, on ETH3D \cite{schops2017multi} and ScanNet \cite{dai2017scannet} respectively. The mapping solution can produce perceivable 3D line structures with either the LiDAR scanner depth from ETH3D \cite{schops2017multi} or the RGB-D sensor from ScanNet \cite{dai2017scannet}, with both conventional LSD detector \cite{von2008lsd} and the recent learning-based one \cite{pautrat2021sold2}. 

\subsection{Results on Line-Assisted Visual Localization}
As in the RGB case, we show here that line mapping with depth maps is also able to help visual localization by combining points and lines. We run our solution of line mapping with depth maps on 7Scenes \cite{7scenes} with depth maps from \cite{7scenes-depth}. Then, we run our proposed point-line visual localization system with hybrid RANSAC \cite{camposeco2018hybrid} as discussed in the paper (detailed in Sec. \ref{sec::supp_localization}). Results are shown in Table \ref{tab:7scenes_fitnmerge}. Integrating line maps into visual localization improves the localization accuracy, in particular contributing to a large performance gain (53.4 $\rightarrow$ 76.0 on 5 cm / 5 deg) on the most challenging scene: \textit{Stairs}. This again demonstrates the usefulness of the acquired 3D line maps.

\section{More Implementation Details}
\label{sec::supp_implementation_details}

\subsection{Datasets}

We test our method quantitatively on Hypersim \cite{roberts:2021} and \textit{Tanks and Temples} \cite{Knapitsch2017}. For the qualitative results across datasets, we rely on two line detectors: SOLD2 and LSD \cite{von2008lsd}. For SOLD2 \cite{pautrat2021sold2} detection, description, and matching, we employ the default parameters provided in their released code repository. As SOLD2 \cite{pautrat2021sold2} is originally focused on indoor and manmade structured environments, we test SOLD2 detection only on indoor datasets: Hypersim \cite{roberts:2021} and ScanNet \cite{dai2017scannet} (Figure \ref{fig::supp_fitnmerge} in supp.), while for the other datasets LSD \cite{von2008lsd} is used to run our line mapping. For all the datasets, we undistort images with the calibration (either provided or estimated with COLMAP \cite{schonberger2016structure}) before performing line detection, which mitigates the issue of straight lines appearing curved due to radial distortion.

\textbf{Hypersim} \cite{roberts:2021} is a photorealistic synthetic dataset for holistic indoor scene understanding. For evaluation, we use the first 8 scenes and resize the image to a maximum dimension of 800 as input to all the tested methods. The average metrics over all the 8 scenes are reported. For neighborhood computation, we use the point triangulator from COLMAP \cite{schonberger2016structure} with SuperPoint \cite{detone2018superpoint} from HLoc \cite{hloc} to build the 3D model from images with known camera poses, and rank neighboring images by the Dice coefficient on the common 3D points. The reconstructed 3D lines are evaluated with respect to the provided ground truth mesh model. To efficiently compute the distance between a query point sampled from the line and the mesh, we use the AABB hierarchy from \cite{libigl} as the data structure. Specifically, we first build the hierarchy from the mesh model and sample points densely and uniformly for each line to compute the overall length recall $R\tau$ and the inlier percentage $P\tau$. 

\textbf{\textit{Tanks and Temples}} \cite{Knapitsch2017} is a benchmark for image-based 3D reconstruction and is widely used for evaluating multi-view stereo and novel view synthesis \cite{riegler2021stable}. They provide dense ground truth point cloud from a FARO Focus 3D X330 HDR scanner. We input the images with their original resolution (around 2 Megapixels) for both our method and L3D++ \cite{hofer2017efficient}. The provided point cloud is cleaned such that it only contains the main subject in the middle of the scene. Since our method can reconstruct lines that are far away from the model (see Figure \ref{fig::scale_invariance} and Figure 5 \textit{Barn} of the main paper), we compute the axis-aligned bounding box for each point cloud and stretch it one meter in all three dimensions. At evaluation, only lines within the stretched bounding box are considered. We evaluate on the \textit{train} split and take the average over all the scenes except for \textit{Ignatius} which has no observable line structures. For efficient computation of the distance between the sampled point and the ground truth point cloud, we build a KD-Tree over the ground truth point cloud with nanoflann \cite{blanco2014nanoflann}. 

To further demonstrate the effectiveness and generalization of our system, we also present qualitative results on unstructured image collections on Aachen v1.1 Day-Night dataset \cite{sattler2018benchmarking} and Rome city from BigSFM \cite{snavely2006photo,snavely2008modeling,agarwal2011building}. In the supplementary material, we also present additional results of our line mapping on Cambridge \cite{kendall2017geometric}, PhotoTourism \cite{snavely2006photo,jin2021image}, and also line mapping given depth maps on ETH3D \cite{schops2017multi,schops2019bad} and ScanNet \cite{dai2017scannet}. The oracle test in Figure 8 of the main paper is conducted on AdelaideRMF dataset \cite{AdelaideRMF}. The localization experiments are run on Cambridge \cite{kendall2015posenet}, 7Scenes \cite{7scenes, 7scenes-depth}, and InLoc \cite{taira2018inloc} (in this supplementary material).

\subsection{Hyperparameters}
Similar to all existing point-based solutions such as COLMAP \cite{schonberger2016structure}, our library also has a number of hyperparameters that can be changed in each module while using our default ones at release should work on most in-the-wild cases due to our scale-invariant design. We keep the hyperparameters unchanged throughout all experiments across datasets. 

The scaling factors introduced at scoring and track building are set as follows: $\tau_a = 10$ degrees for the angle in 3D and $\tau_a = 8$ degrees in 2D, $\tau_o = 0.05$ for the overlap in 2D and 3D, $\tau_p = 5$ pixels for the perpendicular distance in 2D, and $\tau_s = 0.015$ for the scale-invariant endpoint distance in 3D. The threshold for 2D point-line association is set to 2 pixels, and the inlier threshold for 2D VP-line association with JLinkage \cite{toldo2008robust} is set to 1 pixel. The detected vanishing points with at least 5 inliers (associated lines) are kept on each image. The minimum triangulation angle between the camera ray and the plane spanned by camera rays $\vec{x}_1^m$ and $\vec{x}_2^m$ is 1 degree.

\begin{figure}[tb]
\small
\setlength\tabcolsep{6pt} 
\begin{tabular}{cc}
\includegraphics[width=0.33\columnwidth, height=75pt]{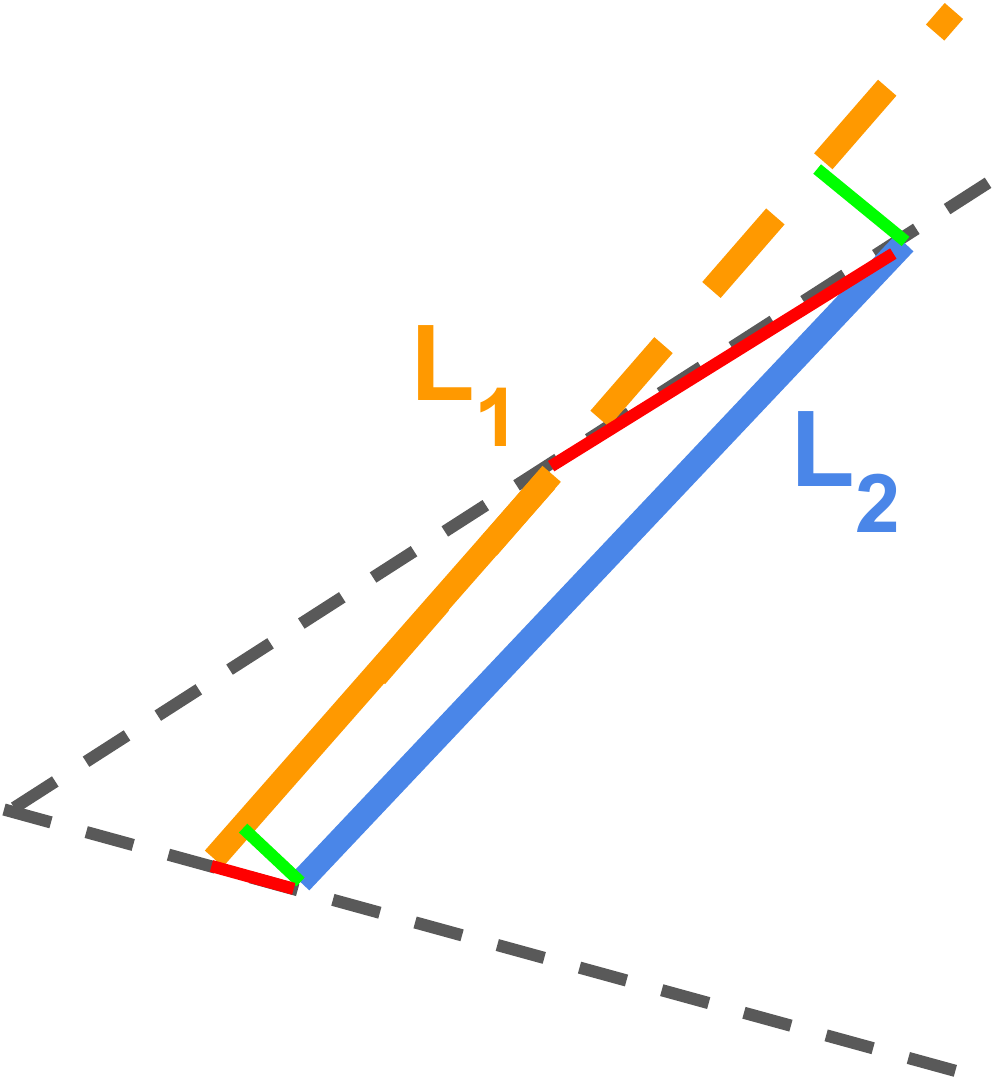} &
\includegraphics[width=0.53\columnwidth]{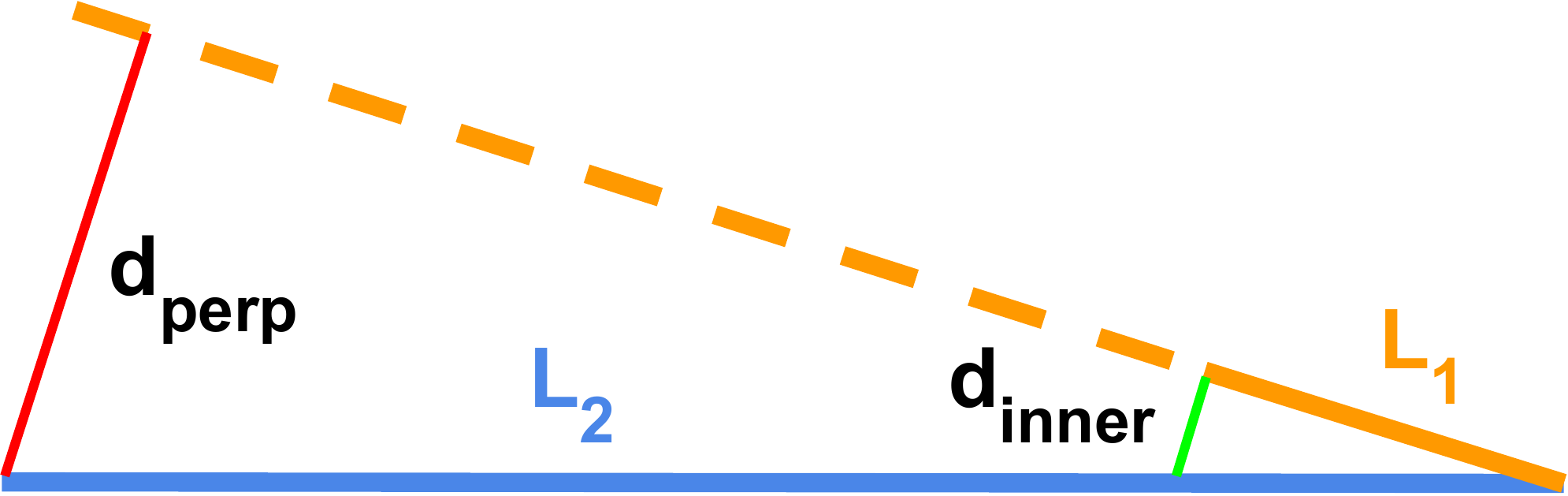} \\
(a) Perspective distance & (b) InnerSeg distance
\end{tabular}
\centering
\caption{\textbf{Benefits of our scoring methods.} (a) Ill-posed line triangulations may have a low perpendicular score (in green), while our perspective distance (in red) can filter them out. (b) The perpendicular distance (in red) is detrimental for very long segments, while our proposed innerseg distance (in green) remains unbiased.}
\label{fig::scoring_supp}
\end{figure}

\subsection{Additional Discussions on Distance Measurements} \label{sec::distance_measurements}

\noindent
\textbf{Benefits of \textit{Perspective Distance}.}
The perspective distance was originally proposed to filter out ill-posed line triangulations that are almost colinear with the ray endpoints of the corresponding 2D segment. In Figure~\ref{fig::scoring_supp}, we show on the left that such ill-posed triangulations may still have a small perpendicular error (in green), and thus, cannot be filtered out with such a classic distance. On the contrary, our proposed perspective scoring (proportional to the endpoint distances in red) will penalize such bad triangulations.

\noindent
\textbf{Benefits of \textit{Innerseg Distance}.}
Another drawback of the perpendicular distance is that it penalizes long segments, as visualized on the right of Figure~\ref{fig::scoring_supp} in red. These long lines are however quite important to get clean reconstructions. Our proposed InnerSeg distance (in green) can effectively avoid this negative bias.

\noindent
\textbf{Scale Factor $\sigma$.}
The uncertainty of the line segment depends on its depth with respect to the two views from which it is triangulated. To make our triangulation invariant to scale changes, we define the scale factor $\sigma$ as the depth of the midpoint divided by the focal length. This essentially encodes how far the midpoint moves in 3D before reaching 1 pixel error on the image. When testing if the \textit{InnerSeg Distance} is within a certain threshold, we rescale the distance with the minimum scale factors of the two, which results in the scale factor $\sigma$ defined in the paper.

\subsection{More Details on the System Design}
\noindent
\textbf{Weak Epipolar Constraint.} We also employ weak epipolar constraints for filtering out matches for line triangulation following \cite{hofer2017efficient}. We measure the IoU between the matched segment and the intersected segments from two epipolar lines from the reference endpoints, and filter out matches if the IoU is below 0.1. For a fair comparison, we also update this hyperparameter in L3D++ \cite{hofer2017efficient} to 0.1, as it empirically gives better performance than its default parameter 0.25 (see Tables \ref{tab::main-hypersim-nv3} and \ref{tab::main-tnt-nv3}). 

\noindent
\textbf{Endpoint Aggregation.}
As discussed in the main paper, after associating the best candidates from 2D line segments into the 3D track, we take the mean and the principle directions over all the 3D endpoints (of the candidates) in the track to get an initial estimate of the infinite 3D line. As we eventually aim to get 3D line segments, we need to compute the 3D endpoints. This is done by projecting all the endpoints from the candidates in the track onto the estimated infinite 3D line. In practice, we take the third outermost endpoints on both sides to give better robustness to unstable triangulations in the track. This robust selection of the third outmost endpoint is also done after joint optimization, when the optimized infinite 3D line with Pl\"ucker coordinate is converted into a 3D line segment. 

\noindent
\textbf{Track Remerging.}
Optionally, we also support remerging similar tracks together after track building, as some tracks may have very close 3D lines. Specifically, we can recompute the pairwise scores among the re-fitted 3D lines of each track, and greedily merge tracks with a stricter threshold. 

\subsection{Details on L3D++ \cite{hofer2017efficient} and ELSR \cite{wei2022elsr}}

For L3D++ \cite{hofer2017efficient}, we use the open-sourced implementation from their official repository \cite{Line3Dpp}. For SOLD2 detector \cite{pautrat2021sold2}, we detect line segments in advance and save the segments into a compatible format that can be processed from L3D++ \cite{Line3Dpp}. We update two of their hyperparameters: visual neighbors from 10 to 20, and IoU threshold for the weak epipolar constraint from 0.25 to 0.10, to enable fair comparison, while we also present their results with the default hyperparameters in Tables \ref{tab::main-hypersim-nv3} and \ref{tab::main-tnt-nv3}. 

For ELSR \cite{wei2022elsr}, we use the official release from the authors on their website. Since they only support VisualSfM input \cite{wu2011visualsfm} with LSD detector \cite{von2008lsd}, we convert the COLMAP model triangulated on Hypersim \cite{roberts:2021} and provided from \textit{Tanks and Temples} \cite{Knapitsch2017} into the VisualSfM format \cite{wu2011visualsfm} such that it is compatible with their implementation.

\section{More Results on Line Mapping}
\label{sec::results_mapping}
\subsection{Scale invariance}
\begin{figure}[tb]
\scriptsize
\setlength\tabcolsep{2pt} 
\begin{tabular}{cccc}
{\includegraphics[trim={0 0 0 0}, clip, width=0.22\linewidth, height=15pt]{figs/imgs/tnt/horse/00069.jpg}} &
{\includegraphics[trim={0 0 0 0}, clip, width=0.22\linewidth, height=15pt]{figs/imgs/tnt/horse/00074.jpg}} &
{\includegraphics[trim={0 0 0 0}, clip, width=0.22\linewidth, height=15pt]{figs/imgs/tnt/horse/00095.jpg}} &
{\includegraphics[trim={0 0 0 0}, clip, width=0.22\linewidth, height=15pt]{figs/imgs/tnt/horse/00102.jpg}} \\
\hline
\end{tabular}
\begin{tabular}{|c|c|}
{\includegraphics[trim={0 50 0 100}, clip, width=0.45\linewidth, height=65pt]{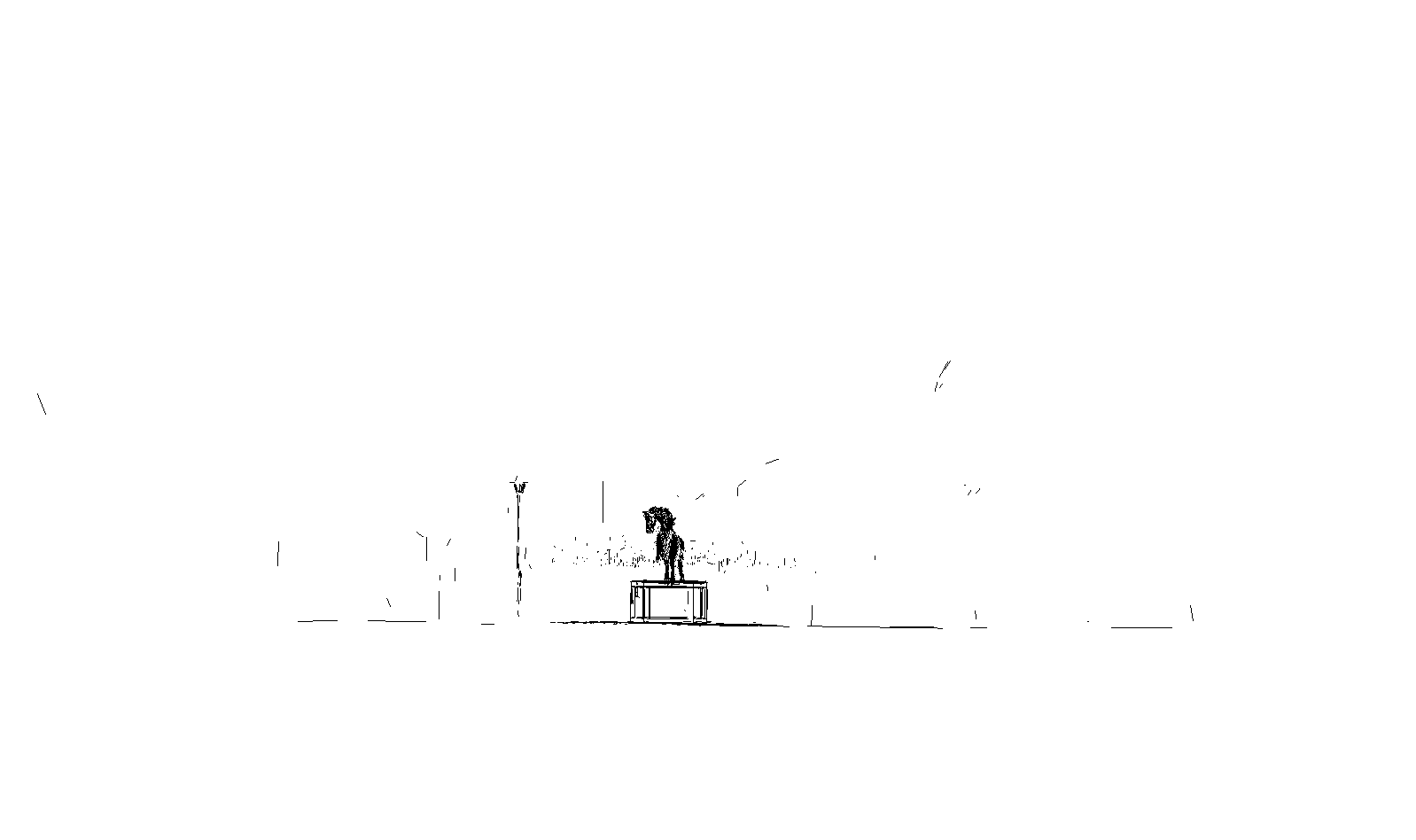}} & 
{\includegraphics[trim={0 50 0 100}, clip, width=0.45\linewidth, height=65pt]{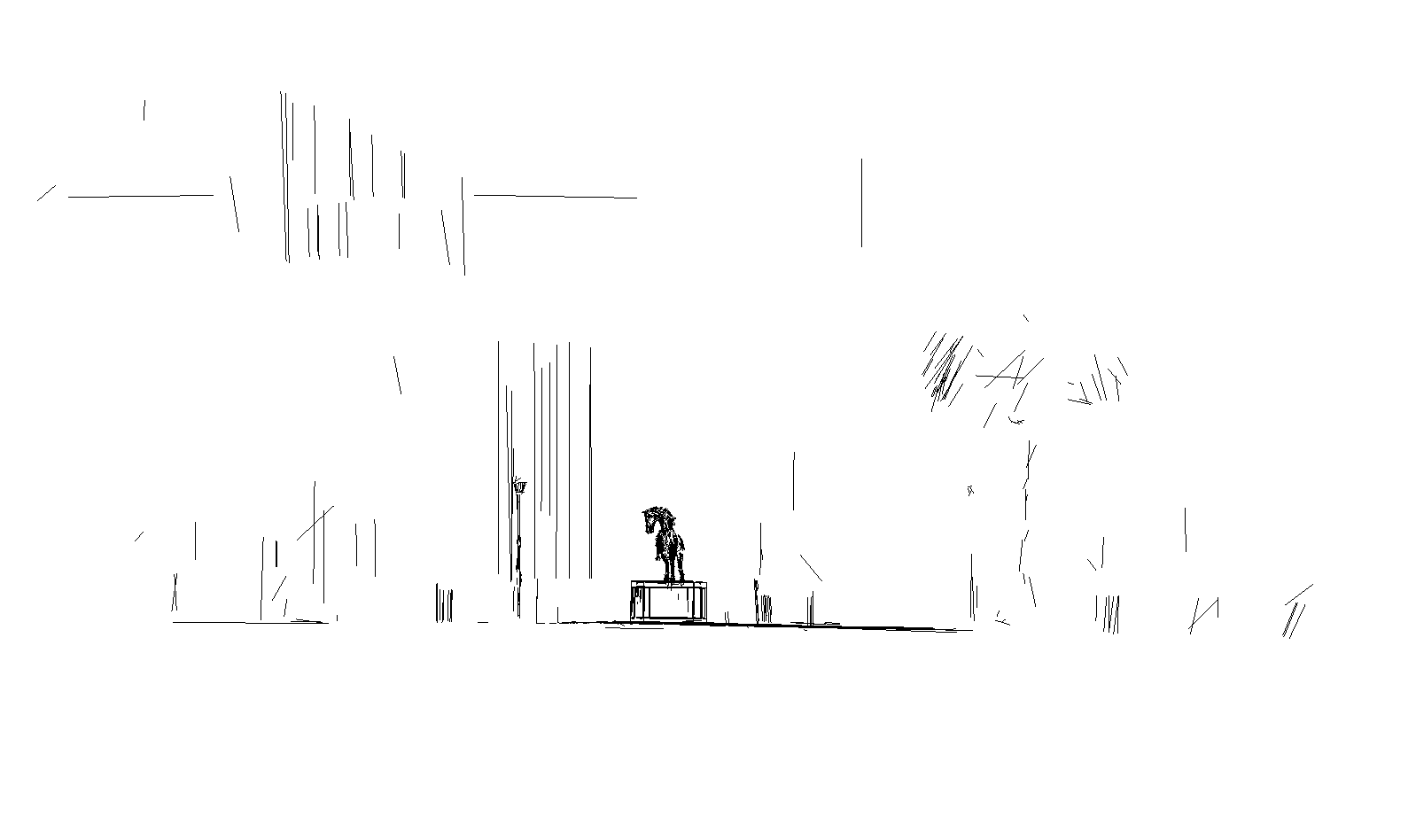}}
\\
\hline
{\includegraphics[trim={0 50 0 100}, clip, width=0.45\linewidth, height=65pt]{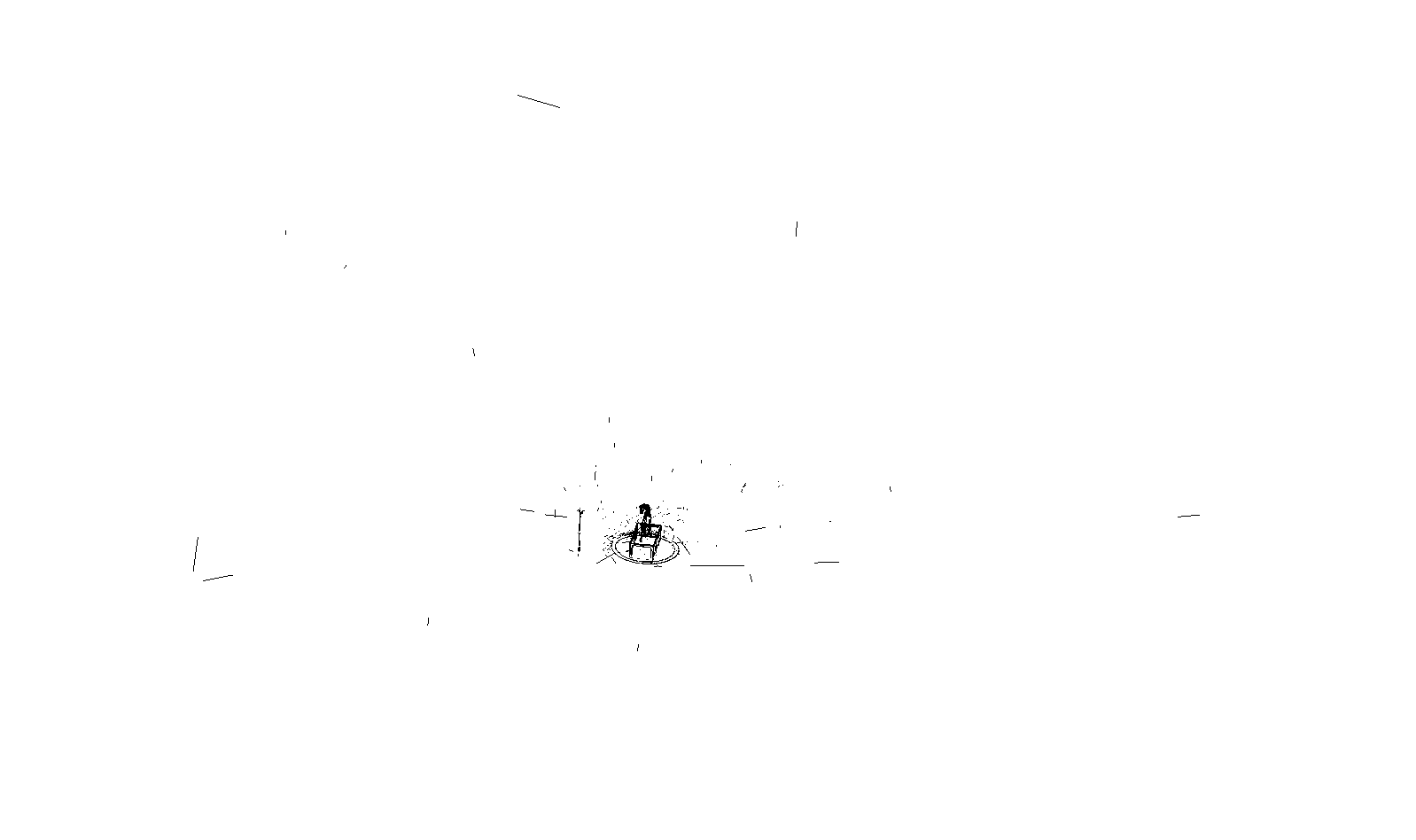}} &
{\includegraphics[trim={0 50 0 100}, clip, width=0.45\linewidth, height=65pt]{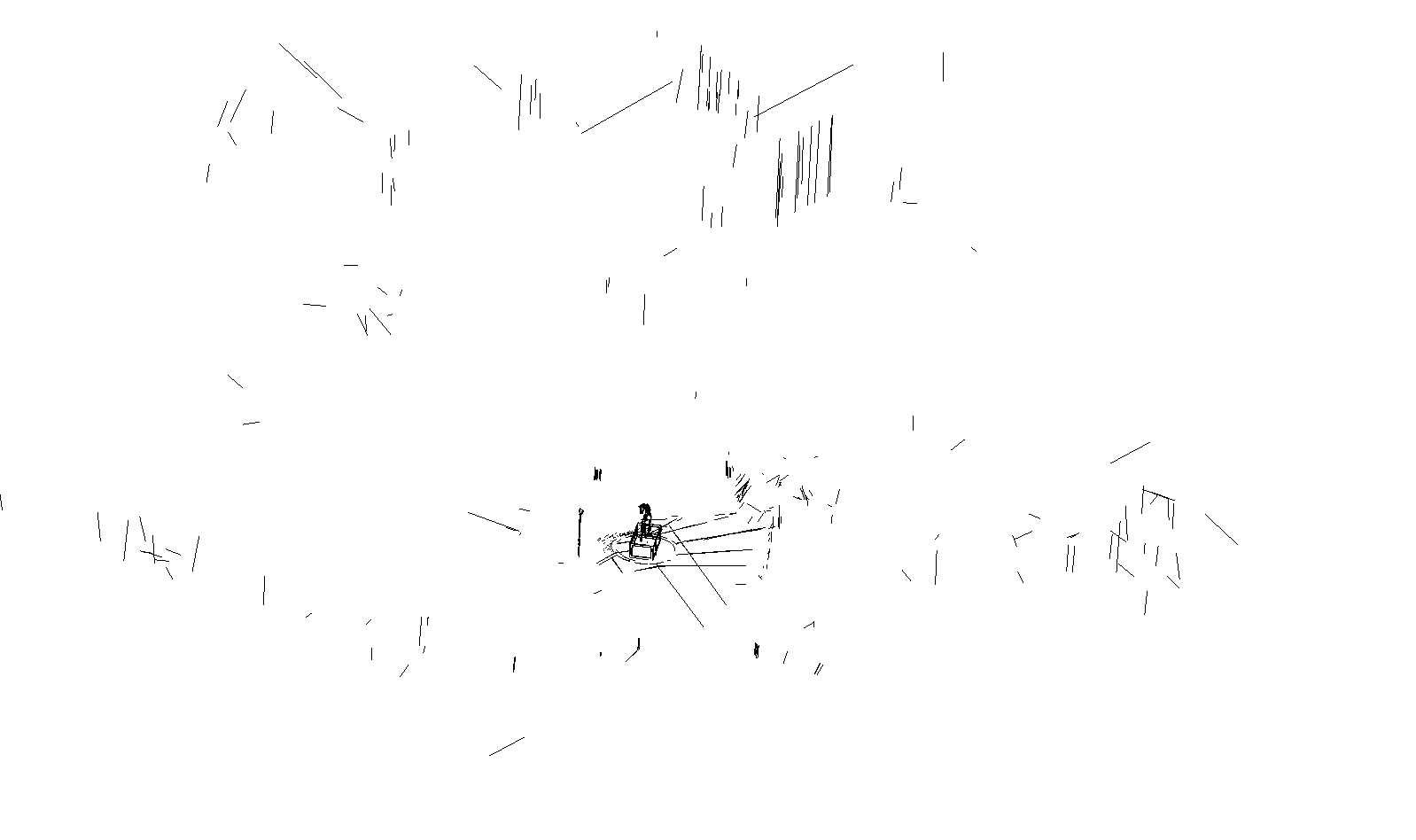}}
\\
\hline
\end{tabular}
\centering
\caption{\textbf{Illustration on the advantages of our scale-invariant design.} \textbf{Left:} L3D++ \cite{hofer2017efficient}. \textbf{Right:} Ours. Both systems are run on \textit{Horse} from \textit{Tanks and Temples} \cite{Knapitsch2017}. We show two different views on the full scene where our method can reconstruct lines that are far away from the main subject. Refer back to Figure 4 in the main paper for zoom-in comparison on the horse.}
\label{fig::scale_invariance}
\end{figure}

Our pipeline is robust to scale changes. This not only refers to the global scale of the scene, but also refers to the local scale of the sub-components of the whole scene: it is very common that one gets different layers of subjects and buildings with large depth changes. 

To demonstrate the benefits of our scale-invariant design, we here show a visualization of how the scale-invariant design can help to reconstruct lines across very different scales in Figure \ref{fig::scale_invariance}. Specifically, We compare the reconstructions of \textit{Horse} from \textit{Tanks and Temples}~\cite{Knapitsch2017} given by L3D++ \cite{hofer2017efficient} and our method. Our method is able to reconstruct many more far-away lines in the background compared to L3D++ \cite{hofer2017efficient}, while providing a very accurate reconstruction for close-by details as well (see Figure 4 of the main paper).

\subsection{More Comparisons with L3D++ \cite{hofer2017efficient}}
\begin{table*}[tb]
\begin{center}
\scriptsize
\setlength{\tabcolsep}{10pt}
\begin{tabular}{clccccccc}
\toprule
Line type & Method & R1 & R5 & R10 & P1 & P5 & P10 & \# supports \\
\midrule
\multirow{8}{*}{\makecell{LSD\\\cite{von2008lsd}}} 
& (nv = 4) L3D++ default param. \cite{hofer2017efficient} & 34.6 & 139.9 & 196.6 & 53.3 & \textbf{82.6} & \textbf{92.6} & (11.6 / 12.5) \\
& (nv = 4) L3D++ \cite{hofer2017efficient} & 37.0 & 153.1 & 218.8 & 53.1 & 80.8 & 90.6 & (14.8 / 16.8) \\
& (nv = 4) ELSR \cite{wei2022elsr} & 13.9 & 59.7 & 96.5 & 55.4 & 72.6 & 82.2 & (N/A / N/A) \\
& (nv = 4) Ours & \textbf{48.6} & \textbf{185.2} & \textbf{251.3} & \textbf{60.1} & 82.4 & 90.0 & (\textbf{16.4} / \textbf{20.5}) \\
\cline{2-9}
& (nv = 3) L3D++ default param. \cite{hofer2017efficient} & 40.9 & 166.2 & 235.8 & 49.3 & 76.7 & 86.9 & (8.7 / 9.4) \\
& (nv = 3) L3D++ \cite{hofer2017efficient} & 40.6 & 168.8 & 242.8 & 50.3 & 77.3 & \textbf{87.6} & (12.1 / 13.6) \\
& (nv = 3) ELSR \cite{wei2022elsr} & 13.9 & 59.7 & 96.5 & 55.4 & 72.6 & 82.2 & (N/A / N/A) \\
& (nv = 3) Ours & \textbf{51.9} & \textbf{198.1} & \textbf{271.0} & \textbf{56.7} & \textbf{78.2} & 86.3 & (\textbf{13.6} / \textbf{16.8}) \\
\midrule
\multirow{6}{*}{\makecell{SOLD2\\ \cite{pautrat2021sold2}}} 
& (nv = 4) L3D++ default param. \cite{hofer2017efficient} & 29.7 & 84.7 & 102.3 & 67.2 & \textbf{88.5} & \textbf{96.0} & (9.9 / 12.4) \\
& (nv = 4) L3D++ \cite{hofer2017efficient} & 36.9 & 107.5 & 132.8 & 67.2 & 86.8 & 93.2 & (13.2 / 20.4) \\
& (nv = 4) Ours & \textbf{54.3} & \textbf{151.1} & \textbf{191.2} & \textbf{69.8} & 84.6 & 90.0 & (\textbf{16.5} / \textbf{38.7}) \\
\cline{2-9}
& (nv = 3) L3D++ default param. \cite{hofer2017efficient} & 34.9 & 102.3 & 127.1 & 61.3 & \textbf{82.2} & \textbf{90.4} & (7.4 / 9.2) \\
& (nv = 3) L3D++ \cite{hofer2017efficient} & 40.3 & 118.7 & 148.0 & 62.3 & 82.0 & 89.6 & (10.6 / 16.0) \\
& (nv = 3) Ours & \textbf{55.6} & \textbf{155.4} & \textbf{197.4} & \textbf{66.8} & 82.0 & 88.0 & (\textbf{14.1} / \textbf{32.5}) \\
\bottomrule
\end{tabular}
\caption{More results on Hypersim \cite{roberts:2021} with minimum 3 supporting images. We also add the results of L3D++ \cite{hofer2017efficient} with its default hyperparameters. ``nv" denotes the minimum number of supporting images for a line track to be included in the final output. Note that even compared to L3D++ with nv = 3, our mapping with nv = 4 achieves significantly better recall while being better on precision at all thresholds as well. This demonstrates the superiority of our method over L3D++ \cite{hofer2017efficient} when moving along the recall-precision trade-off curve.}
\label{tab::main-hypersim-nv3}
\end{center}
\end{table*}

\begin{table*}[tb]
\begin{center}
\scriptsize
\setlength{\tabcolsep}{10pt}
\begin{tabular}{lccccccc}
\toprule
Method & R5 & R10 & R50 & P5 & P10 & P50 & \# supports \\
\midrule
(nv = 4) L3D++ default param. \cite{hofer2017efficient} & 215.4 & 477.7 & 1543.6 & 41.3 & 55.8 & \textbf{87.2} & (6.4 / 6.5) \\
(nv = 4) L3D++ \cite{hofer2017efficient} & 373.7 & 831.6 & 2783.6 & 40.6 & 54.5 & 85.9 & (8.8 / 9.3) \\
(nv = 4) ELSR \cite{wei2022elsr} & 139.2 & 322.5 & 1308.0 & 38.5 & 48.0 & 74.5 & (N/A / N/A) \\
(nv = 4) Ours (line-only) & 472.1 & 1058.8 & 3720.7 & \textbf{46.8} & \textbf{58.4} & 86.1 & (10.3 / 11.8) \\
(nv = 4) Ours & \textbf{508.3} & \textbf{1154.5} & \textbf{4179.5} & 46.0 & 56.9 & 83.7 & (\textbf{10.4} / \textbf{12.0}) \\
\midrule
(nv = 3) L3D++ default param. \cite{hofer2017efficient} & 313.1 & 698.3 & 2351.6 & 32.2 & 44.0 & 72.5 & (4.7 / 4.8) \\
(nv = 3) L3D++ \cite{hofer2017efficient} & 473.7 & 1058.1 & 3622.4 & 35.6 & 48.5 & 79.6 & (6.6 / 7.0) \\
(nv = 3) ELSR \cite{wei2022elsr} & 139.2 & 322.5 & 1308.0 & 38.5 & 48.0 & 74.5 & (N/A / N/A) \\
(nv = 3) Ours (line-only) & 564.8 & 1267.2 & 4539.0 & \textbf{43.2} & \textbf{54.5} & \textbf{83.7} & (7.8 / 8.9) \\
(nv = 3) Ours & \textbf{606.7} & \textbf{1379.5} & \textbf{5047.1} & 42.1 & 52.8 & 80.9 & (\textbf{7.9} / \textbf{9.0}) \\
\bottomrule
\end{tabular}
\caption{More results on Tanks and Temples \cite{Knapitsch2017} with minimum 3 supporting images. We also add the results of L3D++ \cite{hofer2017efficient} with its default hyperparameters. ``nv" denotes the minimum number of supporting images for a line track to be included in the final output.}
\label{tab::main-tnt-nv3}
\end{center}
\end{table*}

\begin{figure}[tb]
\scriptsize
\setlength\tabcolsep{6pt} 
\begin{tabular}{cc}
{\includegraphics[trim={300 120 400 150}, clip, width=0.47\linewidth, height=70pt]{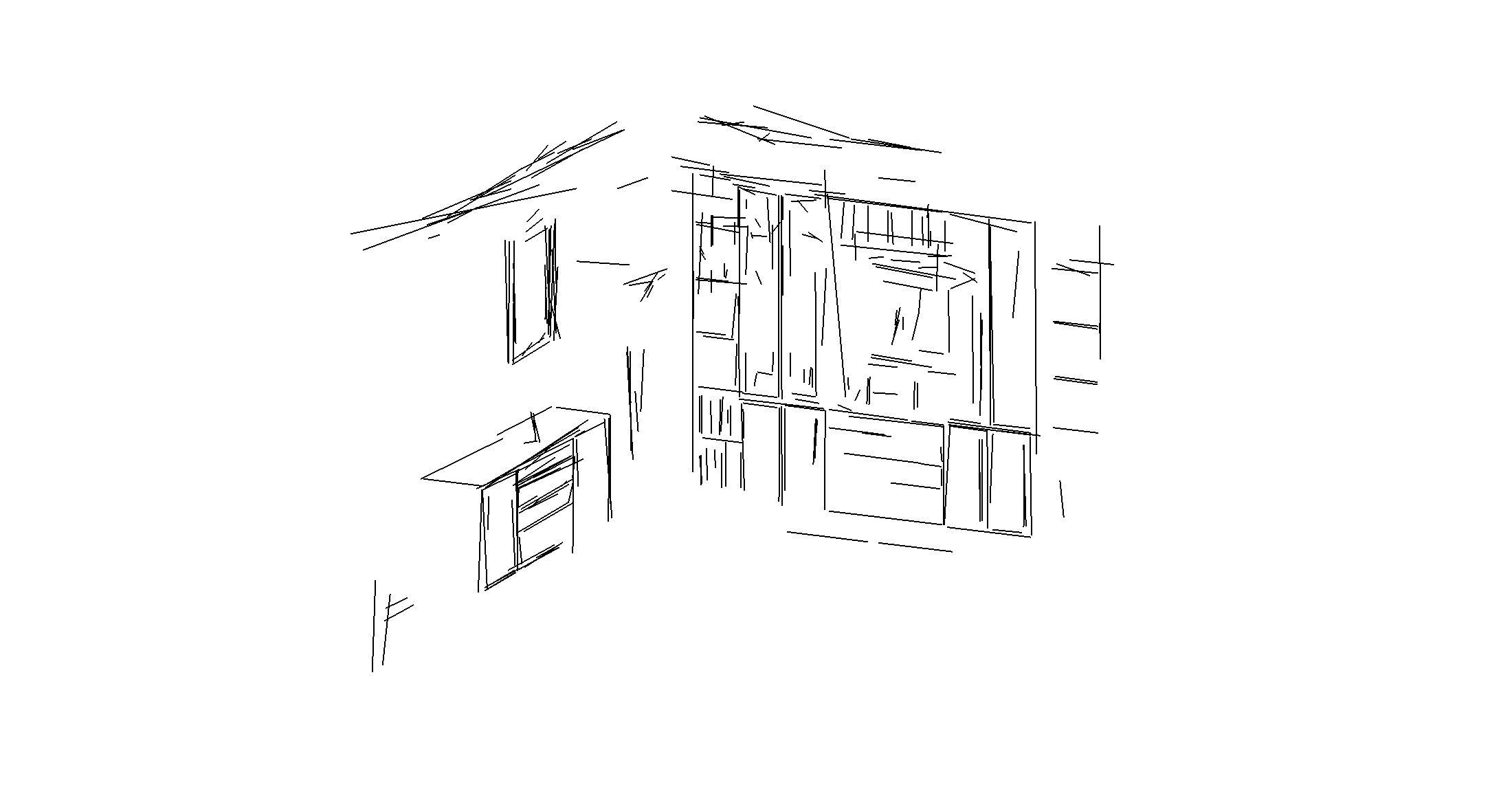}} &
{\includegraphics[trim={300 120 300 150}, clip, width=0.47\linewidth, height=70pt]{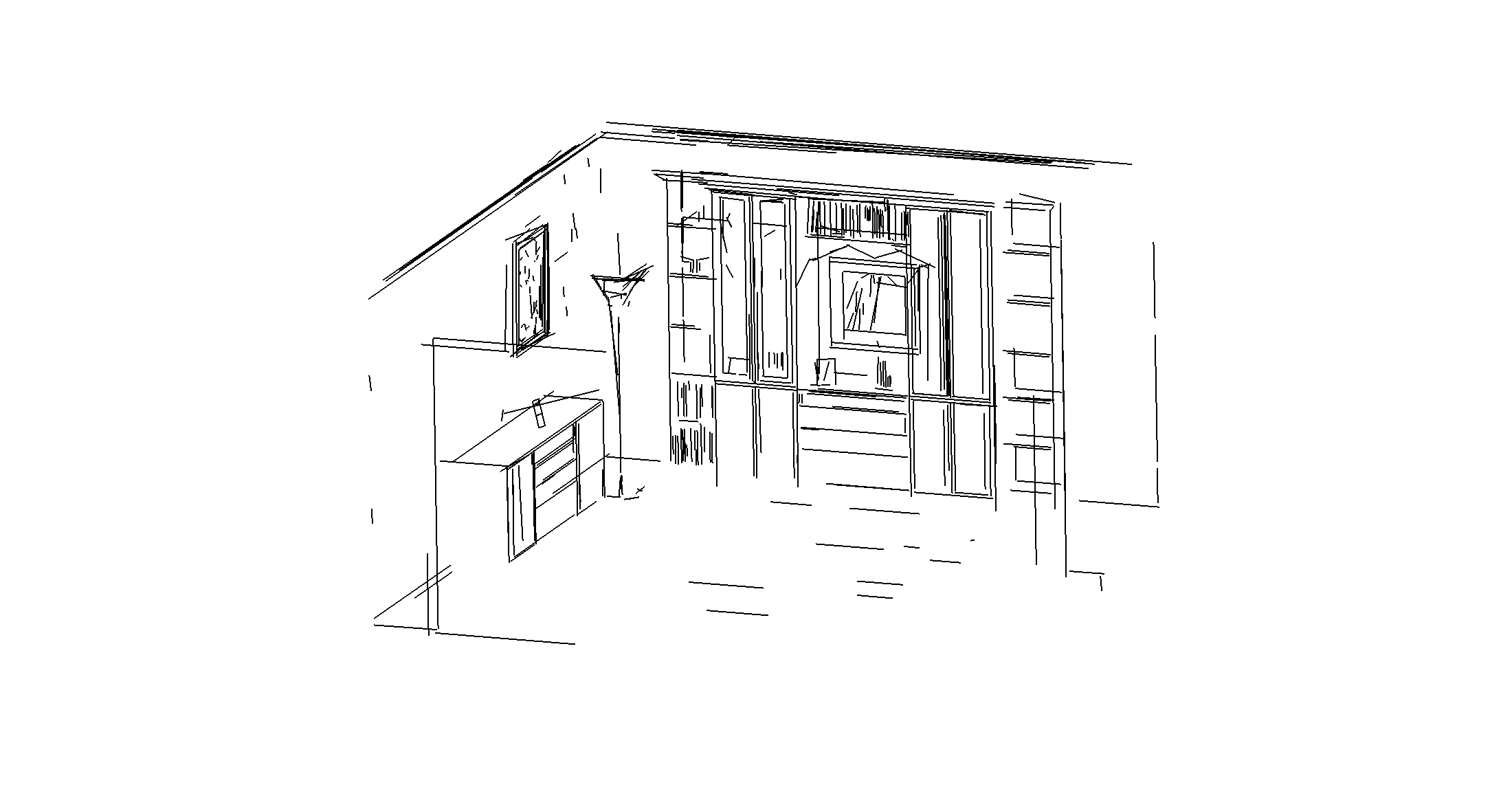}} \\
{\includegraphics[trim={200 200 350 0}, clip, width=0.47\linewidth, height=70pt]{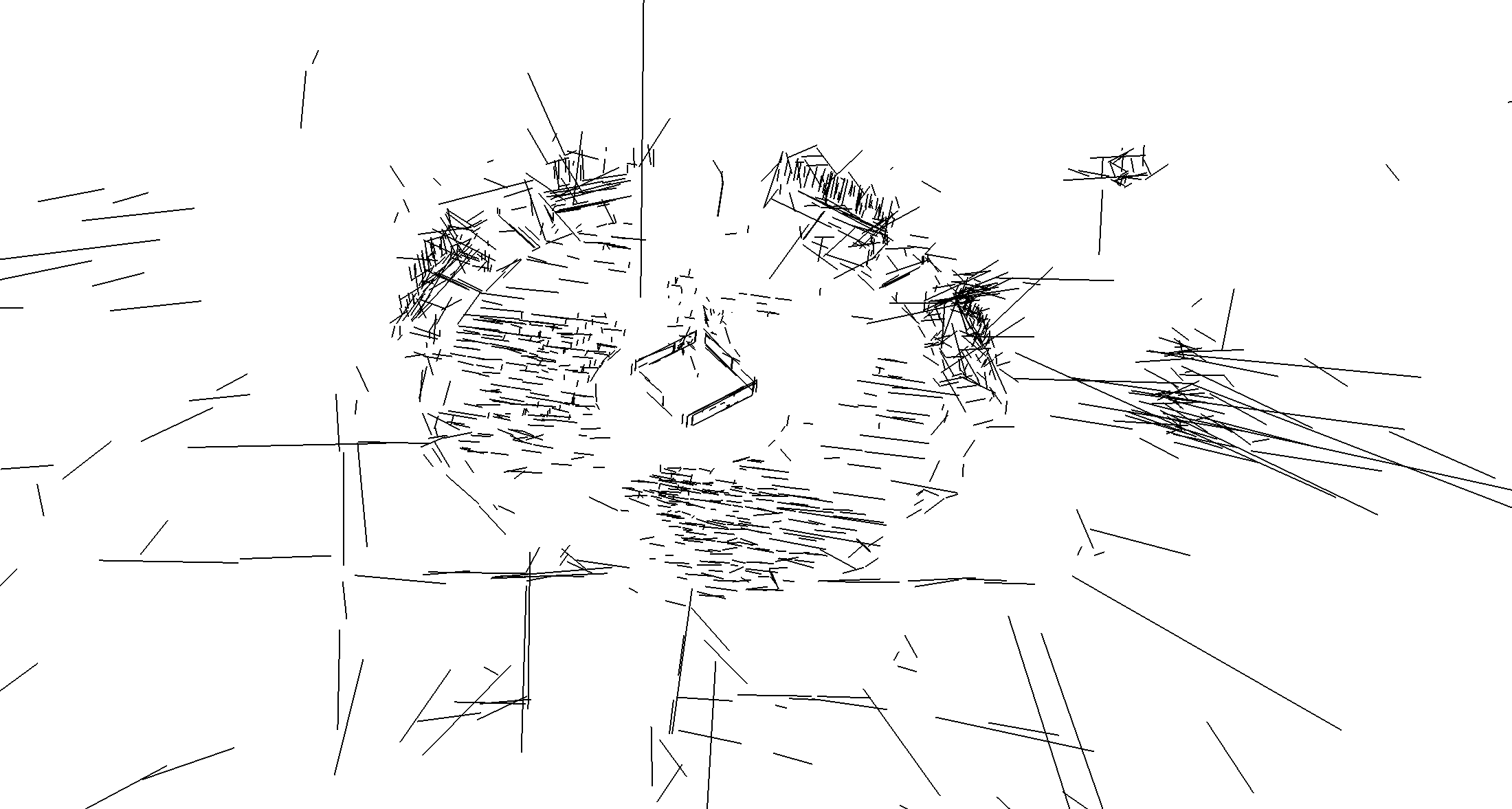}} &
{\includegraphics[trim={0 0 0 40}, clip, width=0.47\linewidth, height=70pt]{figs/imgs/tnt/family/family_ours.png}} \\
ELSR \cite{wei2022elsr} & LIMAP (Ours) \\
\end{tabular}
\centering
\caption{\textbf{Qualitative comparisons with ELSR \cite{wei2022elsr}.} We show two examples from Hypersim \cite{roberts:2021} and Tanks and Temples \cite{Knapitsch2017}. }
\label{fig::elsr_comparison}
\end{figure}

\begin{table*}[tb]
    \centering
    \scriptsize
    \setlength{\tabcolsep}{5pt}
    \begin{tabular}{l|ccccc}
        \toprule
        \backslashbox[34mm]{Matcher}{Detector} & LSD \cite{von2008lsd} & HAWPv3 \cite{xue2020holistically} & TP-LSD \cite{huang2020tp} & SOLD2 \cite{pautrat2021sold2} & DeepLSD \cite{Pautrat_2023_DeepLSD} \\
        \midrule
        LBD \cite{zhang2013efficient} & 42.2 / 58.5 / (14.0 / 14.6) & 6.0 / 58.0 / (7.8 / 9.8) & 21.6 / 73.2 / (9.1 / 9.3) & 30.7 / 69.3 / (12.2 / 18.7) & 64.6 / 70.0 / (15.8 / 18.1) \\
        SOLD2 \cite{pautrat2021sold2} & 48.3 / 59.2 / (15.8 / 19.1) & 14.7 / 62.7 / (11.2 / 20.1) & 44.4 / 76.4 / (14.3 / 16.7) & 50.8 / 74.4 / (15.1 / 32.2) & 72.0 / 71.4 / (18.1 / 24.9) \\
        L2D2 \cite{abdellali2021l2d2} & 44.4 / 59.6 / (15.0 / 16.8) & 13.5 / 63.4 / (10.7 / 18.3) & 39.5 / \textbf{78.1} / (13.7 / 15.4) & 43.9 / 72.8 / (13.7 / 24.9) & 69.2 / 70.4 / (17.0 / 22.2) \\
        LineTR \cite{yoon2021line} & 37.0 / 58.3 / (12.8 / 13.3) & 5.4 / 60.5 / (8.4 / 10.7) & 43.0 / 76.3 / (14.5 / 16.7) & 29.0 / 70.1 / (12.3 / 19.9) & 71.9 / 69.4 / (17.6 / 23.9) \\
        Endpts SP \cite{detone2018superpoint} + NN & 48.8 / 58.6 / (15.5 / 18.2) & 16.2 / 63.2 / (11.2 / 20.0) & 43.7 / 75.8 / (14.3 / 16.5) & 49.1 / 73.7 / (14.7 / 31.4) & 72.8 / 70.3 / (17.7 / 24.0) \\
        Endpts SP \cite{detone2018superpoint} + SG \cite{sarlin2020superglue}  & 48.4 / 58.0 / (15.8 / 18.9) & 16.0 / 61.9 / (11.3 / 20.9) & 47.1 / 76.1 / (14.5 / 16.8) & 50.0 / 72.8 / (15.5 / \textbf{34.4}) & \textbf{74.6} / 69.5 / (\textbf{18.2} / 24.8) \\
        \bottomrule
    \end{tabular}
    \caption{\textbf{Extensibility of the framework to different line detectors and matchers.} We show results of ``R1 / P1 / \#supports" for line mapping on Hypersim \cite{roberts:2021} with only line-line proposals. Also, we present here two customized line matchers by matching the line endpoints with either the nearest neighbor strategy (``Endpts SP + NN") or an advanced point-based matcher SuperGlue \cite{sarlin2020superglue} (``Endpts SP + SG"). }
    \label{tab:main_different_line_methods}
\end{table*}

\begin{table}[tb]
\begin{center}
\scriptsize
\setlength{\tabcolsep}{3pt}
\begin{tabular}{clccccccc}
\toprule
Line type & Triangulation & R1 & R5 & R10 & P1 & P5 & P10 & \# supports \\
\midrule
\multirow{2}{*}{\makecell{LSD\\\cite{von2008lsd}}} & Endpoints & 27.5 & 102.3 & 140.9 & 57.4 & \textbf{83.6} & \textbf{92.3} & (13.1 / 13.3) \\
& Line & \textbf{48.4} & \textbf{185.4} & \textbf{255.2} & \textbf{58.0} & 80.7 & 88.6 & (\textbf{15.8} / \textbf{18.9}) \\
\midrule
\multirow{2}{*}{\makecell{SOLD2\\ \cite{pautrat2021sold2}}} & Endpoints & 29.4 & 87.6 & 111.3 & 67.0 & 83.8 & \textbf{90.4} & (12.3 / 20.2) \\
& Line & \textbf{50.0} & \textbf{144.0} & \textbf{181.5} & \textbf{72.8} & \textbf{85.3} & 90.2 & (\textbf{15.5} / \textbf{34.4}) \\
\bottomrule
\end{tabular}
\caption{\textbf{Comparison between endpoint and line triangulation with ``Endpts SP + SG" matcher} on Hypersim \cite{roberts:2021}. While endpoint-based line matcher achieves very promising performance in Table \ref{tab:main_different_line_methods}, the endpoints of the matched line still do not necessarily match each other, indicating that the performance gain may come from the effectiveness of advanced point features \cite{detone2018superpoint}. }
\label{tab::supp_ablations_endpoints}
\end{center}
\end{table}

\begin{figure*}
\scriptsize
\setlength\tabcolsep{2pt} 
\begin{tabular}{ccc}
{\includegraphics[width=0.3\linewidth, height=114pt]{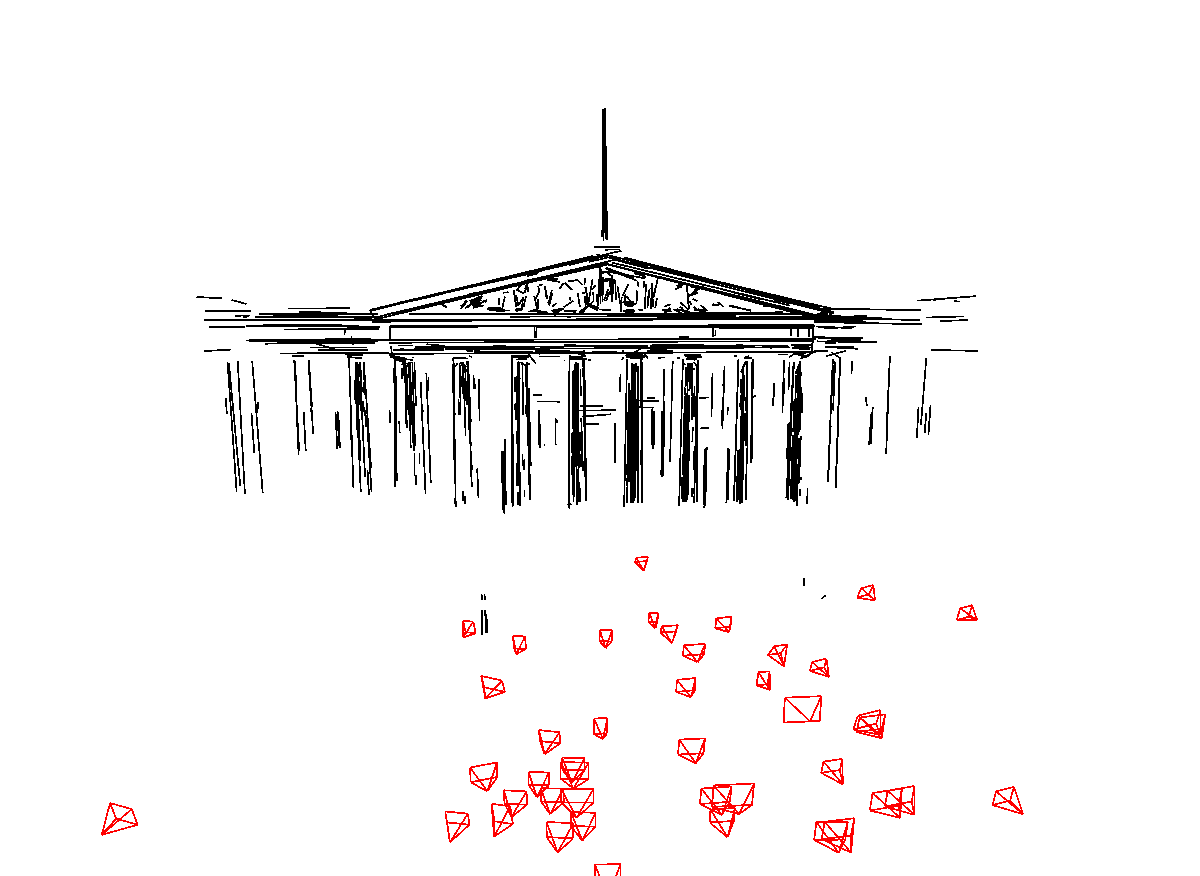}} & 
{\includegraphics[width=0.3\linewidth, height=114pt]{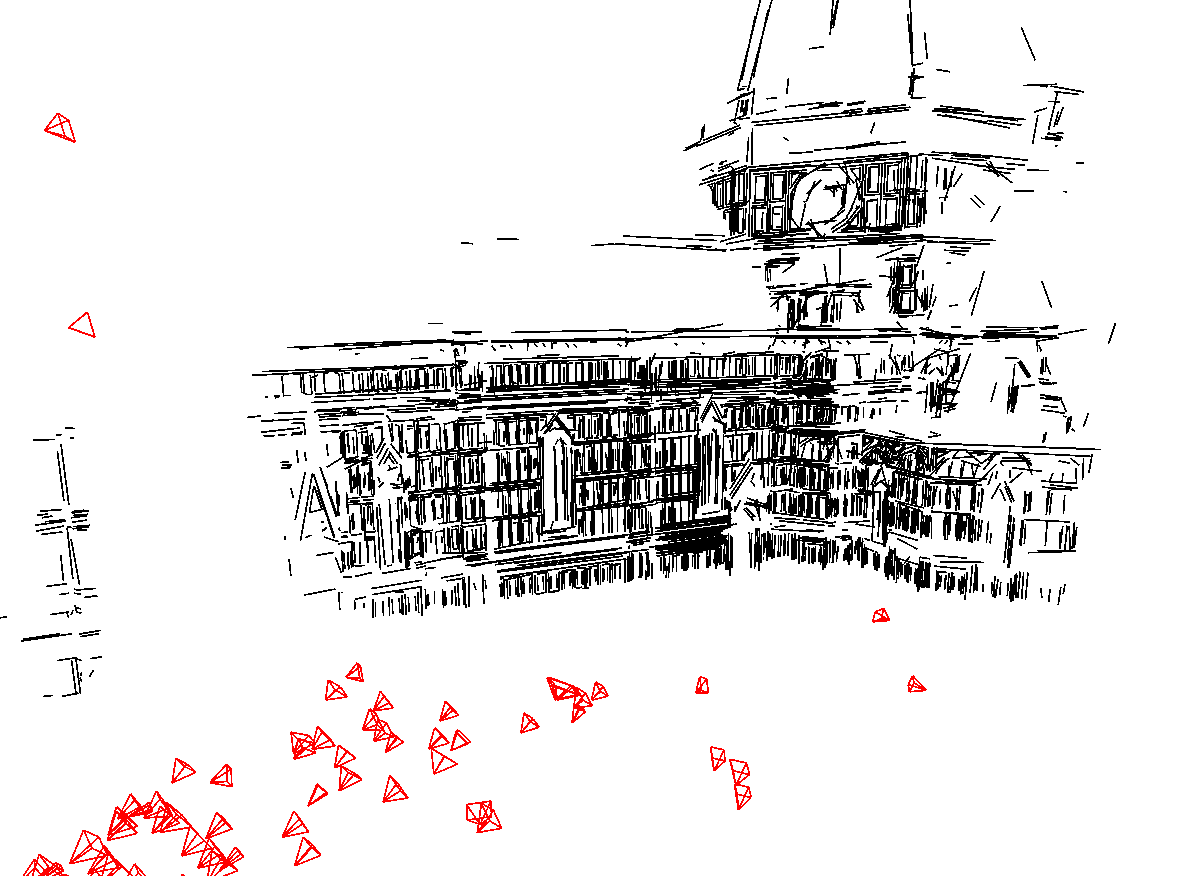}} &
{\includegraphics[width=0.3\linewidth, height=114pt]{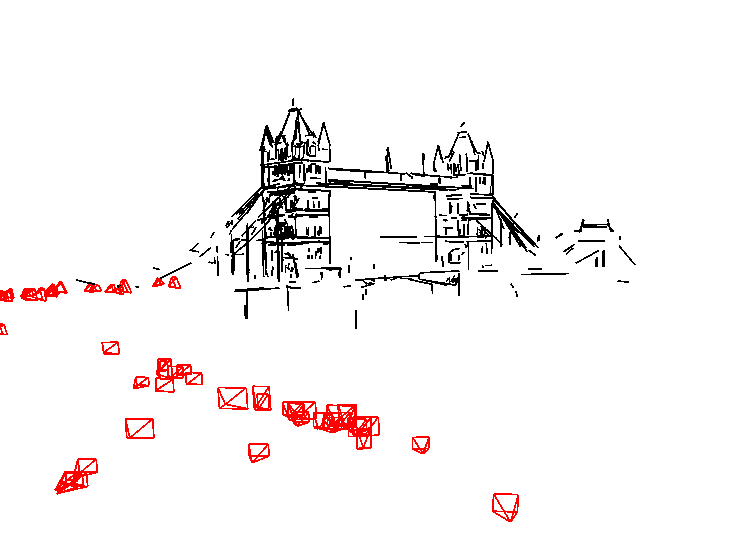}} \\
\textit{British Museum} from \cite{snavely2006photo} &
\textit{Florence Cathedral Side} from \cite{snavely2006photo} &
\textit{London Bridge} from \cite{snavely2006photo}
\\
{\includegraphics[width=0.3\linewidth, height=114pt]{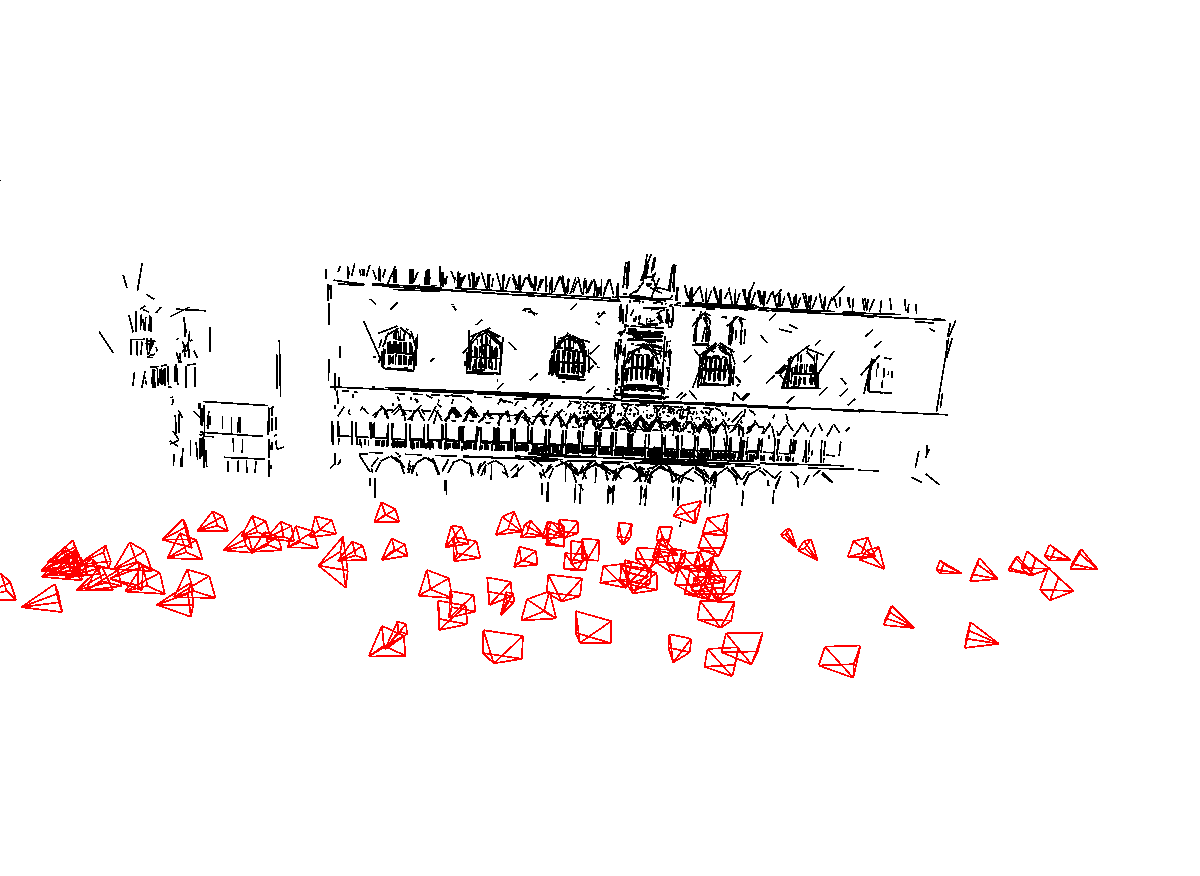}} & 
{\includegraphics[width=0.3\linewidth, height=114pt]{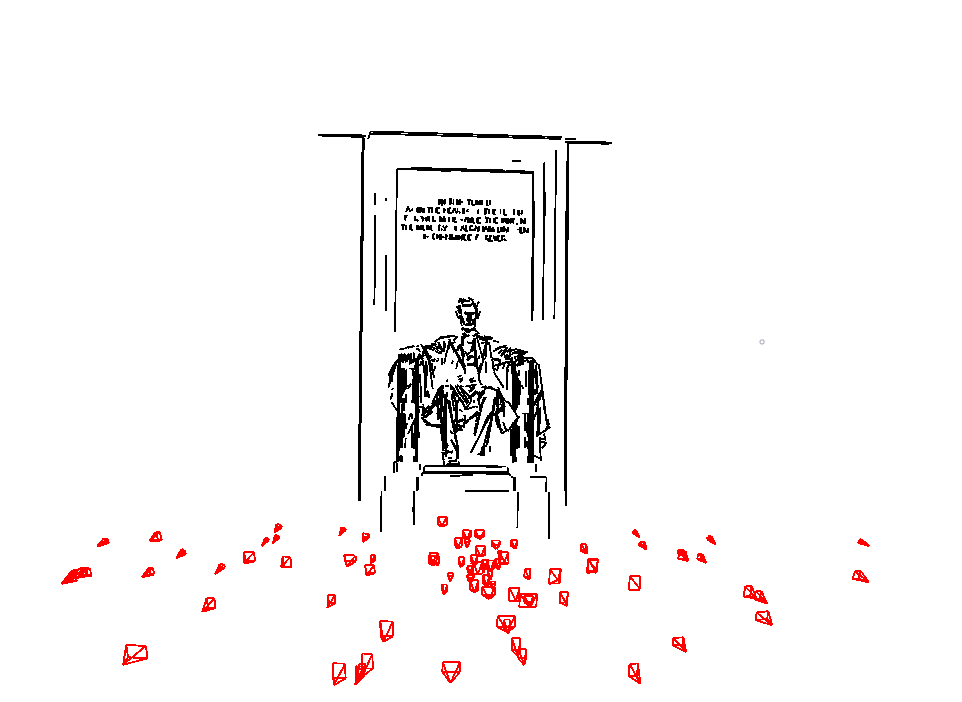}} &
{\includegraphics[width=0.3\linewidth, height=114pt]{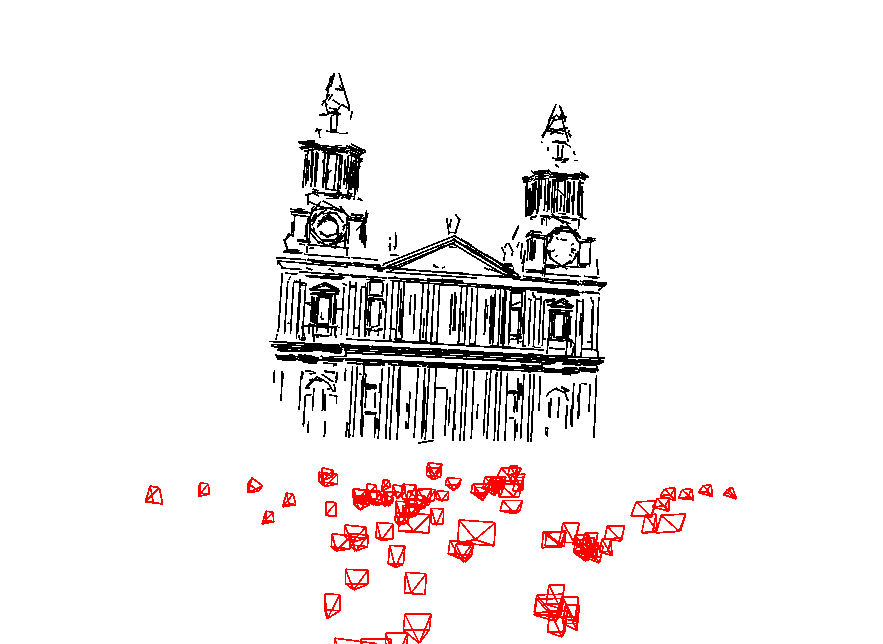}}
 \\
\textit{Piazza San Marco} from \cite{snavely2006photo}&
\textit{Lincoln Memorial Statue} from \cite{snavely2006photo} &
\textit{St. Paul's Cathedral} from \cite{snavely2006photo}
\\
{\includegraphics[width=0.3\linewidth, height=114pt]{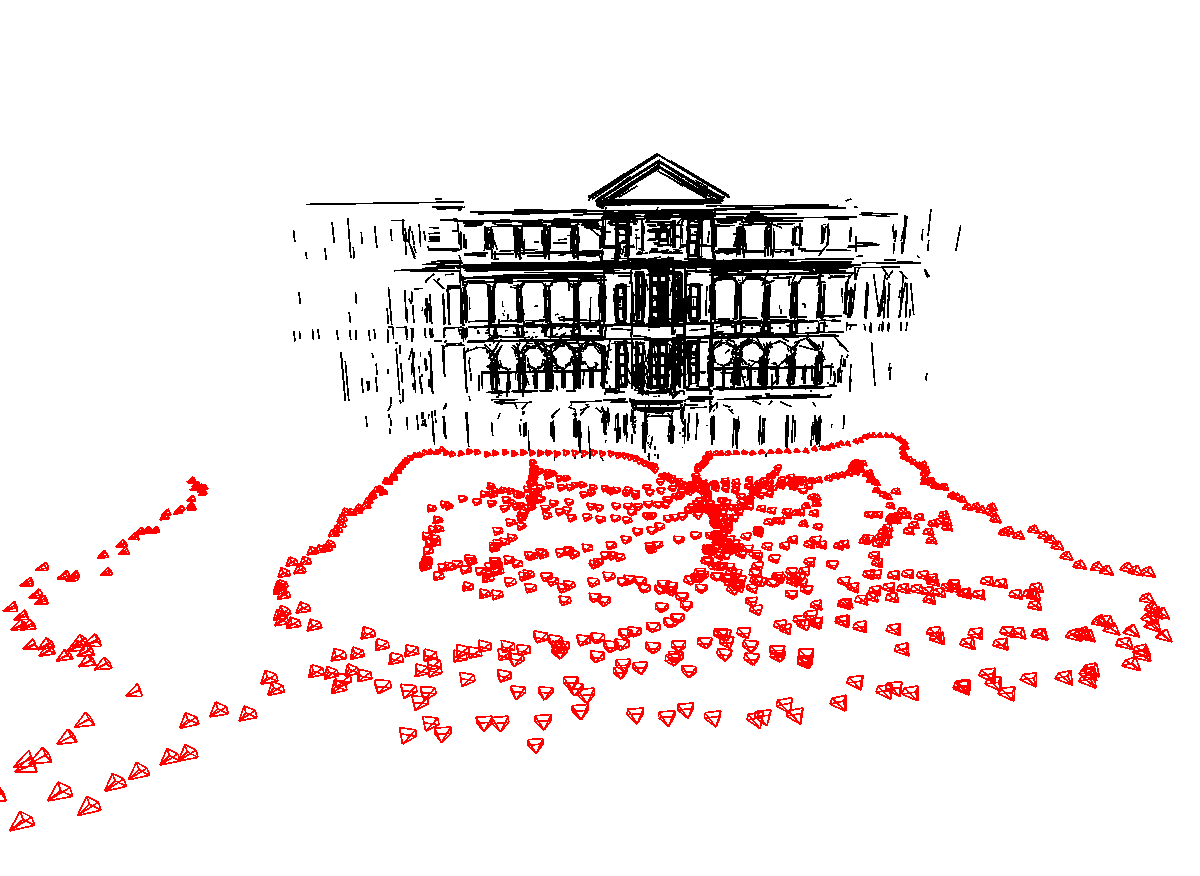}} & 
{\includegraphics[width=0.3\linewidth, height=114pt]{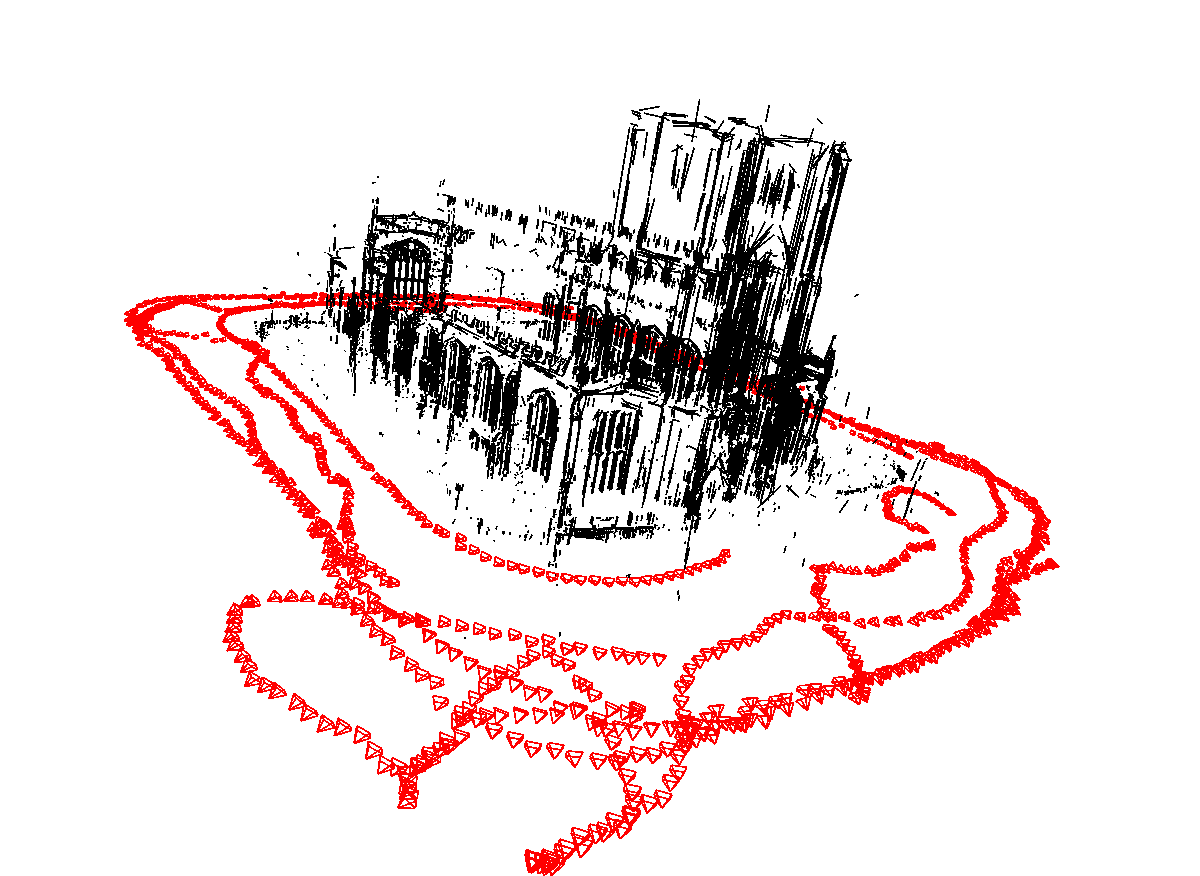}} & 
{\includegraphics[width=0.3\linewidth, height=114pt]{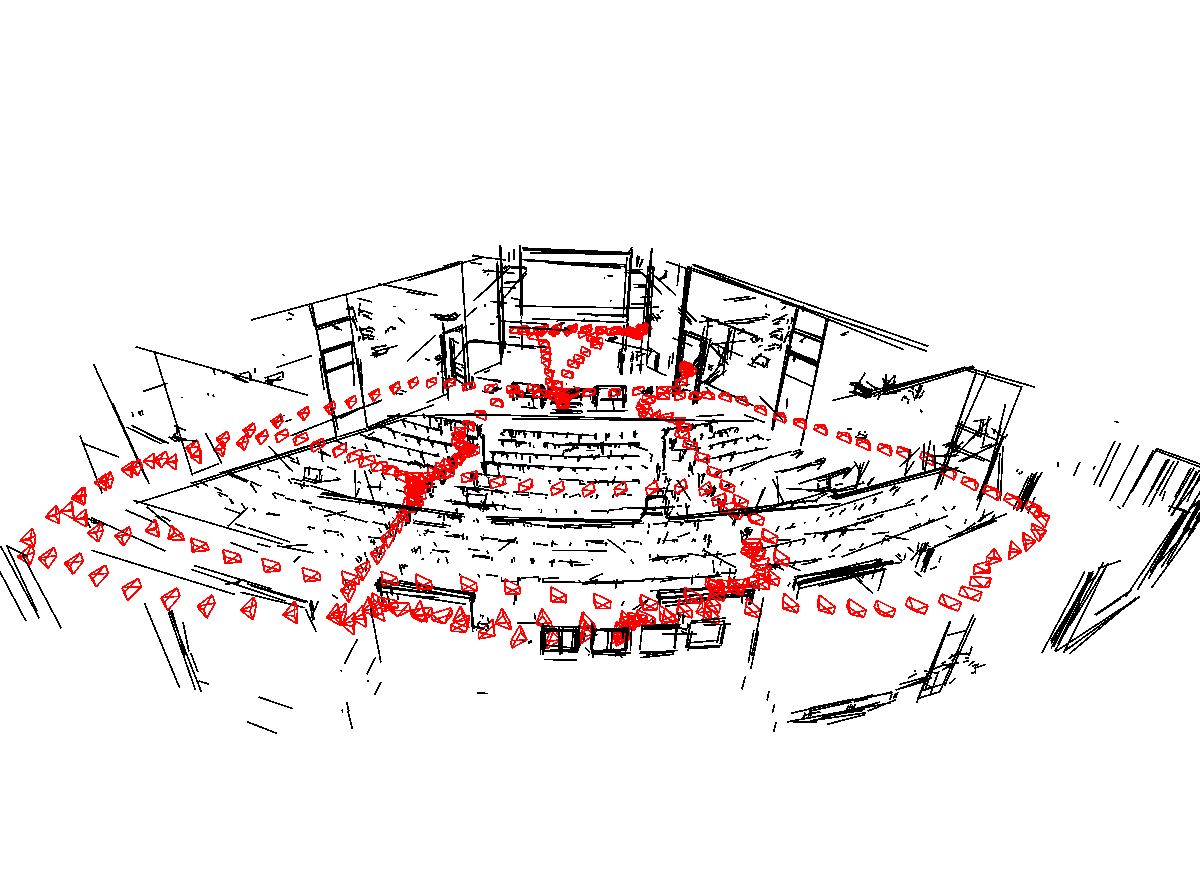}} \\
\textit{Old Hospital} from \cite{kendall2015posenet} &
\textit{St. Mary's Church} from \cite{kendall2015posenet} &
\textit{Auditorium} from \cite{Knapitsch2017}
\\
{\includegraphics[width=0.3\linewidth, height=114pt]{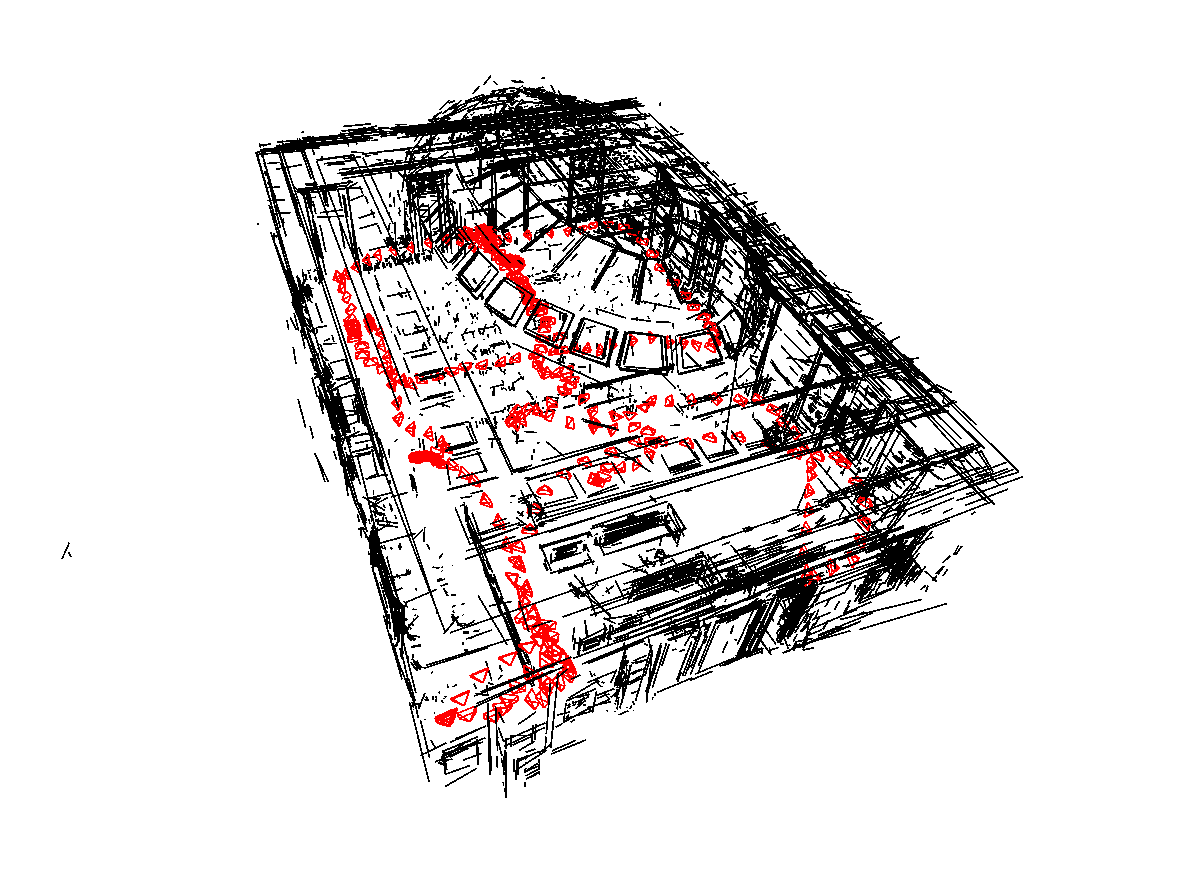}
\hspace{-0.09\linewidth}\includegraphics[width=0.09\linewidth]{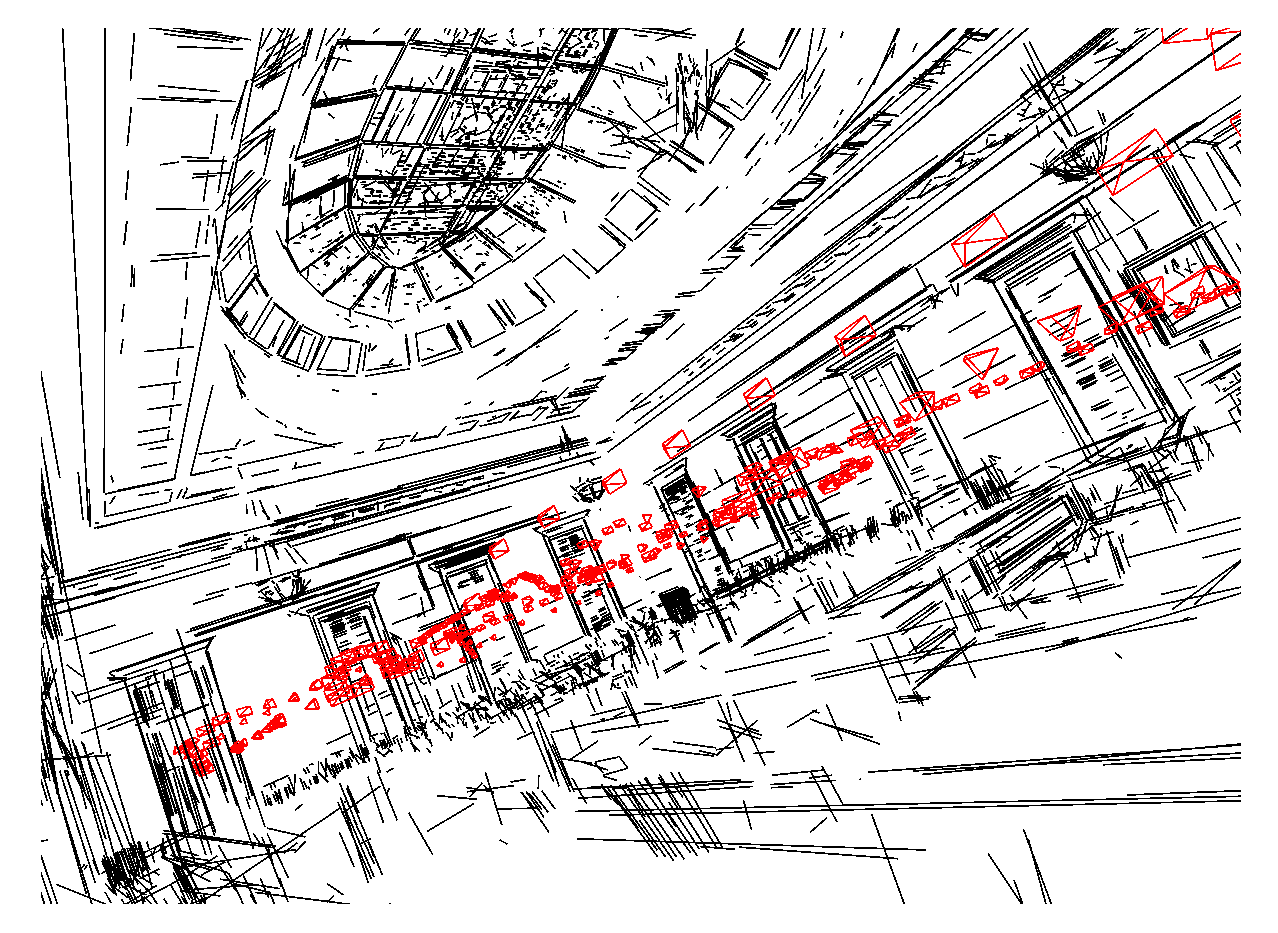}} & 
{\includegraphics[width=0.3\linewidth, height=114pt]{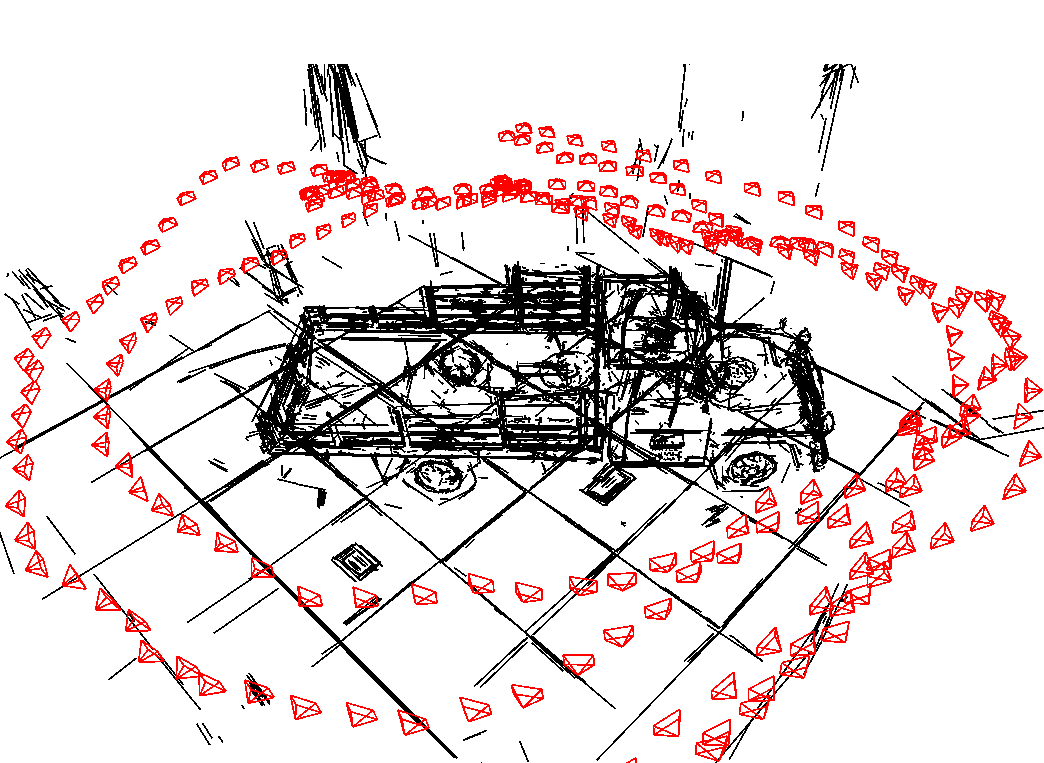}} & 
{\includegraphics[width=0.3\linewidth, height=114pt]{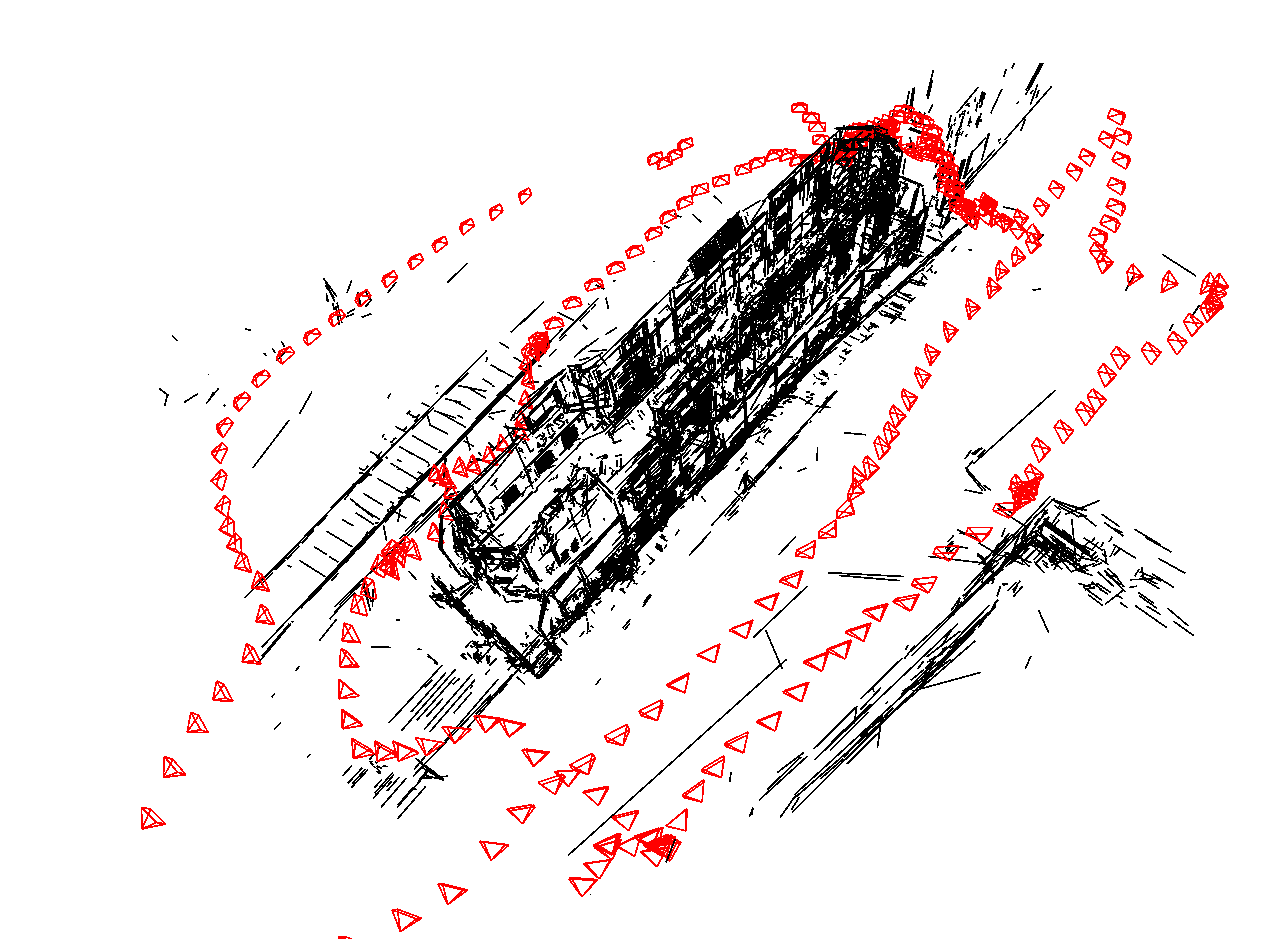}} \\
\textit{Courtroom} (indoor and outdoor) from \cite{Knapitsch2017} &
\textit{Truck} from \cite{Knapitsch2017} &
\textit{Train} from \cite{Knapitsch2017}
\\
{\includegraphics[width=0.3\linewidth, height=114pt]{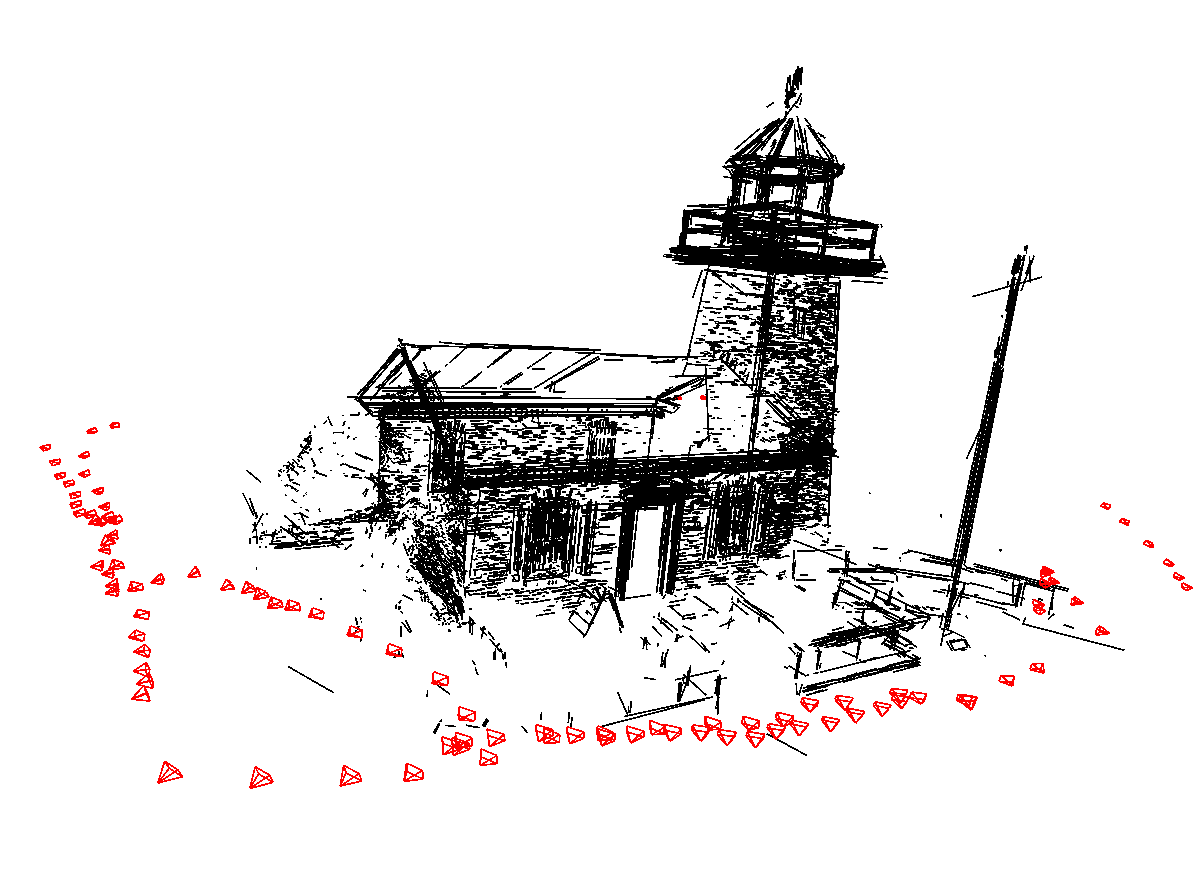}} & 
{\includegraphics[width=0.3\linewidth, height=114pt]{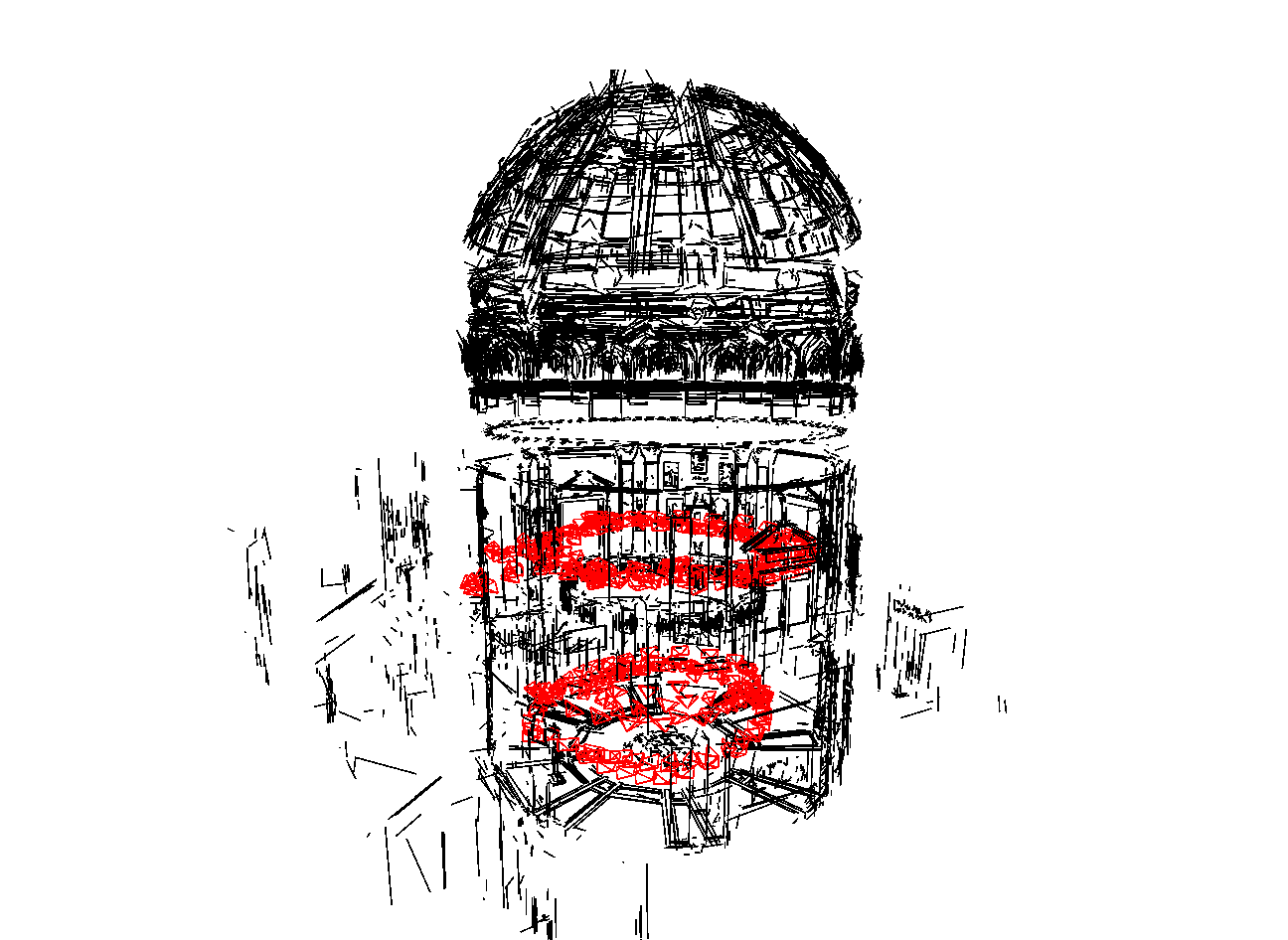}} & 
{\includegraphics[width=0.3\linewidth, height=114pt]{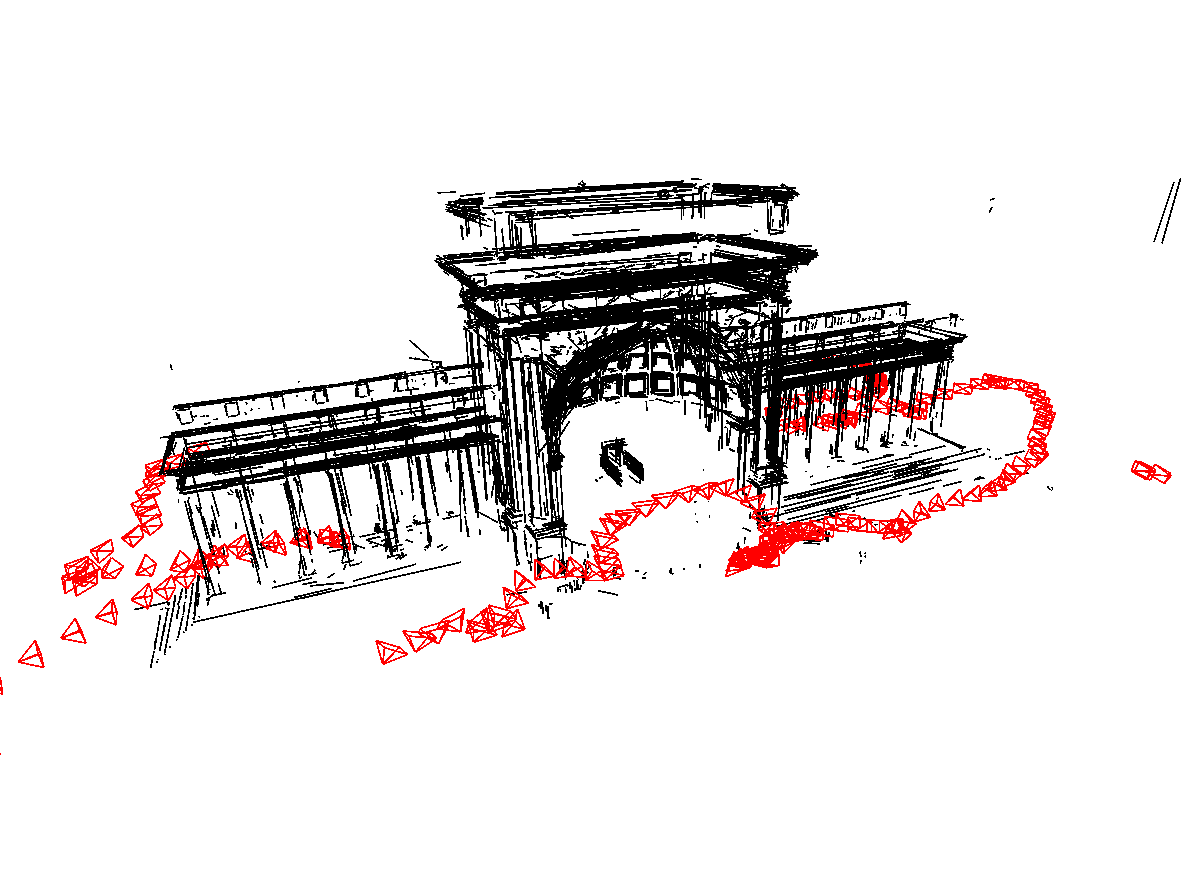}} \\
\textit{Lighthouse} from \cite{Knapitsch2017} &
\textit{Museum} from \cite{Knapitsch2017} &
\textit{Temple} from \cite{Knapitsch2017}
\\
\end{tabular}
\centering
\caption{\textbf{More qualitative results of the 3D line maps recovered by our framework.}}
\label{fig::supp_qualitative}
\end{figure*}

To further highlight the advantage of our proposed line mapping over L3D++ \cite{hofer2017efficient}, we study the recall-precision trade-off by relaxing the requirements for minimum number of supporting images in the final output 3D line tracks from 4 views to 3. This is actually the default setting for L3D++ \cite{hofer2017efficient} but we updated it in our main experiments for a fair comparison. We also compare with the default hyperparameters used in L3D++ \cite{hofer2017efficient} using 10 visual neighbors and 0.25 IoU threshold for the weak epipolar constraints.

Tables \ref{tab::main-hypersim-nv3} and \ref{tab::main-tnt-nv3} show the results on Hypersim \cite{roberts:2021} and \textit{Tanks and Temples} \cite{Knapitsch2017} respectively. The relative positions of L3D++ \cite{hofer2017efficient} and our method are similar when we relax the required minimum supporting images to 3 views. Here it is worth mentioning that, when comparing our method with nv = 4 against L3D++ \cite{hofer2017efficient} with nv = 3 (the default in L3D++ release), we can see that our method is significantly better in both the length recall and precision on all thresholds. This further demonstrates our advantages over L3D++ \cite{hofer2017efficient} on the precision-recall curve. The performance gain becomes larger when comparing our line mapping with L3D++ \cite{hofer2017efficient} using its default parameters. Since ELSR \cite{wei2022elsr} does not provide 2D-3D track association, we cannot filter their output lines with a minimum number of supporting views.

\subsection{Qualitative Results of ELSR \cite{wei2022elsr}}

For completeness, we compare qualitatively our mapping results with those from ELSR \cite{wei2022elsr} in Figure \ref{fig::elsr_comparison}. While ELSR \cite{wei2022elsr} is able to produce reasonable 3D line maps, it often fails to recover lines where the point and plane features are limited. On the contrary, our method recovers significantly more complete structures with better accuracy, and provides rich 2D-3D track association that is critical for downstream applications.

\subsection{Extensibility to Different Line Detectors and Matchers}

To further show the flexibility of our framework to be extended to different line detectors and matchers, we test over several existing line detectors \cite{von2008lsd,xue2020holistically,huang2020tp,pautrat2021sold2} and matchers \cite{zhang2013efficient,pautrat2021sold2,abdellali2021l2d2,yoon2021line} on the first eight scenes of Hypersim \cite{roberts:2021}. We also present two new matchers that are based on endpoint correspondences. Specifically, we extract SuperPoint features \cite{detone2018superpoint} over the two endpoints of the line and measure the structured endpoint distance with either nearest neighbor matching or an advanced point matcher SuperGlue \cite{sarlin2020superglue}. 

Table \ref{tab:main_different_line_methods} shows the results of all $6 \times 4$ combinations. Some interesting facts can be observed from the table. TP-LSD \cite{huang2020tp} achieves the highest precision with L2D2 \cite{abdellali2021l2d2}, while SOLD2 \cite{pautrat2021sold2} achieves the highest length recall. LSD \cite{von2008lsd} is consistently good on the length recall, while struggling on precision and track association due to its nature of being less structural. LineTR \cite{yoon2021line} is particularly good at matching TP-LSD \cite{huang2020tp} lines, while struggling on other detections compared to other matchers. 

We also added the recent strong detectors DeepLSD \cite{Pautrat_2023_DeepLSD} for completeness. Results in the last column of Table \ref{tab:main_different_line_methods} show that DeepLSD \cite{Pautrat_2023_DeepLSD} largely outperforms all competing methods in terms of both length recall and track quality. The number of supporting lines for DeepLSD \cite{Pautrat_2023_DeepLSD} is second best due to the highly occurring over-segmented lines in SOLD2 \cite{pautrat2021sold2}.

The endpoint-based line matcher is surprisingly effective, as ``Endpoint SP + SG" consistently achieves the best track association under all detectors. We further test again the comparison between endpoint triangulation and algebraic line triangulation to see whether the endpoints of the matched line from the endpoint-based line matcher correspond to each other. Results in Table \ref{tab::supp_ablations_endpoints} show similar trends as Table 3 in the main paper, where performing algebraic line triangulation is significantly better than directly triangulating endpoints. This finding indicates that the endpoint-based matcher is surprisingly effective on matching lines with, however, unmatched endpoints, which may be due to the advantages of the rich point features \cite{detone2018superpoint}. This encourages more research towards integrating the success of existing point description and matching solutions to improve line matching.

We believe that, with the flexible design and modular Python bindings, our line mapping system can help benchmarking and facilitate the progress of developing advanced line detection and matching algorithms.


\subsection{More Qualitative Results of Our Line Maps}

We show more qualitative results of our reconstructed 3D line maps across datasets \cite{kendall2015posenet, snavely2006photo, Knapitsch2017} in Figure \ref{fig::supp_qualitative}.

\section{More Results on Visual Localization}
\label{sec::supp_localization}

In this section, we first present the design details of our proposed visual localization pipeline with points and lines. Then, we provide experimental details on Cambridge \cite{kendall2015posenet} and 7Scenes \cite{7scenes} as well as per-scene results. Finally, we show additional results on the large-scale InLoc dataset \cite{taira2018inloc}.

\subsection{Details on Our Visual Localization Pipeline}
The input to the proposed visual localization pipeline is a set of 2D-3D point correspondences (from point-based SfM model, e.g. COLMAP \cite{schonberger2016structure}, or depth maps) and line correspondences (from LIMAP). We directly use the point correspondences processed by HLoc \cite{hloc}. Our pipeline is implemented within a hybrid RANSAC framework \cite{camposeco2018hybrid,Sattler2019Github} with local optimization \cite{chum2003locally,Lebeda2012BMVC}. In the hybrid RANSAC we combine four different minimal solvers on 2D-3D point correspondences (PCs) and 2D-3D line correspondences (LCs) in the following:

\begin{itemize} [noitemsep,nolistsep]
    \item P3P \cite{persson2018lambda}: 3 PCs.
    \item P2P1LL \cite{zhou2018stable}: 2 PCs + 1 LC.
    \item P1P2LL \cite{zhou2018stable}: 1 PC + 2 LCs.
    \item P3LL \cite{zhou2018stable}: 3 LCs.
\end{itemize}

We take the implementation from PoseLib \cite{PoseLib} for all four solvers to solve for the absolute camera pose. Following \cite{camposeco2018hybrid}, the sampling probability and termination criterion of each solver depends on the inlier ratio. For scoring the model, we measure reprojection errors on both the PCs and LCs. Specifically, 2D perpendicular distance is employed for lines. Additionally, we perform cheirality tests on both PCs and LCs. The cheirality test for a 2D-3D line correspondence is done by unprojecting both 2D endpoints onto the 3D infinite line. This is achieved by projecting the camera rays onto the infinite 3D line as discussed in Section \ref{sec::supp_plucker_linetoline}. Also, a 2D-3D line correspondence is considered an outlier if the length of the 2D reprojection of the 3D line segment is less than 1 pixel. For local optimization, the joint point-line refinement is applied with 2D reprojection error (perpendicular distance for lines) with an optional weighted Huber Loss via Ceres \cite{ceres}, where the weights for points and lines are set similarly to the weights according to the numbers of PCs and LCs. We apply the final least square optimization on all the inliers after RANSAC terminates.

\subsection{Details and Per-Scene Results on Cambridge and 7Scenes}
\begin{table}[tb]
    \centering
    \scriptsize
    \setlength{\tabcolsep}{5pt}
    \begin{tabular}{lccc}
        \toprule
        Scene & HLoc~\cite{hloc} & PtLine~\cite{gao2022pose} & Ours \\
        \midrule
        Great Court & \textbf{9.5} / \textbf{0.05} / \textbf{20.4} & 11.2 / 0.07 / 17.8 & 9.6 / \textbf{0.05} / 20.3 \\ 
        King's College & 6.4 / \textbf{0.10} / 37.0 & 6.5 / \textbf{0.10} / 37.0 & \textbf{6.2} / \textbf{0.10} / \textbf{39.4} \\
        Old Hospital & 12.5 / 0.23 / 22.5 & 12.7 / 0.24 / 20.9 & \textbf{11.3} / \textbf{0.22} / \textbf{25.4} \\
        Shop Facade & 2.9 / 0.14 / 78.6 & \textbf{2.7} / \textbf{0.12} / 79.6 & \textbf{2.7} / 0.13 / \textbf{81.6} \\ 
        St.Mary's Church & \textbf{3.7} / 0.13 / 61.7 & 4.1 / 0.13 / 62.3 & \textbf{3.7} / \textbf{0.12} / \textbf{63.8} \\
        \midrule
        Avg. & 7.0 / 0.13 / 44.0 & 7.4 / 0.13 / 43.5 & \textbf{6.7} / \textbf{0.12} / \textbf{46.1} \\
        \bottomrule
    \end{tabular}\textbf{}
    \caption{Per-scene results of visual localization on Cambridge Dataset~\cite{kendall2015posenet}. We report the median translation and rotation error in cm and degrees, and the pose accuracy(\%) at 5 cm / 5 deg threshold.}
    \label{tab:cambridge}
\end{table}

\begin{table}[tb]
    \centering
    \scriptsize
    \setlength{\tabcolsep}{3pt}
    \begin{tabular}{lccc}
        \toprule
        Scene & HLoc\cite{hloc} & PtLine\cite{gao2022pose} & Ours \\
        \midrule
        Chess & \textbf{2.4} / \textbf{0.84} / \textbf{93.0} & \textbf{2.4} / 0.85 / 92.7 & 2.5 / 0.85 / 92.3 \\ 
        Fire & 2.3 / 0.89 / 88.9 & 2.3 / 0.91 / 87.9 & \textbf{2.1} / \textbf{0.84} / \textbf{95.5} \\
        Heads & \textbf{1.1} / \textbf{0.75} / \textbf{95.9} & 1.2 / 0.81 / 95.2 & \textbf{1.1} / 0.76 / \textbf{95.9} \\
        Office & 3.1 / 0.91 / 77.0 & 3.2 / 0.96 / 74.5 & \textbf{3.0} / \textbf{0.89} / \textbf{78.4} \\
        Pumpkin & 5.0 / 1.32 / 50.4 & 5.1 / 1.35 / 49.0 & \textbf{4.7} / \textbf{1.23} / \textbf{52.9} \\
        Redkitchen & 4.2 / \textbf{1.39} / 58.9 & 4.3 / 1.42 / 58.0 & \textbf{4.1} / \textbf{1.39} / \textbf{60.2} \\
        Stairs & 5.2 / 1.46 / 46.8 & 4.8 / 1.33 / 51.9 & \textbf{3.7} / \textbf{1.02} / \textbf{71.1} \\  
        \midrule
        Avg. & 3.3 / 1.08 / 73.0 & 3.3 / 1.09 / 72.7 & \textbf{3.0} / \textbf{1.00} / \textbf{78.0} \\
        \bottomrule
    \end{tabular}
    \caption{Per-scene results of visual localization on 7Scenes~\cite{7scenes}. We report the median translation and rotation error in cm and degrees, as well as the pose accuracy at a 5 cm / 5 deg threshold.}
    \label{tab:7scenes}
\end{table}

For the experiments on both Cambridge \cite{kendall2015posenet} and 7Scenes \cite{7scenes} Datasets, we run our line mapping system with LSD line detections \cite{von2008lsd} and SOLD2 matching \cite{pautrat2021sold2}. On both datasets, the point-alone baseline employs the best method combination in HLoc \cite{hloc}: NetVLAD \cite{arandjelovic2016netvlad} + SuperPoint \cite{detone2018superpoint} + SuperGlue \cite{sarlin2020superglue}.
For Cambridge \cite{kendall2015posenet}, we also follow HLoc \cite{hloc} to resize the images to $1024 \times 576$. The inlier thresholds of our hybrid RANSAC for both points and lines are set to 6 pixels on Cambridge \cite{kendall2015posenet} and 5 pixels on 7Scenes \cite{7scenes}. Up to the date of submission, the triangulated COLMAP model \cite{schonberger2016structure} for Cambridge from the official repository of HLoc \cite{hloc} does not consider the radial distortion in the VisualSfM model \cite{wu2011visualsfm}. Fixing the issue and re-triangulating the point-based 3D maps result in much better performance than the original one, so we use our updated one as the point-based baseline to evaluate our results. Our design with hybrid RANSAC enables direct comparison since disabling the three line solvers will fall into a point-based RANSAC with P3P \cite{persson2018lambda} which is equivalent to HLoc \cite{hloc}. 

To compare with the recently proposed PtLine method \cite{gao2022pose}, we reimplement the match filtering and their midpoint-based post-refinement strategy. Because both their line detector and their strong point-based localization baseline is not publicly available, we apply their method with our line mapping over the initial poses retrieved by HLoc \cite{hloc}. We tune the IoU threshold (0.4 for Cambridge, 0.2 for 7Scenes) for filtering to get the best results on both datasets. 

We here provide the per-scene results of both the PtLine \cite{gao2022pose} and our method on both datasets, in Tables \ref{tab:cambridge} and \ref{tab:7scenes} respectively, where our method consistently outperforms the point-based baseline \cite{hloc} and the joint point-line post-refinement from PtLine \cite{gao2022pose}. The results are further supported under settings when depth maps are available in Table \ref{tab:7scenes_fitnmerge}, as already discussed in Section \ref{sec::supp_fitnmerge}. 

\subsection{Results on InLoc dataset}
\begin{table}[tb]
    \centering
    \scriptsize
    \setlength{\tabcolsep}{4pt}
    \begin{tabular}{llcc}
        \toprule
         & & DUC 1 & DUC 2 \\
        \midrule
        Points & HLoc~\cite{sarlin2019coarse} & 49.0 / 69.2 / 80.3 & 52.7 / \textbf{77.1} / 80.9 \\
        \midrule
        \multirow{2}{*}{\makecell{Points\\+ Lines}}
         & PtLine~\cite{gao2022pose} & 49.0 / 69.2 / \textbf{81.8} & 56.5 / 76.3 / 80.2 \\
         & Ours & \textbf{49.5} / \textbf{72.2} / 81.3 & \textbf{60.3} / 76.8 / \textbf{81.7} \\
        \bottomrule
    \end{tabular}
    \caption{\textbf{Results of visual localization on InLoc~\cite{taira2018inloc}.} We report the pose AUC at 0.25m / 0.5m / 1m and 10 degrees error.}
    \label{tab:inloc}
\end{table}
We further test our method on InLoc dataset \cite{taira2018inloc}, again comparing with HLoc \cite{hloc} as our point-only baseline and PtLine \cite{gao2022pose} as the only joint point-line visual localization method. Results on both DUC1 and DUC2 are shown in Table \ref{tab:inloc}, where integrating line features again improves the performance of point-based solution, while our solution is consistently better than PtLine \cite{gao2022pose}. In particular, we improve over the point-only baseline HLoc \cite{hloc} on AUC @ 0.25m by 7.6 on DUC2, by simply combining lines and points in the hybrid RANSAC framework.

\section{More Results on Refining Structure-from-Motion}
\label{sec::supp_sfm_refinement}
\begin{table}[tb]
    \centering
    \scriptsize
    \setlength{\tabcolsep}{4pt}
    \begin{tabular}{lccc}
    \toprule
    & COLMAP \cite{schonberger2016structure} & \cite{schonberger2016structure} + LIMAP (line-only) & \cite{schonberger2016structure} + LIMAP \\
    \midrule
    \textit{ai\_001\_001} & 68.0 / 87.0 / 91.3 & 78.3 / 91.1 / 93.8 & \textbf{80.0} / \textbf{91.7} / \textbf{94.2} \\
    \textit{ai\_001\_002} & 75.2 / 90.2 / 94.0 & 87.5 / 95.6 / 97.3 & \textbf{88.5} / \textbf{96.0} / \textbf{97.6} \\
    \textit{ai\_001\_003} & 83.8 / 94.4 / 96.6 & 82.9 / 94.0 / 96.4 & \textbf{85.7} / \textbf{95.1} / \textbf{97.1} \\
    \textit{ai\_001\_004} & \textbf{79.2} / \textbf{88.9} / \textbf{90.9} & 67.1 / 82.1 / 86.0 & 77.3 / 88.3 / 90.6 \\
    \textit{ai\_001\_005} & 85.1 / 94.9 / 97.0 & 88.4 / 96.1 / 97.7 & \textbf{90.9} / \textbf{97.0} / \textbf{98.2} \\
    \textit{ai\_001\_006} & 83.4 / 93.1 / 95.7 & 80.2 / 92.9 / 95.7 & \textbf{84.4} / \textbf{93.8} / \textbf{96.3} \\
    \textit{ai\_001\_007} & 59.0 / 68.5 / 70.6 & 64.5 / \textbf{70.6} / \textbf{71.9} & \textbf{65.0} / 70.3 / 71.7 \\
    \textit{ai\_001\_008} & 84.9 / 94.9 / 96.9 & 89.5 / 96.5 / 97.9 & \textbf{91.3} / \textbf{97.1} / \textbf{98.2} \\
    \midrule
    Average $\uparrow$ & 77.3 / 89.0 / 91.6 & 79.8 / 89.9 / 92.1 & \textbf{82.9} / \textbf{91.2} / \textbf{93.0} \\ 
    \midrule
    Median error $\downarrow$ & 0.188 & 0.173 & \textbf{0.146} \\ 
    \bottomrule
    \end{tabular}
    \caption{Per-scene results of joint bundle adjustment of points and lines on Hypersim \cite{roberts:2021}. Relative pose errors are measured on all image pairs. AUC @ (1$^{\circ}$ / 3$^{\circ}$ / 5$^{\circ}$) are reported for each method and the average median error is reported an the bottom row.}
    \label{tab:sfm_hypersim_perscene}
\end{table}

We provide per-scene results in Table \ref{tab:sfm_hypersim_perscene} on the joint point-line bundle adjustment experiment presented in the main paper. Combining lines with points consistently improves the accuracy of point-based Structure-from-Motion (SfM) on 7 of the 8 scenes, with notable improvement particularly on AUC@1$^{\circ}$ thanks to better pixelwise alignment with the line structures. For completeness, we also show the results with line-only optimization in Table \ref{tab:sfm_hypersim_perscene}. When the line structures are rich in the scene, optimizing solely over lines from the initialization of point-based SfM is able to achieve reasonable results, while combining both points and lines give the best accuracy and stability.

\section{Line-Assisted Multi-view Stereo}
\label{sec::supp_mvs}
\begin{figure}[tb]
\scriptsize
\setlength\tabcolsep{2pt} 
\begin{tabular}{cccccc}
{\includegraphics[width=0.22\linewidth]{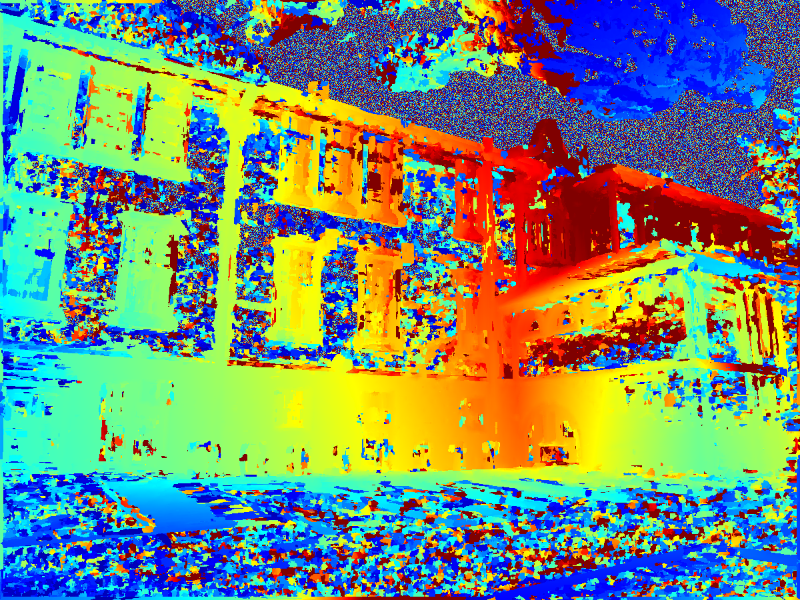}} & 
{\includegraphics[width=0.22\linewidth]{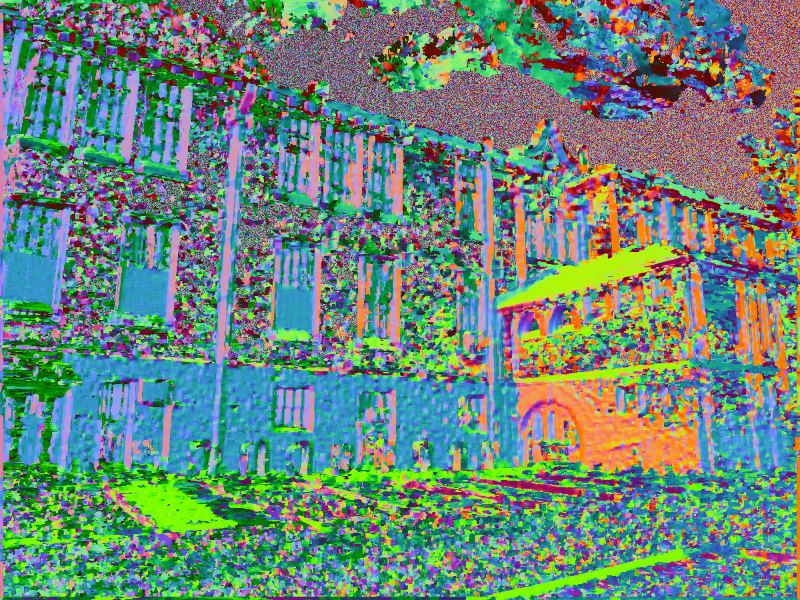}} & & &
{\includegraphics[width=0.22\linewidth]{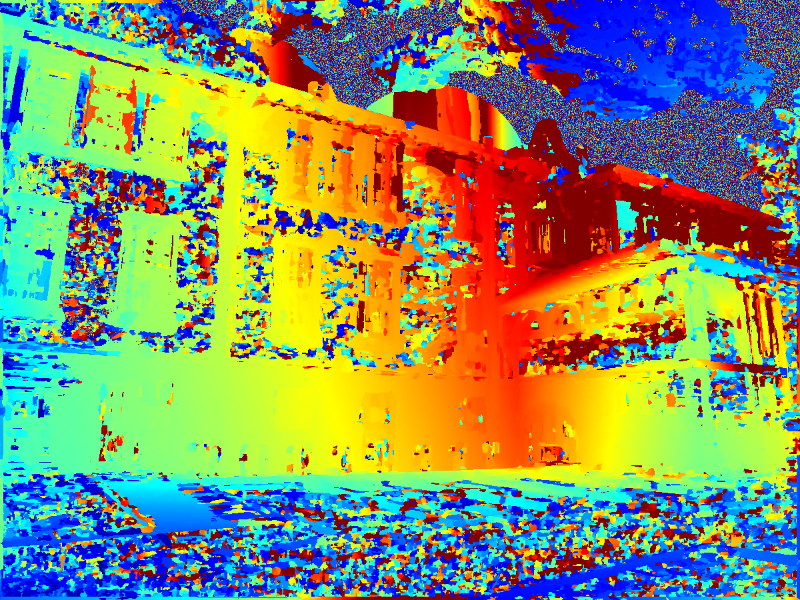}} &
{\includegraphics[width=0.22\linewidth]{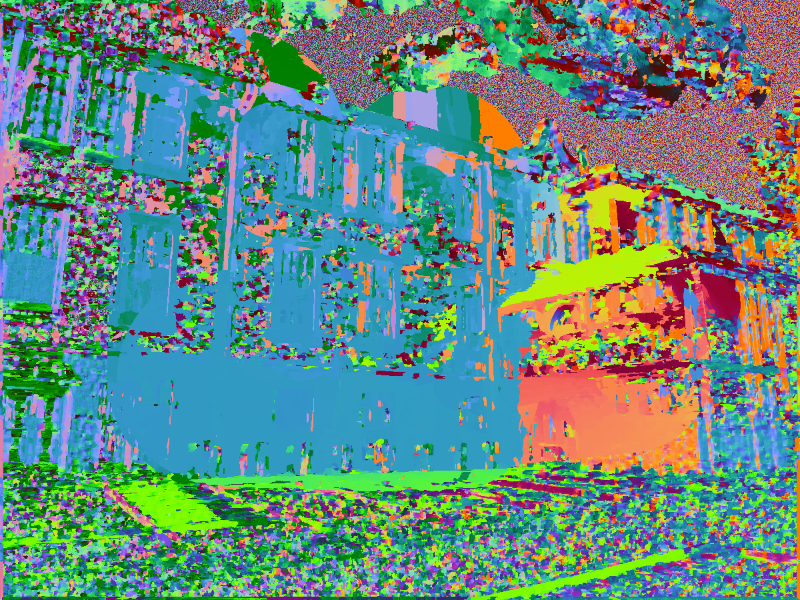}} \\
\multicolumn{2}{c}{Original COLMAP MVS \cite{schonberger2016pixelwise}} & & & \multicolumn{2}{c}{Same with line-based energy}
\end{tabular}
\centering
\caption{\textbf{Line-assisted multi-view stereo (MVS).} Visualization (from COLMAP \cite{schonberger2016structure,schonberger2016pixelwise}) of the depth and normal maps from COLMAP MVS \cite{schonberger2016pixelwise}. Integrating line-based energy into PatchMatch Stereo largely improves completeness.}
\label{fig::results_mvs}
\end{figure}

In this section, we discuss on how to integrate the acquired line maps into PatchMatch Stereo \cite{schonberger2016structure} with the assumption of local planarity. Specifically, at each iteration of the PatchMatch Stereo pipeline, we can add an additional line-based energy that encourages the depth and normal at each pixel to span a plane that crosses some nearby 3D line segments. In practice, for each pixel, we collect all lines that have projections within a 2D perpendicular distance of half the line length. During the proposal selection in PatchMatch Stereo \cite{schonberger2016pixelwise}, we compute the perpendicular distances (sum of the perpendicular distance for both endpoints) of all collected lines to the corresponding plane proposal (spanned at each pixel with its depth and normal in PatchMatch), and sum the two minimum distances from the two closest lines as the line-based energy. This encourages the selected depth-normal pair to span a plane having at least two 3D lines that are very close, implicitly encouraging local planarity on the recovered surface with respect to the 3D line maps.

We show one qualitative result on AdelaideRMF \cite{fan2012robust} in Figure \ref{fig::results_mvs} to illustrate the advantage of line-assisted PatchMatch Stereo \cite{schonberger2016pixelwise}. With the line-based energy, both the depth map and surface normal map become more complete in texture-less regions, and the surface normal map is comparably more smooth thanks to the local planarity implicitly enforced by the 3D line maps.

\section{Extension: Featuremetric Line Refinement}
\label{sec::supp_featuremetric}
\begin{figure}[tb]
\scriptsize
\setlength\tabcolsep{2pt} 
\begin{tabular}{ccc}
{\includegraphics[trim={0 5 0 10}, clip, width=0.32\linewidth, height=70pt]{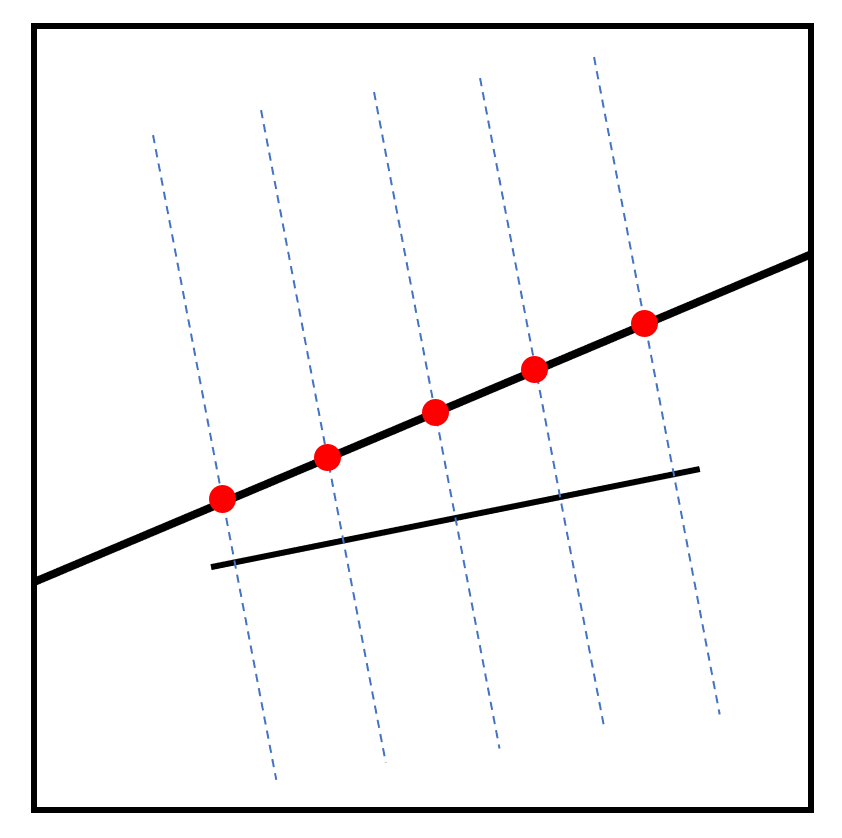}} &
{\includegraphics[trim={0 20 0 17}, clip, width=0.32\linewidth, height=70pt]{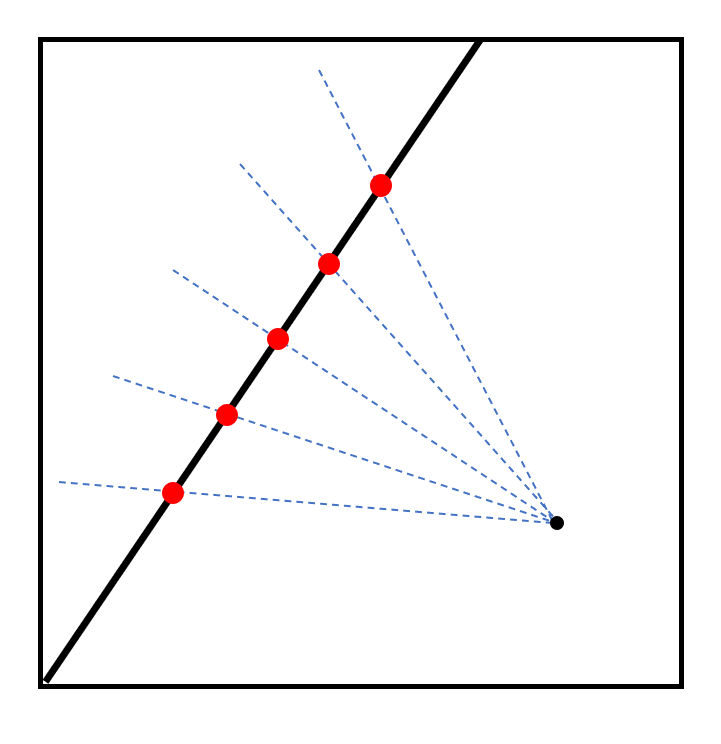}} & 
{\includegraphics[width=0.32\linewidth, height=70pt]{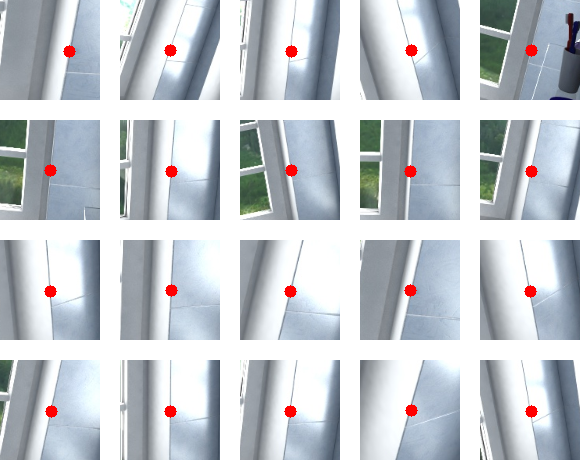}} \\
reference image & target image & correspondences \\
\end{tabular}
\centering
\caption{Illustration on how to extend featuremetric optimization \cite{lindenberger2021pixel} over 3D line tracks.}
\label{fig::featuremetric}
\end{figure}

Inspired by the recent success of featuremetric refinement for point-based Structure-from-Motion~\cite{sarlin2021back,lindenberger2021pixel}, we here present ideas on how to extend its application for 3D line refinement. This type of refinement can be potentially very suitable for lines, since compared to point-based alternatives, line detectors usually have higher localization errors in the image. For points, it is straightforward to define featuremetric consistency loss by simply interpolating the feature map at the point locations and computing the difference. To apply the same framework for lines, it is also necessary to establish point-wise correspondences along the line to be able to measure the feature consistency. One approach is to parameterize the 3D line endpoints explicitly and sample points between them. The two main drawbacks of this approach are that directly optimizing over the endpoints might suffer from endpoint collapse, and that during optimization, the supporting images for each sampled point might change as the endpoints shift.

To avoid these issues, we here present an alternative formulation that instead optimizes over the infinite 3D line and defines the sample points directly in 2D. The idea is to parameterize the sampling of points through line intersection (Figure \ref{fig::featuremetric}). Specifically, we first uniformly sample 3D points between the two initial endpoints. For each sampled 3D point, we determine the supporting images and 2D line segments. From these 2D line segments, we select the reference line segment as the longest one and construct a perpendicular 2D line based on the projection. The sampled 3D points are then discarded. For the optimization, we project the infinite 3D lines onto the images, which are then intersected with the perpendicular 2D lines to give us the 2D sample points in the reference views. These sample points are then mapped to epipolar lines in the set of supporting images, and then by intersecting with the projections of the 3D line, we get the corresponding sample points in the other views. With these point-wise correspondences, we can compute the featuremetric consistency loss used in the optimization. This is illustrated in Figure~\ref{fig::featuremetric}. The infinite line can again be minimally parameterized with Pl\"ucker coordinate discussed in Section \ref{sec::supp_plucker}.

To reduce memory requirements for the feature maps, a specially designed \textit{line patching} strategy can be employed, where an oriented bounding box around the lines is extracted with bilinear sampling on the non-integral coordinates. The 2D rotation and translation are stored along with each line patch for the local-global coordinate transformation. This \textit{line patching} can significantly save memory while still allowing us to accurately interpolate the features of the sampled points lying close to the line segment.

\section{Limitations and Future Work}
\label{sec::supp_limitations}
In this section, we discuss the current limitations of the proposed system and give an overview of potential areas for improvement and future work.

The current system is designed to reconstruct 3D lines from known camera poses, i.e.~we focused on the mapping step of the reconstruction pipeline. While we show in the experiments that the system is robust to imperfect camera poses (e.g.~obtained from SLAM or SfM), it is still dependent on the point-based reconstruction/tracking working and we cannot recover if they fail.
Interesting future work is to integrate our mapping pipeline into an incremental Structure-from-Motion framework. We believe this is a promising direction as  our experiments show that both localization (i.e.~registering a new image to a reconstruction) and bundle adjustment can be improved with our line maps.

The system is also dependent on the reliability of the employed line detector and matcher, as shown in Table \ref{tab:main_different_line_methods}. Improving the detection, description, and matching of 2D lines can benefit a lot on the resulting 3D line maps. While this is beyond the scope of this paper, by sharing our library with the community we hope to facilitate relevant research developments on 2D line-related practice. 

The current system design is partly due to the fact that we have weaker detectors and matchers for lines compared to points. This requires more excessive geometric verification before building tracks compared to point-based triangulation methods which can be more greedy in their selection. While line triangulation is inherently less stable than point triangulation, improvements in line detection and matching might lessen the need for geometric verification.

{\small
\bibliographystyle{ieee_fullname}
\bibliography{egbib}
}

\end{document}